\documentclass{article}
\usepackage{xcolor}
\usepackage{enumitem}
\usepackage{tcolorbox}
\usepackage{booktabs}       
\usepackage{multirow}       
\usepackage[table]{xcolor}  
\usepackage{graphicx}
\usepackage{enumitem}
\usepackage{amsmath}
\usepackage{amssymb}
\usepackage{xcolor}
\usepackage{tcolorbox}
\usepackage{fontawesome5}
\usepackage{adjustbox}
\tcbuselibrary{listings, skins, breakable}

\newtcolorbox{findingbox}[1]{
    enhanced,
    breakable,
    colback=blue!3,
    colframe=blue!55!black,
    boxrule=0.5pt,
    arc=2pt,
    left=5pt,
    right=5pt,
    top=4pt,
    bottom=4pt,
    fonttitle=\bfseries,
    title={#1}
}

\usepackage[preprint]{neurips_2026}
\setcitestyle{numbers,square,sort&compress}

\usepackage[utf8]{inputenc} 
\usepackage[T1]{fontenc}    
\usepackage{hyperref}       
\usepackage{url}            
\usepackage{booktabs}       
\usepackage{amsfonts}       
\usepackage{nicefrac}       
\usepackage{microtype}      
\usepackage{xcolor}         

\usepackage[utf8]{inputenc} 
\usepackage[T1]{fontenc}    
\usepackage{hyperref}       
\usepackage{url}            
\usepackage{booktabs}       
\usepackage{amsfonts}       
\usepackage{nicefrac}       
\usepackage{microtype}      
\usepackage{xcolor}         
\usepackage{multirow}
\usepackage{graphicx}
\usepackage{amsmath}
\usepackage{subfigure}
\usepackage{wrapfig}
\usepackage{listings}
\usepackage{hyperref}      
\usepackage{fontawesome5}  
\definecolor{codegray}{rgb}{0.5,0.5,0.5}
\definecolor{codepurple}{rgb}{0.58,0,0.82}
\definecolor{backcolour}{rgb}{0.96,0.96,0.96} 
\definecolor{framecolor}{rgb}{0.85,0.85,0.85} 
\definecolor{deepgreen}{HTML}{A9D08E} 
\definecolor{lightgreen}{HTML}{E2EFDA}
\newcommand{\best}[1]{\begingroup\setlength{\fboxsep}{1pt}\colorbox{deepgreen}{#1}\endgroup}
\newcommand{\secondbest}[1]{\begingroup\setlength{\fboxsep}{1pt}\colorbox{lightgreen}{#1}\endgroup}
\definecolor{deepred}{HTML}{E6B8B7}
\definecolor{lightred}{HTML}{FBE5E5}
\newcommand{\bestred}[1]{\begingroup\setlength{\fboxsep}{1pt}\colorbox{deepred}{#1}\endgroup}
\newcommand{\secondbestred}[1]{\begingroup\setlength{\fboxsep}{1pt}\colorbox{lightred}{#1}\endgroup}
\definecolor{citecolor}{HTML}{0071bc}
\tcbset{
  simpleprompt/.style={
    breakable, enhanced,
    boxrule=0.7pt, arc=2mm,
    colframe=black!80,              
    left=6pt,right=6pt,top=6pt,bottom=6pt,
    fonttitle=\bfseries,
  }
}
\newtcolorbox{prompt}[2][]{simpleprompt,
  colback=citecolor!6!white,        
  title={#2}, #1}
\title{How Mobile World Model Guides GUI Agents?}

\author{
    \textbf{Weikai Xu$^{1*}$, Kun Huang$^{2*}$, Yunren Feng$^{3*}$, Jiaxing Li$^{1}$, Yuhan Chen$^{1}$, Yuxuan Liu$^{4}$} \\
    \textbf{Zhizheng Jiang$^{3}$, Heng Qu$^{5}$, Pengzhi Gao$^{2}$, Wei Liu$^{2}$, Jian Luan$^{2}$, Xiaolin Hu$^{6}$, Bo An$^{1\ddagger}$} \\
    \textmd{} \\
  $^1$\textmd{Nanyang Technological University} 
  $^2$\textmd{MiLM Plus, Xiaomi Inc.} 
  $^3$\textmd{Independent Researchers} \\
  $^4$\textmd{Gaoling School of Artificial Intelligence, Renmin University of China} 
   \\
   $^5$\textmd{Wuhan University} 
   $^6$\textmd{Xiamen University} \\
 $^*$\textmd{ Equal contribution.}
 $\ddagger$\textmd{ Corresponding authors.} \\
\href{https://huggingface.co/spaces/xwk123/mobileworldmodel}{\faIcon{globe}} \textmd{Project}
\quad
 \href{https://huggingface.co/datasets/xwk123/Mobile-GUI-Worldmodel-SFT}{\faIcon{database}} \textmd{Dataset}
 \quad
\href{https://hf.co/collections/xwk123/mobileworldmodel}{\includegraphics[height=1em]{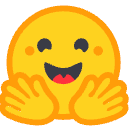}} \textmd{Model}
  }

\begin{document}

\maketitle

\begingroup
\renewcommand{\thefootnote}{}
\footnotetext{
Emails: \texttt{WEIKAI002@e.ntu.edu.sg}, \texttt{huangkun813@gmail.com}, and \texttt{yunrenfeng@gmail.com}.}
\endgroup

\begin{abstract}
Recent advances in vision-language models have enabled mobile GUI agents to perceive visual interfaces and execute user instructions, but reliable prediction of action consequences remains critical for long-horizon and high-risk interactions.
Existing mobile world models provide either text-based or image-based future states, yet it remains unclear which representation is useful, whether generated rollouts can replace real environments, and how test-time guidance helps agents of different strengths.
To answer the above questions, we filter and annotate mobile world-model data, then train world models across four modalities: delta text, full text, diffusion-based images, and renderable code.
These models achieve SoTA performance on both MobileWorldBench and Code2WorldBench. 
Furthermore, by evaluating their downstream utility on AITZ, AndroidControl, and AndroidWorld, we obtain three findings.
First, renderable code reconstruction achieves high in-distribution fidelity and provides effective multimodal supervision for data construction, while text-based feedback is more robust for online out-of-distribution (OOD) execution.
Second, world-model-generated trajectories can provide transferable interaction experience in the training process and improve agents' end-to-end task performance, although these data do not preserve the original distribution.
Last, for overconfident mobile agents with low action entropy, posterior self-reflection provides limited gains, suggesting that world models are more effective as prior perception or training supervision than as universal post-hoc verifiers.
\end{abstract}

\begin{figure}[t]
    \centering
    \vspace{-0.35em}
    \includegraphics[width=\linewidth]{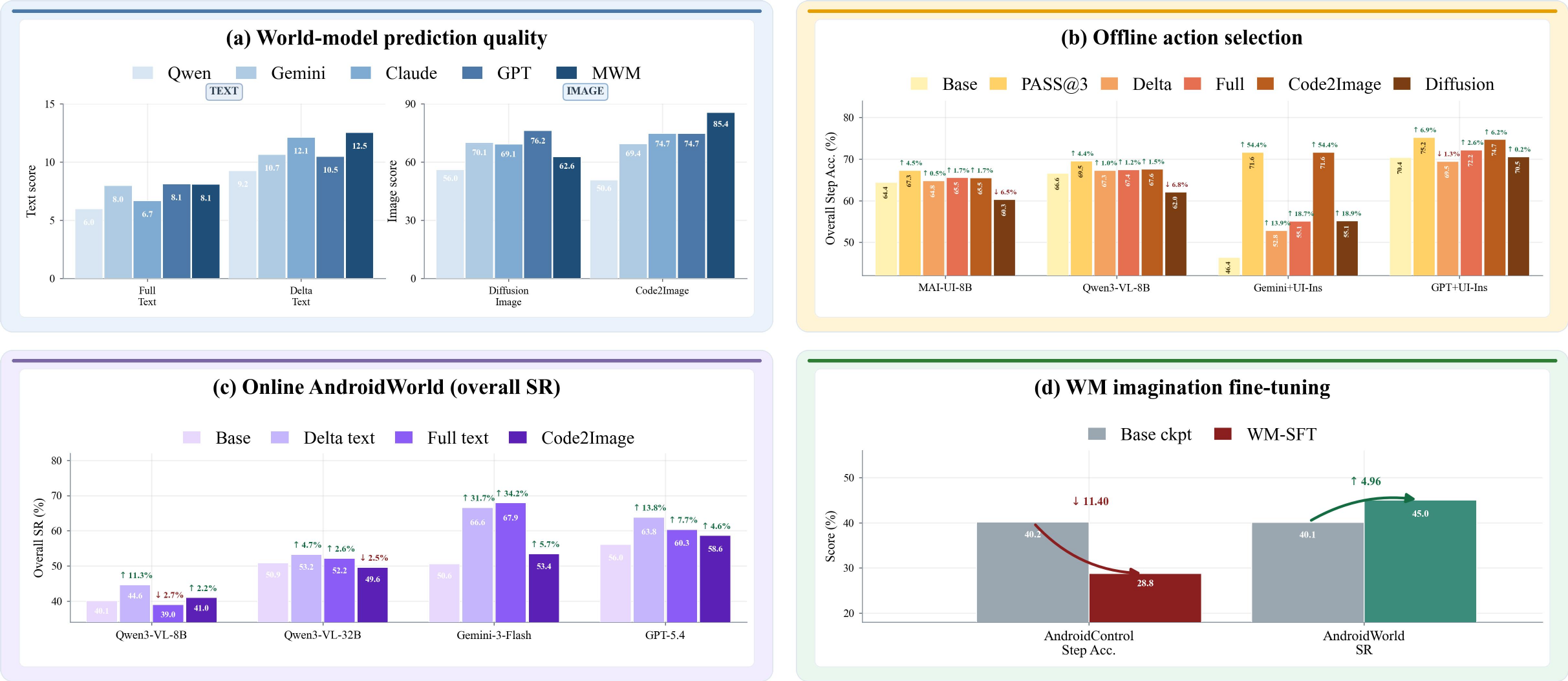}
    \vspace{-0.4em}
    \caption{%
    Overview of empirical results across prediction formats, test-time guidance, and imagination-based training.
    \textbf{(a)}~World-model prediction quality across judge VLMs and \textsc{MobileWorldModel} (MWM).
    \textbf{(b)}~Offline GUI navigation: overall step accuracy of base agents vs.\ world-model interaction variants (including PASS@3).
    \textbf{(c)}~Online AndroidWorld overall success rate (SR) under M3A with text vs.\ image feedback.
    \textbf{(d)}~Fine-tuning on world-model imagined trajectories (WM-SFT) vs.\ the base checkpoint on AndroidControl step accuracy and AndroidWorld SR\@.}
    \vspace{-0.4em}
    \label{fig:overview}
\end{figure}

\section{Introduction}
Recent advances in vision-language models~\citep{achiam2023gpt,guo2025seed1,yang2025qwen3} have enabled autonomous mobile agents~\citep{qin2025ui,yang2026gui,wang2025ui} to execute user instructions based on historical actions and current states. 
However, these agents still lack the foresight to anticipate the consequences of their actions.
Lacking such foresight prevents agents from distinguishing the divergent consequences of similar UI elements and avoiding irreversible or high-risk states, thereby depriving them of the opportunity for early self-reflection.
To bridge this gap, \textbf{Mobile World Models} can provide this crucial prospective capability for agents, serving as a simulator to predict the expected interface transitions following a candidate action.
While, mobile world modeling is challenging because UI transitions can be highly discontinuous: consecutive screens may be entirely different, the valid action space changes at every step, and the agent must reason over mixed visual, textual, and spatial signals. 
These properties make it unclear what kind of future-state prediction is most useful for downstream mobile agents.

A central design choice in mobile world models is therefore the prediction format exposed to the agent. 
According to the final form consumed by downstream agents, prior methods can be broadly divided into text-based and image-based outputs. 
Text-based outputs include \textit{Delta Text}, which describes only the state change after an action~\citep{chae2024web}, and \textit{Full Text}, which describes the entire next state~\citep{gu2024your,cao2026mobiledreamer}. 
Image-based outputs include \textit{Diffusion Image}, which directly synthesizes the future screenshot with generative models~\citep{luo2025vimo}, and \textit{Code2Image}, which first predicts renderable interface code, such as HTML or structured view representations, and then renders it into an image~\citep{koh2026generative,zheng2026code2world}. 
This taxonomy covers the major design choices in existing mobile world models while focusing on what information is ultimately provided to the downstream agent.

These prediction formats expose different trade-offs between semantic abstraction, visual fidelity, computational cost, and downstream interpretability. 
Text-based outputs are efficient and semantically informative, but may discard visual layout, spatial grounding, and fine-grained interface details important for mobile agents. 
Image-based outputs preserve richer visual information such as icons, colors, layout, and spatial relations, but may introduce visual hallucinations and require substantially higher training or inference costs. 
Among image-based methods, Code2Image provides a structured route to image prediction, but its reliability under distribution shift remains unclear. 
Moreover, the usefulness of a predicted future also depends on how the agent consumes it: a world model may serve as a prior perceptual module before action generation, a posterior critic for selecting among sampled actions, or a training-time proxy environment for producing imagined trajectories. 
Thus, the central question is not merely whether mobile world models can predict future states, but whether their predictions can be represented, consumed, and converted into useful agent behavior.

We therefore organize our study around three research questions:
\begin{itemize}[leftmargin=1.4em,itemsep=0.15em,topsep=0.25em]
    \item \textit{\textbf{RQ1: What Should Mobile World Models Predict?}}
    \item \textit{\textbf{RQ2: How Should {Mobile World Models} Assist {GUI Agents} at Test Time?}}
    \item \textit{\textbf{RQ3: Can {GUI Agents} Learn from {Mobile World Model} Imagination?}}
\end{itemize}

Before addressing these questions, Section~\ref{sec:mwm} defines the prediction formats and introduces \textsc{MobileWorldModel}, a transition-level training pipeline that instantiates text-based and image-based world models under a unified comparison setting. 
This shared foundation supports the reconstruction analysis, test-time guidance, and imagination-based training studied in Sections~\ref{sec:rq1}--\ref{sec:rq3}.

\begin{figure}[t]
    \setlength{\abovecaptionskip}{0.15cm}
    \setlength{\belowcaptionskip}{-0.1cm}
    \centering
    \includegraphics[width=\linewidth]{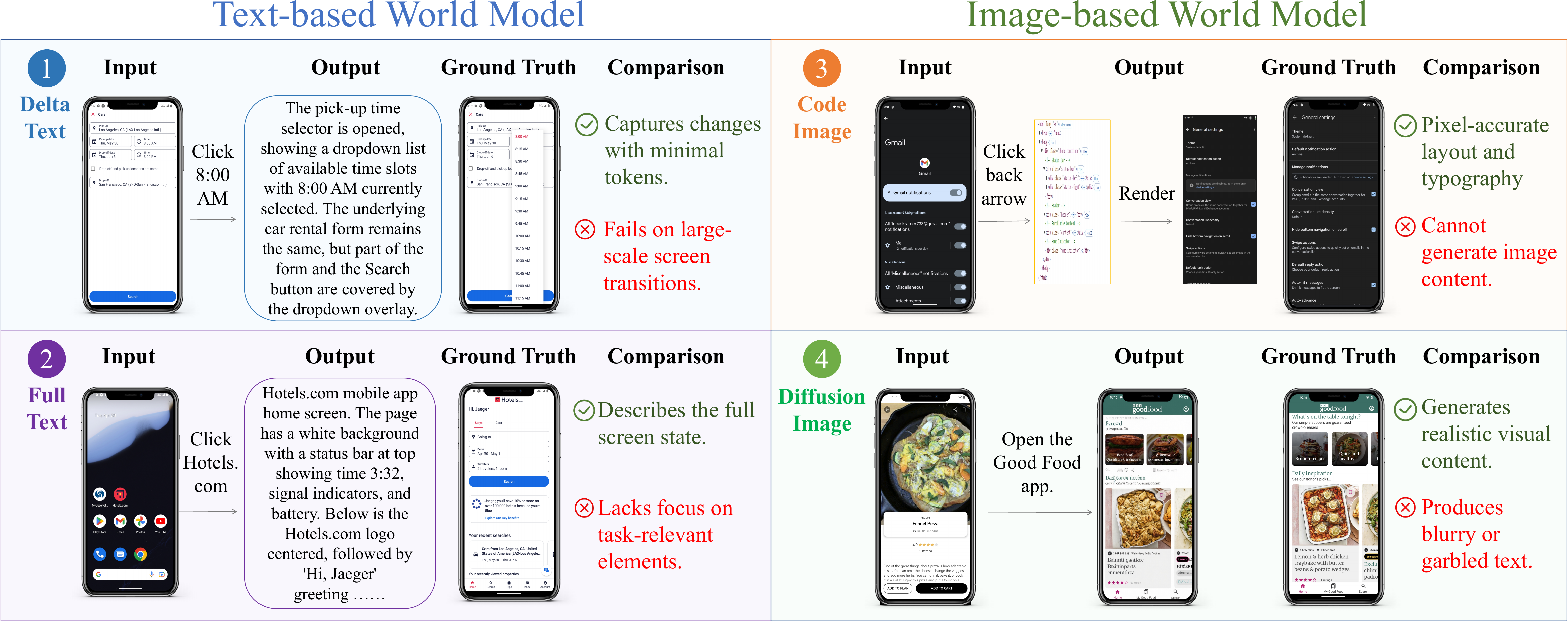}
    \caption{
    Comparison between text-based and image-based world models for GUI state prediction.
    }
    \label{fig:world_model_comparison}
\end{figure}
\begin{figure}[t]
\vspace{-0.3cm}
    \centering
    \adjustbox{margin=0pt 0pt 15mm 0pt}{%
        \includegraphics[width=\linewidth]{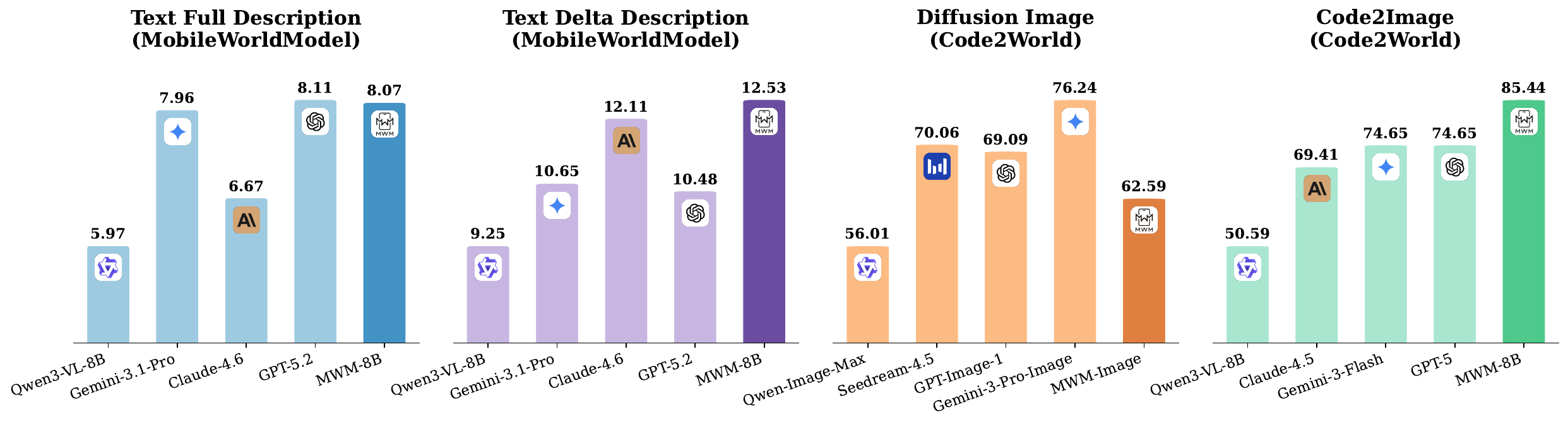}
    }
    \vspace{-0.7cm}
    \caption{
    Comparison of different world-modeling paradigms across four generation settings.
    }
    \vspace{-0.6cm}
    \label{fig:wm_exp_bar}
\end{figure}

To answer \textbf{RQ1}, Section~\ref{sec:rq1} evaluates the four prediction formats from three complementary perspectives: next-state reconstruction quality, offline step-level action selection, and online end-to-end task completion. 
This design allows us to distinguish whether a modality is merely easy to reconstruct, useful for individual action decisions, or actually beneficial for long-horizon mobile task completion. 
Our results show that no single modality dominates across all settings. 
Code2Image provides strong in-distribution reconstruction and effective visual guidance for capable agents, whereas text-based predictions are more robust in online and out-of-distribution environments. 
Diffusion image prediction offers visual context but suffers from text-rendering errors and high inference cost. 
These findings suggest that mobile world models should be selected according to generation reliability, downstream interpretability, deployment distribution, and efficiency, rather than visual fidelity alone.

To answer \textbf{RQ2}, Section~\ref{sec:rq2} studies how mobile world model predictions should be incorporated into agents during test-time decision making. 
We analyze three factors that determine the effectiveness of world model guidance: decision confidence, interaction mode, and sampling budget. 
Our results show that mobile agents are often highly overconfident and low-entropy, which limits the diversity of candidate actions that a posterior world model critic can evaluate. 
Prior perception can bypass this bottleneck by exposing possible consequences before the agent commits to an action, while test-time scaling can partly mitigate the sampling-space limitation when the world-model representation provides distinguishable feedback. 
These results indicate that mobile world models are most effective when used as prior perceptual modules rather than purely posterior critics, and that their test-time benefit is jointly constrained by agent confidence, candidate diversity, and inference budget.

To answer \textbf{RQ3}, Section~\ref{sec:rq3} studies whether agents can acquire new capabilities from mobile world model imagined interactions. 
Motivated by the limitations of test-time guidance, 
we train agents on trajectories collected from the environment simulated by mobile world model and evaluate whether the resulting policy improves beyond the original training setting. 
Our results are mixed but informative: training on imagined trajectories degrades performance on AndroidControl, yet improves success rate on AndroidWorld. 
This discrepancy suggests that imagined trajectories do not necessarily preserve the source data distribution; instead, their value lies in providing transferable interaction experience that can improve generalization when the generated behaviors are useful.

Across these three questions, our study shows that the value of mobile world models depends not only on how accurately they predict future states, but also on how their predictions are represented, consumed, and converted into agent experience. 
Our results provide a unified empirical analysis of four representative prediction formats, reveal how agent confidence, candidate diversity, interaction mode, and inference budget constrain test-time world-model guidance, and further examine the opportunities and risks of using imagined interactions for agent training. 
Visually faithful predictions can be less reliable under distribution shift, posterior guidance can be constrained by overconfident action policies, and imagined trajectories may fail to preserve the source benchmark distribution while still providing useful transferable interaction experience in online environments. 
These findings suggest that mobile world models should be designed as integrated components of mobile agent systems, rather than as standalone next-state predictors.

\section{Constructing Mobile World Models}
\label{sec:mwm}

\subsection{Prediction Formats of Mobile World Models}

As shown in Figure~\ref{fig:world_model_comparison}, existing mobile world models fall into four paradigms along two axes:  text-based approaches (delta and full text) and image-based approaches (code-rendered and diffusion-generated).

\noindent \textbf{Delta Text Model.} 
A delta text model takes the current page and action as input and predicts only how the next page changes relative to the current page, i.e., $(s_t, a_t) \mapsto \Delta y_{t+1}$. 
When the next page differs only slightly from the current one, this format can complete the prediction with only a short textual description, reducing redundancy and potential prediction errors.

\noindent \textbf{Full Text Model.} When the next page is completely different from the previous one, a delta description cannot express this difference. In this case, we expect the model to take $(s_t, a_t)$ as input and predict the full textual description of the next page, i.e., $(s_t, a_t) \mapsto y_{t+1}$, rather than only the change, while it still remains in a purely textual modality. The full description treats all elements equally, making it difficult to highlight task-relevant details among the verbose output.

\noindent \textbf{Code2Image Model.}
This model takes $(s_t, a_t)$ as input and predicts renderable code for the next page, i.e., $(s_t, a_t) \mapsto c_{t+1}$ with $\hat{s}_{t+1} = \mathrm{Render}(c_{t+1})$. 
It complements pure text prediction by preserving textual content, widget shapes, relative positions, and page layout. 
However, much of the generated code is consumed by structural markup, and renderable code still cannot synthesize photographic content, limiting its usefulness for tasks requiring visual judgment over image regions.

\noindent \textbf{Diffusion Image Model.} This model takes the current page and action as input and directly predicts the next screenshot, i.e., $(s_t, a_t) \mapsto \hat{s}_{t+1}$. It compensates for the inability of Code2Image models to insert images. However, because such models are pretrained on natural images and rely on a pixel-level denoising process, they cannot render clear text, are slow to generate, and depend on the input image. Despite the aforementioned defects, we still want to investigate whether it can guide agents with supplementary visual information to enhance performance under limited textual data.

\subsection{Transition-Level Data Construction}
\label{appendix:data_processing_details}

Existing GUI datasets are usually organized as linear trajectories, whereas our world models are trained on action-conditioned state transitions. 
We therefore convert each trajectory into transition triples $(s_t, a_t, s_{t+1})$ and apply graph-level deduplication followed by two quality filtering stages.
\begin{wrapfigure}{r}{0.35\textwidth}
    \vspace{-0.3cm}
    \includegraphics[width=0.35\textwidth]{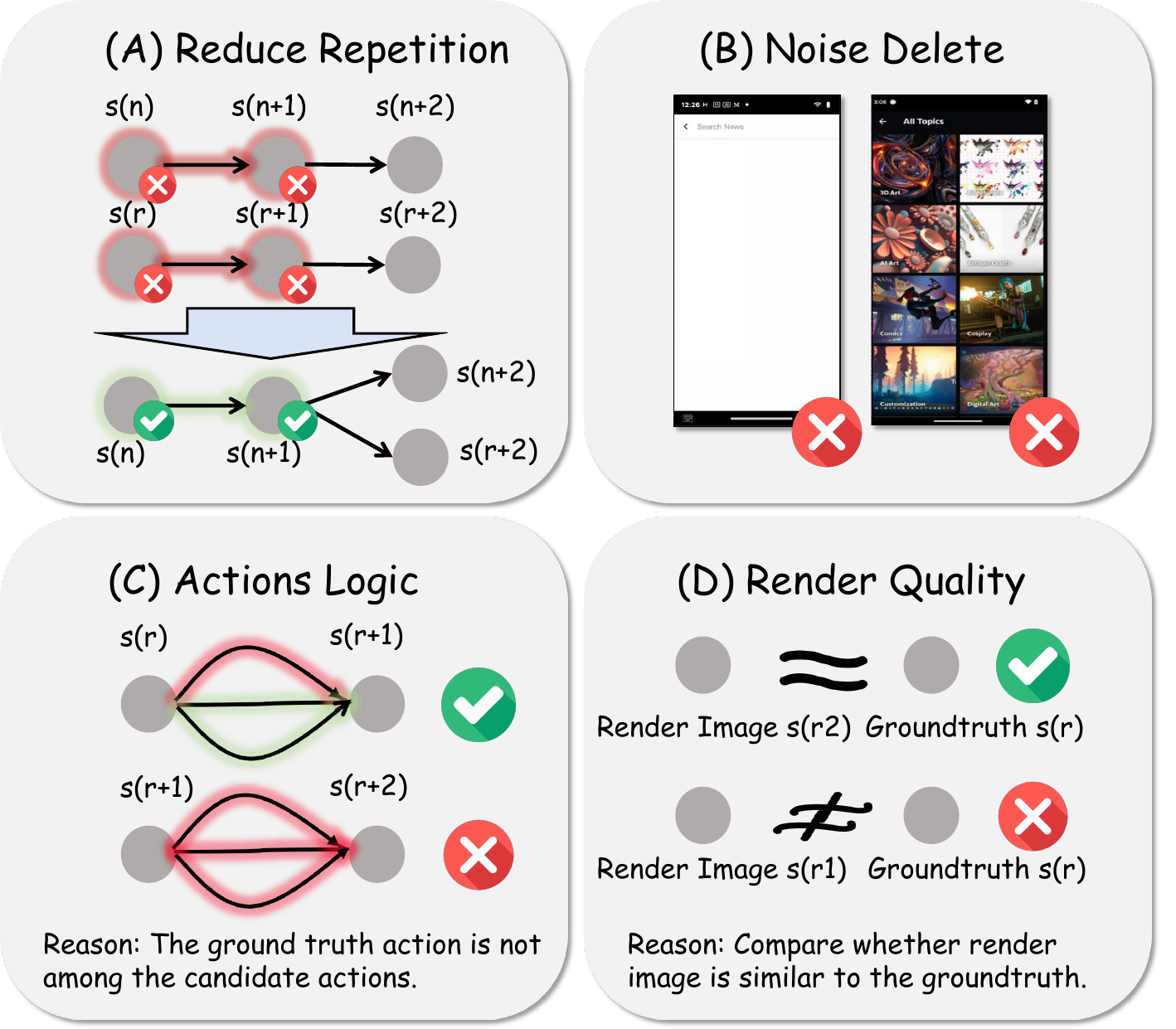}
    \vspace{-0.3cm}
    \caption{Four data filtering methods during our data construction.}
    \vspace{-0.3cm}
    \label{fig:datafilter}
\end{wrapfigure}

\noindent \textbf{Graph-level transition deduplication.}
To reduce repeated state transitions, we merge transition triples with similar start states, similar next states, and the same action type. 
Following the node-merging strategy in Mobile3M~\citep{wu2024mobilevlm}, for two triples $\tau_n=(s_n,a_n,s_{n+1})$ and $\tau_r=(s_r,a_r,s_{r+1})$, we treat them as duplicates if
$D(s_n,s_r)>0.95$, $D(s_{n+1},s_{r+1})>0.95$, and $\mathrm{type}(a_n)=\mathrm{type}(a_r)$, where $D(\cdot,\cdot)$ denotes DINOv2 similarity and $\mathrm{type}(a)$ denotes the action type.
This step removes visually redundant transitions while preserving different interaction semantics.

\noindent \textbf{Rendered-state fidelity filtering.}
For render-code annotations, we verify whether the generated code can faithfully reproduce the target next state. 
Let $c_{t+1}$ denote the annotated render code and $\hat{s}_{t+1}=\mathrm{Render}(c_{t+1})$ denote the screenshot rendered from this code at the same resolution as the ground-truth image $s_{t+1}$.
We discard a sample if $D(\hat{s}_{t+1},s_{t+1})<\gamma$, where $\gamma$ is a preset similarity threshold. 
This filtering is necessary because HTML-based rendering cannot always reproduce photographic content, native widgets, or dynamic visual effects, and low-fidelity rendered states may mislead downstream agents.

\noindent \textbf{Action-logic filtering.}
We further filter transitions whose annotated action is weakly supported by the observed state change. 
Inspired by Code2World~\citep{zheng2026code2world}, we use Gemini-3-Flash as a verifier and provide the current screenshot $s_t$ and ground-truth next screenshot $s_{t+1}$ as input. 
The verifier samples five plausible actions $\{\hat{a}^{(k)}\}_{k=1}^{5}$, and we keep the transition only if the annotated action appears among these candidates, i.e., $a_t \in \{\hat{a}^{(k)}\}_{k=1}^{5}$. 
This step removes noisy annotations where the action description, coordinates, or observed state transition are inconsistent.

\noindent \textbf{Page-noise removal.}
In addition to the above filters, we remove highly homogeneous blank search pages that appear frequently in search trajectories. 
These pages introduce strong visual homogeneity and can bias both diffusion-based generation and HTML rendering toward black-and-white layouts. 
Removing them improves the diversity and reliability of the final training corpus.



\subsection{Data Statistics} \label{DataStatistics}
We construct the GUI training corpus from multiple mobile datasets to cover different annotation granularities, interface distributions, and trajectory horizons. AITW is drawn from the Google Apps split of Android in the Wild~\citep{rawles2023androidinthewild}. AMEX~\citep{chai2025amex} provides richer page-level and element-level annotations. 
\begin{wraptable}{r}{0.45\textwidth}
    \vspace{-0.6cm}
    \caption{Two-stage data filtering statistics.}
    \centering
    \footnotesize
    \resizebox{\linewidth}{!}{
    \begin{tabular}{lrrrr}
        \toprule
        Dataset & \shortstack{Trajectory\\Steps} & \shortstack{Graph\\Dedup} & \shortstack{Quality\\Dedup} & \shortstack{Training\\Steps} \\
        \midrule
        AITW & 18,995 & 12,657 & 10,048 & 10,048 \\
        AMEX & 38,630 & 31,206 & 29,494 & 29,494 \\
        AITZ & 14,218 & 9,665 & 9,203 & 9,203 \\
        Android\_control & 74,722 & 33,972 & 31,476 & 31,476 \\
        GUI-Odyssey & 118,506 & 34,710 & 22,394 & - \\
        \midrule
        Total & 263,071 & 122,210 & 102,615 & 80,221\\
        \bottomrule
    \end{tabular}
    }
    \vspace{-0.4cm}
\label{tab:data_filtering_pipeline}
\end{wraptable}
AITZ~\citep{zhang2024android} contributes CoaT-format action description and complete element lists, AndroidControl~\citep{li2024effects} introduces more challenging out-of-distribution settings, and GUI-Odyssey~\citep{lu2025guiodyssey} offers longer and more complex cross-app trajectories. 
As shown in Table~\ref{tab:data_filtering_pipeline}, we use Claude4.6-Opus and Gemini3-Pro (they have different strengths) to relabel all retained samples with delta descriptions, full descriptions, and renderable HTML code.
The largest reduction occurs during page-noise removal, especially for AndroidControl and GUI-Odyssey, which contain more complex pages. GUI-Odyssey is annotated but excluded from the training set, serving instead as the OOD benchmark in later experiments.

\section{What Should Mobile World Models Predict?}
\label{sec:rq1}


To determine which modality of world model output most effectively guides GUI agents, we first validate the reconstruction quality of our world model. We then conduct offline step-level evaluation to isolate the effect of each modality on individual action decisions under controlled conditions, followed by online end-to-end evaluation to verify whether these gains translate into real task completion.

\subsection{How Well Do Mobile World Models Predict? Next-state Reconstruction Quality}

\noindent \textbf{Settings.} Except for the diffusion-based models, we use Qwen3-4B/8B~\citep{yang2025qwen3} as the base models and train them on our relabeled data. For diffusion-based world models, we use Qwen-Image-Edit-20B~\citep{wu2025qwen} and StableDiffusion-3.5-Large-8B~\citep{blattmann2023stable}\footnote{This paper uses \href{https://huggingface.co/stabilityai/stable-diffusion-3.5-large}{stable-diffusion-3.5-large}, which is licensed under the \href{https://stability.ai/community-license-agreement}{Stability Community License}. The authors confirm that the use of this model is strictly for academic research purposes and involves no commercial activities.} as baselines and train them on the same annotated data. For render-code distillation, Appendix~\ref{sec:wm_token_stats} and Table~\ref{tab:wm_token_stats} report the token statistics used to balance annotation quality and output length. Text-based world models are evaluated on MobileWorldBench~\citep{li2025mobileworldbench}. We follow its original experimental setting and use three VLM-as-a-Judge metrics, namely accuracy, completeness, and relevance, with Gemini-3-Pro as the judge and report the average of three evaluation runs. The image-based models are evaluated on Code2World~\citep{zheng2026code2world}. The functional-logic metrics include $S_{ad}$ and $S_{id}$, which measure action adherence and action identifiability, respectively. The visual-quality metrics include $S_{ele}$ and $S_{lay}$, which evaluate element alignment and layout integrity, respectively. The formulas of the functional-logic metrics are provided in Appendix~\ref{appendix:eval_metrics}, while the visual-quality metrics are computed by DINOv2~\citep{oquab2023dinov2} and SigLIP~\citep{tschannen2025siglip}.
\begin{table*}[htbp]
    \vspace{-0.3cm}
    \caption{Results of text-based world models on MobileWorldBench. Each dimension is rated from 1 to 5 and was evaluated by Gemini-3.1-Pro.}
    \centering
    \footnotesize
    \resizebox{0.80\textwidth}{!}{
    \begin{tabular}{l cccc cccc}
        \toprule
        \multirow{2}{*}{\textbf{Model}} &
        \multicolumn{4}{c}{\textbf{Full description}} &
        \multicolumn{4}{c}{\textbf{Delta description}} \\
        \cmidrule(lr){2-5} \cmidrule(lr){6-9}
        & \textbf{Acc.} & \textbf{Com.} & \textbf{Rele.} & \textbf{Overall}
        & \textbf{Acc.} & \textbf{Com.} & \textbf{Rele.} & \textbf{Overall} \\
        \midrule
        Qwen3-VL-4B & 1.98 & 1.83 & 2.10 & 5.83 & 3.28 & 2.45 & 3.56 & 9.28 \\
        Qwen3-VL-8B & 1.98 & 2.02 & 1.97 & 5.97 & 3.24 & 2.44 & 3.56 & 9.25 \\
        Qwen3-VL-32B & 1.93 & 1.92 & 2.11 & 5.95 & 3.36 & 2.75 & 3.77 & 9.87 \\
        Mobile-8B & - & - & - & - & 4.19 & \secondbest{3.70} & 4.50 & 12.39 \\
        Gemini-3.1-Pro & 2.57 & 2.42 & \secondbest{2.97} & 7.96 & 3.66 & 2.94 & 4.05 & 10.65 \\
        Claude-4.6-Sonnet & 2.19 & 2.08 & 2.40 & 6.67 & 4.14 & 3.49 & 4.48 & 12.11 \\
        GPT-5.2 & 2.62 & 2.48 & \best{3.02} & \secondbest{8.11} & 3.55 & 3.11 & 3.82 & 10.48 \\
        MobileWorldModel-4B & 2.47 & 2.59 & 2.43 & 7.48 & 3.55 & 3.34 & 3.99 & 10.87 \\
        \quad + visual hint & 2.49 & 2.60 & 2.50 & 7.60 & 3.64 & 3.39 & 4.04 & 11.07 \\
        MobileWorldModel-8B & \secondbest{2.66} & \secondbest{2.76} & 2.64 & 8.07 & \secondbest{4.26} & 3.65 & \secondbest{4.62} & \secondbest{12.53} \\
        \quad + visual hint & \best{2.78} & \best{2.89} & 2.70 & \best{8.37} & \best{4.29} & \best{3.77} & \best{4.68} & \best{12.74} \\
        \bottomrule
    \end{tabular}
    }
    \vspace{-0.3cm}
    \label{tab:text_world_model_mobileworldbench}
\end{table*}

\noindent \textbf{Experimental Results.}
Table~\ref{tab:text_world_model_mobileworldbench} shows that delta descriptions are consistently easier to evaluate than full-state descriptions. Generic VLMs obtain only 5.83--5.97 overall scores on full descriptions for the Qwen3-VL family, whereas their delta description scores increase to 9.25--9.87. This gap suggests that localized state-change prediction requires more precise grounding of which UI elements changed, while full descriptions provide richer context for the judge. MobileWorldModel closes part of this gap: MobileWorldModel-8B reaches 8.07 on full descriptions and 12.53 on delta descriptions, outperforming general-purpose models of similar or larger scale. Adding visual hints further improves both description formats, with MobileWorldModel-8B increasing from 8.07 to 8.37 on full descriptions and from 12.53 to 12.74 on delta descriptions. These results indicate that domain-specific mobile GUI world modeling benefits from both specialized training and explicit visual grounding, especially when the evaluation requires element-level localization.
\begin{wraptable}{r}{0.58\textwidth}
    \vspace{-0.6cm}
    \caption{Image-based world model results on Code2World.}
    \centering
    \scriptsize
    \resizebox{\linewidth}{!}{
    \begin{tabular}{l cccc ccc}
        \toprule
        \multirow{2}{*}{\textbf{Model}} &
        \multicolumn{2}{c}{\textbf{Functional Logic}} &
        \multicolumn{4}{c}{\textbf{Visual Quality}} &
        \multirow{2}{*}{\textbf{Overall}} \\
        \cmidrule(lr){2-3} \cmidrule(lr){4-7}
        & \textbf{$S_{ad}$} & \textbf{$S_{id}$} & \textbf{$S_{ele}$} & \textbf{$S_{lay}$} & \textbf{\textit{SigLIP}} & \textbf{\textit{DINOv2}} \\
        \midrule
        \multicolumn{7}{c}{\textit{\textbf{Image generation}}} \\
        \midrule
        Gemini-3-Pro-Image & 92.63 & 83.67 & 68.47 & 63.67 & \secondbest{84.89} & 64.11 & \secondbest{76.24} \\
        GPT-Image-1 & 89.59 & 71.40 & 58.78 & 56.22 & 77.58 & 61.00 & 69.09\\
        Doubao-Seedream-4.5 & 85.36 & 86.15 & 59.08 & 55.82 & 81.76 & 52.19 & 70.06 \\
        Qwen-Image-Edit-Max & 57.55 & 54.12 & 54.05 & 46.33 & 70.39 & 53.61 & 56.01\\
        Janus-Pro-7B & 58.10 & 53.68 & 55.22 & 45.90 & 79.85 & 54.30 & 57.84\\
        StableDiffusion-3.5-Large-8B & 54.65 & 70.21 & \bestred{23.06} & \bestred{27.19} & 64.21 & 54.64 & 48.99\\
        MobileWorldModel-Image (Ours) & 79.15 & 80.95 & \secondbestred{32.17} & \secondbestred{34.13} & 82.19 & \secondbest{66.94} & 62.58\\
        \midrule
        \multicolumn{7}{c}{\textit{\textbf{Code generation}}} \\
        \midrule
        GPT-5 & 94.02 & \secondbest{90.15} & 74.13 & 69.78 & 78.07 & 41.74 & 74.64\\
        Gemini-3-Flash & 92.65 & 84.08 & \secondbest{74.52} & 69.74 & 81.17 & 45.72 & 74.65\\
        Claude-4.5-Sonnet & 89.60 & 86.12 & 65.80 & 62.25 & 75.97 & 36.72 & 69.41\\
        JanusCoderV-7B & 57.12 & 56.05 & 30.18 & 31.09 & 59.89 & 21.73 & 42.68\\
        Qwen3-VL-8B & 59.20 & 65.80 & 43.10 & 42.70 & 63.88 & 28.86 & 50.59\\
        Qwen2.5-VL-72B & 73.34 & 70.12 & 47.15 & 48.26 & 68.45 & 30.12 & 56.24\\
        InternVL3-78B & 72.41 & 67.35 & 45.06 & 46.73 & 62.64 & 25.42 & 53.26\\
        GLM-4.6V-106B & 91.62 & 74.23 & 61.95 & 58.74 & 67.26 & 27.42 & 63.53\\
        Code2World-8B & \secondbest{94.28} & 88.64 & 71.35 & \secondbest{70.32} & 79.44 & 49.18 & 75.53\\
        MobileWorldModel-8B (Ours) & \best{96.51} & \best{94.60} & \best{79.14} & \best{81.43} & \best{87.11} & \best{73.87} & \best{85.44}\\
        \bottomrule
    \end{tabular}
    }
    \vspace{-0.6cm}
\label{tab:image_world_model_code2worldbench}
\end{wraptable}
For image-based world models, Figure~\ref{fig:wm_exp_bar} and Table~\ref{tab:image_world_model_code2worldbench} show that MobileWorldModel-8B achieves the best results across all code-generation metrics. 
Its gains on functional-logic metrics support the effectiveness of action-consistency filtering. 
In contrast, diffusion-based models obtain much lower text-sensitive scores than visual-similarity scores ($32.17, 34.13$ vs. $82.19, 66.94$), indicating clear text-rendering weaknesses despite their ability to provide additional visual context.


\vspace{-0.2cm}
\subsection{How Do Modalities Affect Action Selection? Offline Step-Level Evaluation}


\noindent \textbf{Settings.} \label{offlinesettings}
We report results on app-unseen, task-unseen, and in-distribution splits, which respectively evaluate generalization to unseen applications, unseen task instructions, and familiar app-task distributions. We choose MAI-UI-8B~\citep{zhou2025mai} as an SFT GUI agent, Qwen3-VL-8B/32B/238B-A22B to study parameter scaling, Gemini 3-Flash as a weaker closed-source model, and GPT-5.4 as a strong agent whose performance is close to the annotation model. To quantify how much world models improve downstream agent performance, BO3 serves as a lower bound for self-reflection, while PASS@3 approximates the upper bound under three sampled actions. Percentage measures the fraction of sampled actions for which the world model is effectively involved, Tokens compares the efficiency of VLM-based methods, and Time measures the efficiency cost of diffusion-based world models.

\begin{table*}[t]
    \vspace{-0.5cm}
    \caption{Offline task navigation results across base GUI agents and world-model interaction methods. App., Task., and IDD denote app-unseen, task-unseen, and in-distribution settings, respectively.}
    \centering
    \scriptsize
    \resizebox{0.99\textwidth}{!}{
    \begin{tabular}{ll | cccc cccc | c | cc}
        \toprule
        \multirow{2}{*}{\textbf{Base Model}} &
        \multirow{2}{*}{\textbf{Settings}} &
        \multicolumn{4}{c}{\textbf{Type Acc.}} &
        \multicolumn{4}{c}{\textbf{Step Acc.}} &
        \multirow{2}{*}{\textbf{Percentage}} &
        \multicolumn{2}{c}{\textbf{Efficiency}} \\
        \cmidrule(lr){3-6} \cmidrule(lr){7-10} \cmidrule(lr){12-13}
        & & \textbf{App.} & \textbf{Task.} & \textbf{IDD} & \textbf{Overall}
        & \textbf{App.} & \textbf{Task.} & \textbf{IDD} & \textbf{Overall}
        & & \textbf{Tokens} & \textbf{Time} \\
        \midrule
        \multirow{7}{*}{MAI-UI-8B}
        & Base & 78.46 & 77.87 & 81.05 & 79.39 & 63.52 & 62.37 & 66.73 & 64.43 & 56.56 & 19320.65 & 41.93 \\
        & BO3 & \secondbestred{77.68} & \bestred{76.44} & \secondbestred{80.16} & \bestred{78.03} & \bestred{59.62} & \bestred{60.67} & \bestred{65.31} & \bestred{62.69} & 56.56 & 59603.47 & 143.52 \\
        & PASS@3 & 81.23 & 80.51 & 83.78 & 82.05 & 65.88 & 64.78 & 70.20 & 67.31 & 56.56 & 58038.11 & 123.46 \\
        & Delta Text & 79.04 & 78.39 & 81.54 & 79.83 & \secondbestred{63.35} & \secondbestred{62.31} & 67.63 & 64.77 & 56.56 & 58368.72 & 144.09 \\
        & Full Text & \best{79.59} & \best{79.02} & \best{82.19} & \best{80.47} & \best{64.08} & \best{63.08} & \secondbest{68.37} & \best{65.53} & 56.56 & 62037.94 & 143.74 \\
        & Code2Image & \secondbest{79.23} & \secondbest{78.71} & \secondbest{82.04} & \secondbest{80.25} & \secondbest{63.92} & \secondbest{62.98} & \best{68.41} & \secondbest{65.51} & 56.56 & 74465.88 & 241.86 \\
        & Diffusion2Image & \bestred{74.86} & \bestred{73.95} & \bestred{77.48} & \bestred{75.62} & \bestred{59.11} & \bestred{57.84} & \bestred{63.52} & \bestred{60.26} & 56.56 & -- & \bestred{78846.64} \\
        \midrule
        \multirow{7}{*}{Qwen3-VL-8B}
        & Base & 80.28 & 79.22 & 82.93 & 81.02 & 66.31 & 64.65 & 68.81 & 66.61 & 44.86 & 21821.74 & 40.71 \\
        & BO3 & 80.42 & 79.41 & 82.97 & 81.15 & 66.39 & \secondbestred{64.52} & \secondbestred{68.04} & \secondbestred{66.43} & 44.86 & 66589.19 & 135.87 \\
        & PASS@3 & 83.36 & 82.39 & 84.62 & 83.46 & 70.07 & 68.20 & 71.10 & 69.55 & 44.86 & 65428.87 & 120.78 \\
        & Delta Text & \best{81.42} & \best{80.60} & \secondbestred{82.83} & \best{81.66} & \secondbest{67.96} & \secondbest{66.19} & \secondbestred{68.51} & 67.27 & 44.86 & 66784.92 & 137.51 \\
        & Full Text & 81.33 & \secondbest{80.57} & \secondbestred{82.83} & \secondbest{81.62} & 67.79 & 66.11 & \best{69.02} & \secondbest{67.42} & 44.86 & 69232.45 & 140.87 \\
        & Code2Image & \secondbest{81.40} & 80.51 & \best{82.99} & \best{81.66} & \best{68.11} & \best{66.40} & \best{69.02} & \best{67.59} & 44.86 & 86437.64 & 240.19 \\
        & Diffusion2Image & \bestred{76.41} & \bestred{75.08} & \bestred{78.56} & \bestred{76.72} & \bestred{62.11} & \bestred{60.42} & \bestred{64.37} & \bestred{62.05} & 44.86 & -- & \bestred{71135.14} \\
        \midrule
        \multirow{7}{*}{Qwen3-VL-32B}
        & Base & 82.27 & 81.43 & 83.51 & 82.43 & 67.97 & 66.53 & 69.22 & 67.83 & 57.37 & 21898.75 & 74.96 \\
        & BO3 & 82.34 & 81.57 & \secondbestred{83.46} & \secondbestred{82.38} & 68.11 & \secondbestred{66.41} & \secondbestred{69.08} & 67.95 & 57.37 & 61912.63 & 255.08 \\
        & PASS@3 & 84.61 & 83.51 & 85.29 & 84.36 & 70.13 & 68.50 & 71.40 & 69.88 & 57.37 & 65663.73 & 234.46 \\
        & Delta Text & \best{82.81} & \best{81.65} & \secondbest{83.64} & \best{82.61} & \secondbestred{67.84} & \secondbestred{66.46} & 69.61 & 67.97 & 57.37 & 80303.21 & 402.53 \\
        & Full Text & 82.29 & \secondbestred{81.26} & \best{83.75} & \secondbest{82.44} & \secondbestred{67.93} & \best{66.71} & \secondbest{69.87} & \secondbest{68.13} & 57.37 & 81614.67 & 433.72 \\
        & Code2Image & \secondbest{82.57} & \secondbestred{81.36} & \secondbestred{83.48} & \secondbestred{82.39} & \secondbestred{67.91} & \secondbest{66.63} & \best{69.88} & \best{68.16} & 57.37 & 87787.07 & 442.63 \\
        & Diffusion2Image & \secondbestred{81.46} & \bestred{80.34} & \secondbestred{83.36} & \secondbestred{81.79} & \secondbestred{67.12} & \secondbestred{65.70} & \secondbestred{68.65} & \secondbestred{67.57} & 57.37 & -- & \bestred{76884.45} \\
        \midrule
        \multirow{7}{*}{Qwen3-VL-238B-A22B}
        & Base & 85.65 & 84.31 & 86.67 & 85.47 & 70.09 & 68.64 & 71.87 & 70.26 & 44.58 & 22282.30 & 84.63 \\
        & BO3 & \secondbestred{85.58} & 84.44 & \secondbestred{86.52} & \secondbestred{85.39} & 70.22 & \secondbestred{68.51} & \secondbestred{71.74} & 70.33 & 44.58 & 72301.86 & 234.49 \\
        & PASS@3 & 86.79 & 85.58 & 88.67 & 86.89 & 72.13 & 70.41 & 73.62 & 71.92 & 44.58 & 66865.45 & 230.26 \\
        & Delta Text & \secondbestred{85.54} & \secondbest{84.34} & \secondbest{87.21} & \secondbest{85.72} & \best{70.54} & \best{68.84} & \secondbest{72.12} & \secondbest{70.41} & 44.58 & 83273.24 & 234.89 \\
        & Full Text & \secondbestred{85.23} & \secondbestred{84.11} & 86.84 & \secondbestred{85.45} & 70.34 & 68.70 & 71.91 & \secondbestred{70.23} & 44.58 & 94254.32 & 258.59 \\
        & Code2Image & \secondbestred{85.48} & \best{84.40} & \best{87.30} & \best{85.82} & \secondbest{70.36} & \secondbest{68.79} & \best{72.34} & \best{70.48} & 44.58 & 104971.49 & 304.59 \\
        & Diffusion2Image & \bestred{82.41} & \bestred{80.93} & \bestred{83.12} & \bestred{82.05} & \bestred{66.84} & \bestred{65.27} & \bestred{68.11} & \bestred{66.73} & 44.58 & -- & \bestred{71952.54} \\
        \midrule
        \multirow{7}{*}{\shortstack{Gemini 3-Flash\\+ UI-Ins-7B}}
        & Base & 76.51 & 76.57 & 77.91 & 77.21 & 48.00 & 48.07 & 45.69 & 46.39 & 90.18 & 16049.36 & 39.44 \\
        & BO3 & 77.63 & \bestred{74.82} & \bestred{75.97} & \bestred{75.98} & 50.31 & 49.69 & 50.81 & 50.37 & 90.18 & 47629.65 & 149.37 \\
        & PASS@3 & 82.30 & 82.44 & 83.13 & 82.74 & 71.67 & 71.50 & 71.81 & 71.63 & 90.18 & 45353.52 & 137.61 \\
        & Delta Text & \bestred{63.49} & \bestred{64.18} & \bestred{64.21} & \bestred{64.15} & 53.05 & 53.54 & 52.05 & 52.84 & 90.18 & 49613.12 & 239.44 \\
        & Full Text & \bestred{65.63} & \bestred{66.32} & \bestred{67.25} & \bestred{66.75} & \secondbest{54.23} & \secondbest{54.71} & 55.39 & 55.06 & 90.18 & 54631.47 & 223.15 \\
        & Code2Image & \best{81.09} & \best{81.08} & \best{82.34} & \best{81.68} & \best{71.67} & \best{71.50} & \best{71.81} & \best{71.63} & 90.18 & 64344.12 & 240.07 \\
        & Diffusion2Image &  \secondbest{76.77} & \secondbest{77.25} & \secondbest{79.7} & \secondbest{78.15} & \secondbest{54.45} & 53.81 & \secondbest{56.61} & \secondbest{55.15} & 90.18 & -- & \bestred{82586.16} \\
        \midrule
        \multirow{7}{*}{GPT-5.4 + UI-Ins-7B}
        & Base & 86.97 & 87.55 & 89.46 & 88.46 & 69.20 & 69.42 & 71.63 & 70.38 & 85.27 & 11493.12 & 41.20 \\
        & BO3 & 89.73 & 88.69 & 90.06 & 89.86 & 69.77 & 70.31 & 71.90 & 70.94 & 85.27 & 36192.04 & 149.36 \\
        & PASS@3 & 97.87 & 97.88 & 98.09 & 97.99 & 74.88 & 75.13 & 76.18 & 75.22 & 85.27 & 35035.67 & 123.57 \\
        & Delta Text & 87.91 & 88.61 & 89.47 & 89.03 & \secondbestred{68.37} & \secondbestred{69.07} & \bestred{69.94} & \secondbestred{69.49} & 85.27 & 37768.21 & 152.19 \\
        & Full Text & \secondbest{87.92} & 88.59 & \secondbest{91.60} & \secondbest{90.05} & \secondbest{70.70} & \secondbest{71.15} & \secondbest{73.35} & \secondbest{72.21} & 85.27 & 42384.51 & 161.56 \\
        & Code2Image & \best{92.43} & \best{92.76} & \best{93.73} & \best{93.28} & \best{74.21} & \best{74.57} & \best{74.77} & \best{74.71} & 85.27 & 54835.59 & 242.56 \\
        & Diffusion2Image & 87.08 & \secondbestred{87.41} & \secondbestred{89.32} & 88.51 & 69.34 & \secondbestred{69.27} & \secondbestred{71.48} & 70.52 & 85.27 & -- & \bestred{80938.29} \\
        \bottomrule
    \end{tabular}
    }
    \vspace{-0.6cm}
    \label{tab:offline_task_navigation}
\end{table*}

\noindent \textbf{Experimental Results.} Table~\ref{tab:offline_task_navigation} shows that simple self-reflection is not sufficient without additional inputs. We observe a diminishing performance margin between BO3 and PASS@3 evaluations corresponding to the increase in model parameters within the Qwen series. In contrast, closed-source agents show larger gaps, especially Gemini 3-Flash, where PASS@3 exceeds BO3 by $21.26$ in overall Step Acc. Moreover, BO3 often fails to improve over the base policy, indicating that self-reflection alone cannot provide reliable corrective signals. With world-model feedback, several agents approach PASS@3 upper bound. For example, Code2Image improves Gemini 3-Flash from $46.39$ to $71.63$ in overall Step Acc., matching PASS@3 and improving over the base model by $25.24$. \textbf{\textit{This indicates that small-scale models are nearly incapable of performing self-reflection autonomously.}}

\noindent \textbf{More Click Actions Repair By Guidance.} Gemini 3-Flash exhibits a mismatch between Type Acc. and Step Acc. under text-based judging. The text world-model judge tends to assign high scores to \texttt{wait}, which hurts action-type prediction, while the Step Acc. improvement mainly comes from repairing click coordinates. This phenomenon is common in error repair; the original erroneous action may switch from one error type to another. Image-based Code2Image guidance avoids this issue and improves both metrics simultaneously, because their outputs are easier to distinguish.

\noindent \textbf{Text Value Analysis.} The main failure mode of image-based guidance on smaller agents is text recognition. In judge models outputs, we frequently observe abnormal or unreadable descriptions for diffusion images, on Qwen3-VL-8B, diffusion image guidance reduces overall Step Acc. from $66.61$ to $62.05$, and on Gemini 3-Flash, text-based world-model judging reduces overall Type Acc. from $77.21$ to $64.15$ for delta text. Closed-source agents can better extract useful information from rendered images, which explains the gains from Code2Image on Gemini 3-Flash and GPT-5.4. Nevertheless, diffusion-based image generation remains inefficient because the 40-step denoising process incurs a much higher Time cost than VLMs. \textbf{\textit{Therefore, the additional image modeling from diffusion models is insufficient to compensate for shortcomings in text processing and rendering efficiency.}}

\subsection{How Do Modalities Affect Task Completion? Online End-to-End Evaluation}
\noindent \textbf{Settings.}
We evaluate world-model guidance on AndroidWorld~\citep{rawles2024androidworld} using the M3A framework and Mobile-Agent-v3.5~\citep{xu2026mobile} (MA3.5). Each agent samples $k=3$ candidate actions; the world model predicts the next states; and Gemini 3-Flash scores the candidates and selects the highest to execute. 

\noindent \textbf{Benchmark and Metrics.} \label{androidworldsettings}
We employ the official AndroidWorld benchmark, which categorizes unseen application tasks into three distinct difficulty levels: (1) Easy and Medium tasks serve to evaluate the agent's baseline instruction-following and local navigation capabilities. (2) Hard tasks represent a more challenging setting, requiring multi-step reasoning, cross-app navigation, and robustness against deep UI hierarchies. We report the exact Success Rate (SR) determined by programmatic state verification, alongside execution efficiency metrics including Mean Steps, Tokens per Task, and Execution Time per Task.

\begin{table*}[htbp]
    \vspace{-0.3cm}
    \centering
    \caption{Full AndroidWorld results with M3A agents under different world-model feedback formats. Success rate (SR) is reported for easy, medium, hard, and overall task groups, together with tokens per task, mean interaction steps, and run time.}
    \footnotesize
    \resizebox{\textwidth}{!}{
    \begin{tabular}{l c c c c c c c}
        \toprule
        \textbf{Model / WM} & \textbf{Easy SR} & \textbf{Medium SR} & \textbf{Hard SR} & \textbf{Overall SR} & \textbf{Tokens/Task} & \textbf{Mean Steps} & \textbf{Run Time} \\
        \midrule
        Qwen3-VL-8B (M3A) & 54.64 & 27.77 & 16.66 & 40.08 & 1698.13 & 13.69 & 144.56 \\
        + Delta Text & \best{59.01} & \best{34.72} & \secondbest{17.10} & \best{44.61} & 54461.79 & 13.02 & 439.93 \\
        + Full Text & \bestred{50.82} & \best{33.33} & \bestred{11.84} & \bestred{39.01} & 58548.32 & 13.87 & 1469.09 \\
        + Code2Image & \secondbest{54.92} & \best{31.94} & \bestred{13.16} & \secondbest{40.95} & 214360.66 & 13.44 & 2600.86 \\
        \midrule
        Qwen3-VL-32B (M3A) & 62.57 & 44.90 & 24.56 & 50.86 & 1907.01 & 13.78 & 553.32 \\
        + Delta Text & \secondbest{64.75} & \best{50.69} & \bestred{21.05} & \secondbest{53.23} & 47497.21 & 13.87 & 958.14 \\
        + Full Text & \secondbest{63.11} & \best{50.00} & \bestred{21.05} & \secondbest{52.16} & 51497.53 & 14.10 & 1095.01 \\
        + Code2Image & \secondbest{63.11} & \secondbestred{43.06} & \bestred{18.42} & \secondbestred{49.57} & 209073.95 & 13.89 & 2490.72 \\
        \midrule
        Gemini-3-Flash (M3A) & 62.29 & 44.44 & 24.56 & 50.57 & 1238.86 & 11.05 & 290.61 \\
        + Delta Text & \best{81.15} & \best{58.33} & \best{35.53} & \best{66.59} & 56767.73 & 11.69 & 980.39 \\
        + Full Text & \best{82.79} & \best{61.81} & \best{31.58} & \best{67.89} & 61207.18 & 12.17 & 1042.40 \\
        + Code2Image & \best{72.13} & 44.44 & \bestred{10.53} & \secondbest{53.44} & 217123.29 & 12.72 & 2714.06 \\
        \midrule
        GPT-5.4 (M3A) & 72.68 & 41.67 & 29.82 & 56.03 & 2178.89 & 11.44 & 220.60 \\
        + Delta Text & \best{76.23} & \best{59.72} & \secondbest{31.58} & \best{63.79} & 48557.00 & 12.46 & 1017.37 \\
        + Full Text & \secondbestred{72.13} & \best{54.17} & \best{34.21} & \best{60.34} & 56754.34 & 12.61 & 1353.21 \\
        + Code2Image & \bestred{69.67} & \best{56.94} & \bestred{26.32} & \secondbest{58.62} & 224245.86 & 12.31 & 3107.67 \\
        \bottomrule
    \end{tabular}
    }
    \vspace{-0.3cm}
    \label{tab:androidworld_full_results}
\end{table*}

\begin{wrapfigure}{r}{0.4\textwidth}
    \vspace{-0.4cm}
    \centering
    \includegraphics[width=\linewidth]{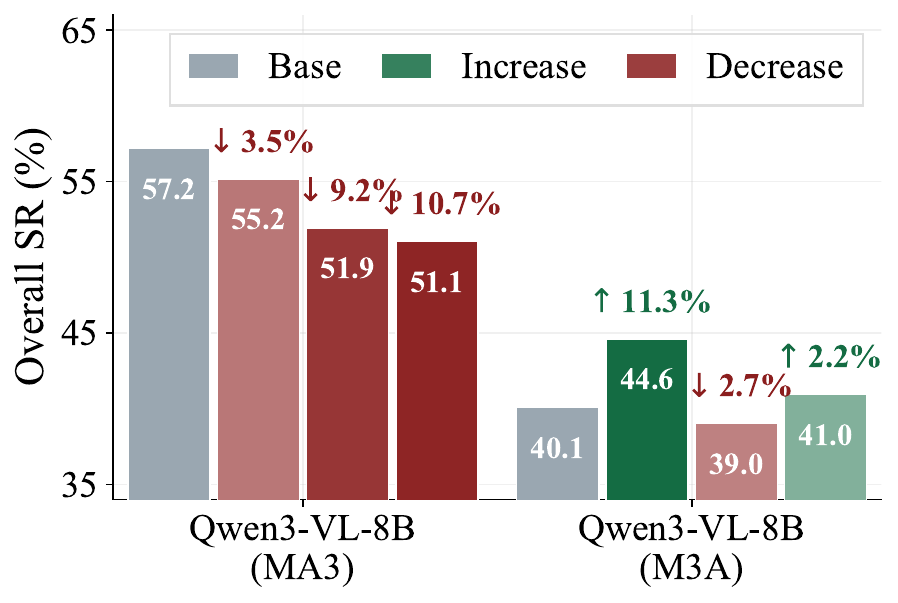}
    \vspace{-0.5cm}
    \caption{Overall SR on AndroidWorld under two agent frameworks.}
    \vspace{-0.2cm}
\label{fig:androidworld_qwen8b_framework_sr}
\end{wrapfigure}
\begin{figure}[t]
\vspace{-0.4cm}
    \centering
    \includegraphics[width=0.9\linewidth]{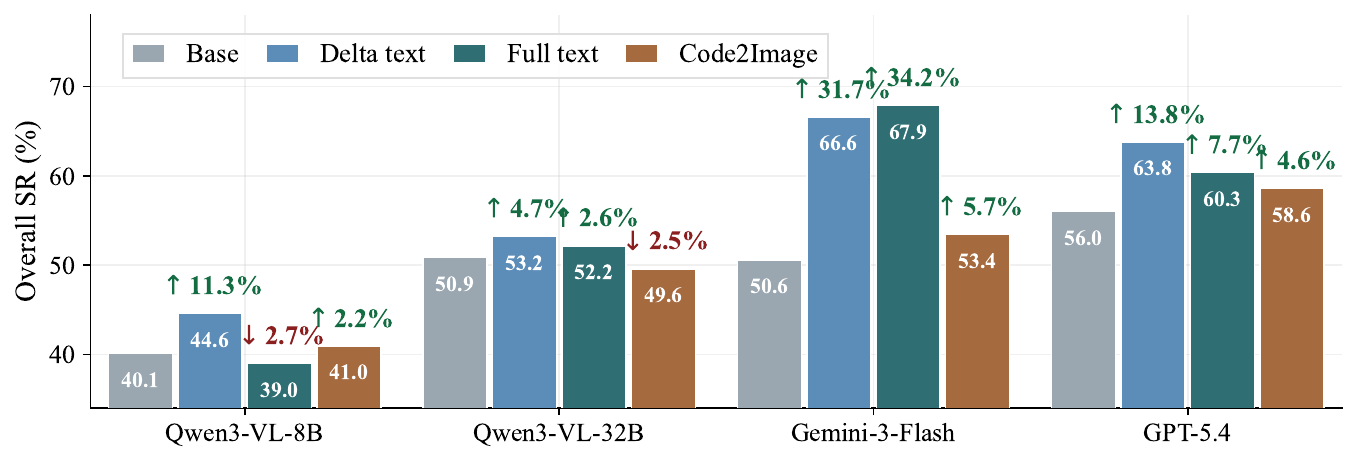}
    \vspace{-0.4cm}
    \caption{Overall success rate on AndroidWorld with M3A agents. Bars report overall SR, and labels above world-model variants show the relative change over the corresponding base agent.}
    \label{fig:androidworld_overall_sr}
    \vspace{-0.4cm}
\end{figure}

\noindent \textbf{Experimental Results.}
Figure~\ref{fig:androidworld_overall_sr} shows that a purely reactive M3A policy remains insufficient for many medium and long-horizon AndroidWorld tasks, where agents must verify intermediate UI states before acting. Text-level world-model feedback supplies this lookahead signal and consistently improves strong foundation models: delta-text feedback raises Gemini 3-Flash overall SR from $50.6$ to $66.6$ while GPT-5.4 improves from $56.0$ to $63.8$ overall SR.
However, stronger base models do not automatically yield better world-model guidance. For Qwen3-VL, scaling the unguided model from 8B to 32B improves overall SR, but adding Code2Image to the 32B model reduces it to $49.6$. 
This drop is most evident in text-input and text-edit tasks, where rendered images can miss fine-grained UI transitions and provide the judge with misleading progress signals (see Figure \ref{fig:androidworld_code2image_case_study} for case studies).
These results suggest that posterior selection is limited by both world-model fidelity and the agent's proposal ability. Larger agents such as Gemini 3-Flash generate more diverse executable candidates, whereas weaker agents often repeat actions (refer to the case in Figure  \ref{fig:small_scale_model_generation_case_study}). 
\textbf{\textit{Therefore, text-based world models can provide more accurate predictions in OOD environments, helping agents improve their task completion rates.}}

\begin{table*}[htbp]
    \centering
    \vspace{-0.3cm}
    \caption{AndroidWorld results for Qwen3-VL-8B under two agent settings. }
    \footnotesize
    \resizebox{\textwidth}{!}{
    \begin{tabular}{l c c c c c c c}
        \toprule
        \textbf{Model / WM} & \textbf{Easy SR} & \textbf{Medium SR} & \textbf{Hard SR} & \textbf{Overall SR} & \textbf{Tokens/Trial} & \textbf{Run Time/Task} & \textbf{Mean Steps} \\
        \midrule
        Qwen3-VL-8B (MA3) & 66.67 & 58.33 & 24.56 & 57.18 & 5161.22 & 241.04 & 11.29 \\
        + Delta Text & 70.49 & 50.00 & 15.79 & 55.17 & 37788.39 & 756.63 & 14.18 \\
        + Full Text & 69.26 & 41.67 & 15.79 & 51.94 & 50283.94 & 892.47 & 14.94 \\
        + Code2Image & 65.98 & 43.06 & 18.42 & 51.07 & 156940.67 & 3290.32 & 14.19 \\
        \midrule
        Qwen3-VL-8B (M3A) & 54.64 & 27.77 & 16.66 & 40.08 & 1698.13 & 144.56 & 13.69 \\
        + Delta Text &59.01 & 34.72 & 17.10 & 44.61 & 54461.79 & 439.93 & 13.02 \\
        + Full Text & 50.82 & 33.33 & 11.84 & 39.01 & 58548.32 & 1469.09 & 13.87 \\
        + Code2Image & 54.92 & 31.94 & 13.16 & 40.95 & 214360.66 & 2600.86 & 13.44 \\
        \bottomrule
    \end{tabular}
    }
    \vspace{-0.3cm}
    \label{tab:androidworld_qwen8b_framework_results}
\end{table*}

\noindent \textbf{Comparison of Different Agent Frameworks.}
Figure~\ref{fig:androidworld_qwen8b_framework_sr} and Table \ref{tab:androidworld_qwen8b_framework_results} show that the agent framework strongly affects whether world-model guidance helps. With Qwen3-VL-8B, the MA3.5 baseline reaches $57.2$ overall SR, driven by more comprehensive reasoning steps and reflection mechanisms. Yet adding the world-model pipeline to MA3.5 leads to consistent regression: Delta Text drops to $55.2$, Full Text to $51.9$, and Code2Image to $51.1$. 
When MA3.5 already proposes the correct action, an explicit world model can still hallucinate an incorrect future state, especially when the task depends on application-specific business logic. (see Appendix Figure \ref{fig:AndroidWorld_Text_Case_Study}). 


\begin{findingbox}{RQ1: What should mobile world models predict?}
Modality selection presents a strict trade-off: \textbf{text-based predictions} ensure superior OOD robustness, while \textbf{Code2Image} optimizes ID visual guidance. Conversely, \textbf{diffusion-based images} remain impractical due to severe text-rendering artifacts and prohibitive inference overhead.
\end{findingbox}

\section{How Should Mobile World Models Assist {GUI Agents} at Test Time?}
\label{sec:rq2}
After identifying the output modal characteristics of world models, we analyze the reasons behind the failure of world model guidance, investigate three interaction modes (posterior selection, self-reflection, and prior perception), and explore the test-time scaling law under guidance.

\subsection{Why Does Guidance Fail? Decision Confidence Analysis}

\begin{wrapfigure}{r}{0.6\textwidth}
    \vspace{-0.4cm}
    \centering
    \begin{minipage}[c]{0.44\linewidth}
        \includegraphics[width=\linewidth]{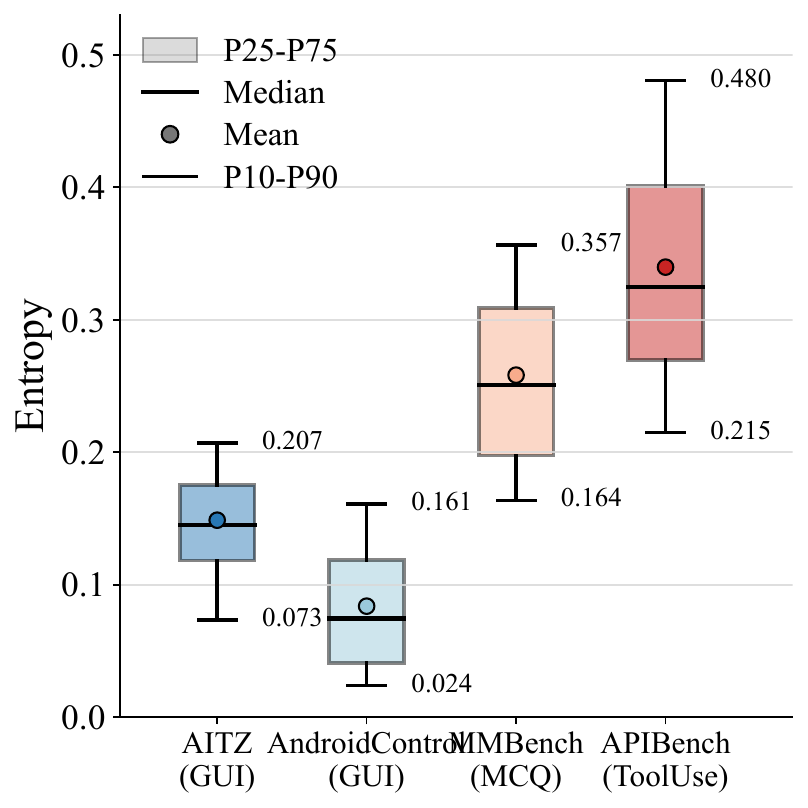}
    \end{minipage}
    \hspace{-0.2em}
    \begin{minipage}[c]{0.52\linewidth}
        \includegraphics[width=\linewidth]{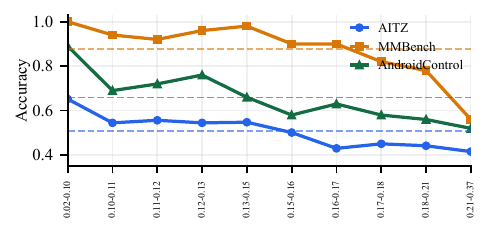}
        \vspace{-0.4em}
        \includegraphics[width=\linewidth]{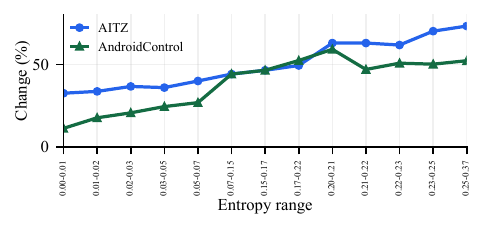}
    \end{minipage}
    \vspace{-0.2cm}
    \caption{Entropy statistics and entropy-conditioned behavior across GUI, MCQ, and tool-use benchmarks.}
    \vspace{-0.4cm}
    \label{fig:entropy_boxplot}
\end{wrapfigure}

We first examine whether action entropy reflects decision uncertainty in GUI tasks. 
Higher-entropy steps are generally less accurate, suggesting that uncertain actions are more likely to be erroneous; the full entropy-bin statistics are provided in Appendix Table~\ref{tab:entropy_step_accuracy} and Figure~\ref{fig:entropy_accuracy_bins}.
We further observe that more erroneous action corrections occur (28.05 to 59.64) in high-entropy steps in Table \ref{tab:self_reflection_to_perception}, but the overall entropy of GUI action distributions remains low, which is the main factor limiting the improvement from test-time world-model guidance. To further contextualize this effect, we compare GUI benchmarks with MMBench~\citep{liu2024mmbench}, a multiple-choice multimodal benchmark, and APIBench~\citep{patil2024gorilla}, a tool-use benchmark based on API-style interactions.
As shown in Figure~\ref{fig:entropy_boxplot}, Qwen3-VL-8B is highly confident in GUI settings: the mean entropy is only $0.15$ on AITZ and $0.08$ on AndroidControl, much lower than $0.26$ on MMBench and $0.34$ on APIBench.
Although APIBench also uses structured tool-use outputs (GUI task actions are used the similar output format), its entropy is substantially higher than that of GUI tasks, suggesting that the low entropy is not merely caused by structured action formatting. This pattern is also reflected by the Percentage column in Table~\ref{tab:offline_task_navigation}, where the sampling distribution of smaller agents is highly concentrated. In the general-domain MMBench setting, Qwen3-VL-8B reaches a score of $87.30$, but its output distribution is still clearly less concentrated than in GUI scenarios. This leads to the following takeaway: \textbf{\textit{Compared to the entropy of general tasks, GUI agents have a lower entropy, which decreases the diversity of candidate actions. This higher confidence further makes agents more inclined to keep their original actions, even when world models provide a clear signal of reflection regarding their erroneous actions.}}

\begin{table*}[t]
    \vspace{-0.3cm}
    \caption{Interaction-setting results for AITZ on action-type and step-level navigation metrics. In the greedy element interaction setting, 35.18 elements are predicted by world models per screenshot.}
    \centering
    \tiny
    \resizebox{0.9\textwidth}{!}{
    \begin{tabular}{l|c cc cc cc cc|c|c}
        \toprule
        \multirow{2}{*}{\textbf{Method / WM}} &
        \multirow{2}{*}{\textbf{Scroll}} &
        \multicolumn{2}{c}{\textbf{Click}} &
        \multicolumn{2}{c}{\textbf{Input}} &
        \multirow{2}{*}{\textbf{Press}} &
        \multirow{2}{*}{\textbf{Stop}} &
        \multicolumn{2}{c}{\textbf{Total}} &
        \multirow{2}{*}{\shortstack{\textbf{Corrected/}\\\textbf{Edited/WM Engagement}}} &
        \multirow{2}{*}{\textbf{Tokens}} \\
        \cmidrule(lr){3-4} \cmidrule(lr){5-6} \cmidrule(lr){9-10}
        & & \textbf{Type} & \textbf{Step} & \textbf{Type} & \textbf{Step}
        & & & \textbf{Type} & \textbf{Step} & & \\
        \midrule
        \multicolumn{12}{c}{\textbf{\textit{Qwen3-VL-8B -- Baseline Decoding}}} \\
        \midrule
        Base & 52.16 & 81.69 & 56.54 & 70.76 & 53.51 & 53.21 & 62.36 & 72.91 & 55.99 & -- & 28912.32 \\
        BO3 & 51.29 & 81.97 & 57.25 & 74.26 & 56.14 & 55.84 & 61.49 & 73.45 & 56.67 & -- & 64985.87 \\
        PASS@3 & 54.53 & 86.35 & 63.43 & 78.07 & 59.65 & 58.11 & 66.95 & 77.63 & 61.84 & -- & 63671.74 \\
        PASS@9 & 56.10 & 87.12 & 64.58 & 83.04 & 65.58 & 63.10 & 83.60 & 79.26 & 66.04 & -- & 189177.85 \\
        \midrule
        \multicolumn{12}{c}{\textbf{\textit{Qwen3-VL-8B -- Qwen-Judged Posterior Selection}}} \\
        \midrule
        Delta Text & 52.59 & 81.65 & 57.11 & 77.49 & 57.60 & 55.47 & 59.77 & 73.59 & 56.70 & --/--/66.41 & 92689.83 \\
        Full Text & 51.29 & 82.22 & 57.01 & 76.61 & 56.73 & 55.85 & 60.34 & 73.74 & 56.47 & --/--/66.41 & 97366.29 \\
        Code2Image & 53.80 & 81.79 & 58.25 & 76.02 & 58.31 & 56.82 & 65.51 & 73.94 & 58.18 & --/--/66.41 & 145842.96 \\
        \midrule
        \multicolumn{12}{c}{\textbf{\textit{Qwen3-VL-8B -- Gemini-Judged Posterior Selection}}} \\
        \midrule
        Delta Text & 52.80 & 81.84 & 57.39 & 76.90 & 58.19 & 55.09 & 59.20 & 73.59 & 56.87 & --/--/66.41 & 89570.31 \\
        Full Text & 55.37 & 85.50 & 60.32 & 79.32 & 60.02 & 58.85 & 62.48 & 76.94 & 59.73 & --/--/66.41 & 94282.22 \\
        Code2Image & 56.80 & 85.36 & 61.73 & 79.61 & 61.02 & 59.47 & 64.06 & 76.96 & 61.07 & --/--/66.41 & 143671.72 \\
        \midrule
        \multicolumn{12}{c}{\textbf{\textit{Qwen3-VL-8B -- Agent Self-Reflection}}} \\
        \midrule
        Delta Text & 50.65 & 83.49 & 59.39 & 78.94 & 59.64 & 50.56 & 50.86 & 73.31 & 56.75 & 23.69/28.05/100 & 31674.81 \\
        Full Text & 50.86 & 83.45 & 59.77 & 79.24 & 59.65 & 51.32 & 51.15 & 73.51 & 57.10 & 14.66/43.69/100 & 53266.62 \\
        Code2Image & 55.02 & 82.98 & 63.96 & 78.95 & 64.06 & 55.94 & 56.44 & 78.00 & 61.45 & 25.95/37.68/100 & 49307.51 \\
        \midrule
        \multicolumn{12}{c}{\textbf{\textit{Qwen3-VL-8B -- Entropy-Gated Self-Reflection}}} \\
        \midrule
        Delta Text & 48.73 & 84.61 & 61.91 & 79.29 & 70.95 & 60.95 & 61.47 & 79.43 & 60.94 & 49.62/59.64/12.5 & 28703.16 \\
        Full Text & 49.74 & 84.61 & 61.92 & 80.39 & 72.09 & 61.33 & 61.92 & 84.72 & 61.26 & 47.00/67.76/12.7 & 29029.94 \\
        Code2Image & 49.38 & 84.54 & 61.63 & 79.41 & 71.19 & 60.57 & 61.62 & 84.46 & 60.87 & 45.13/66.25/12.4 & 32727.72 \\
        \midrule
        \multicolumn{12}{c}{\textbf{\textit{Qwen3-VL-8B -- Greedy Element Interaction with Qwen Selection}}} \\
        \midrule
        Delta Text & 57.44 & 94.88 & 79.41 & 91.73 & 77.22 & 67.87 & 66.55 & 93.62 & 67.69 & -- & 1019588.13 \\
        Full Text & 58.12 & 94.51 & 80.30 & 92.84 & 78.08 & 67.96 & 69.12 & 93.16 & 68.64 & -- & 1071029.19 \\
        Code2Image & 61.03 & 95.76 & 81.24 & 94.21 & 80.83 & 69.14 & 70.03 & 94.24 & 70.39 & -- & 1604272.56 \\
        \midrule
        \multicolumn{12}{c}{\textbf{\textit{Gemini3 Flash -- Greedy Element Interaction with Gemini Selection}}} \\
        \midrule
        Delta Text & 70.18 & 97.33 & 84.10 & 95.67 & 83.45 & 72.91 & 73.22 & 96.29 & 74.95 & -- & 1112277.96 \\
        Full Text & 71.04 & 98.06 & 84.46 & 96.82 & 84.05 & 73.33 & 73.75 & 97.34 & 75.16 & -- & 1168395.48 \\
        Code2Image & 72.68 & 98.14 & 86.12 & 97.36 & 85.04 & 73.47 & 74.02 & 97.66 & 79.03 & -- & 1750115.52 \\
        \bottomrule
    \end{tabular}
    }
    \vspace{-0.4cm}
    \label{tab:self_reflection_to_perception}
\end{table*}

\subsection{Which Interaction Mode Works Best? Posterior, Reflection, and Prior}

\noindent \textbf{Settings.}  
We compare following interaction modes. (1) \textbf{Posterior guidance} but judged by different models. (2) \textbf{Hybrid prior-posterior guidance} asks agents to rethink after observing the world-model prediction. (3) \textbf{Prior guidance} queries world models over all elements before making decisions.

\noindent \textbf{Benchmark And Metrics.} We use AITZ~\citep{qin2025ui} because it provides element lists, which are suitable for prior-guidance. Corrected, Edited, and WM Engagement denote the rate of action correction, the ratio of accepted rethinking decisions, and the fraction of steps involving world models. 

\noindent \textbf{Limitations of Posterior Selection.}
Even with world models and Gemini-judge, Qwen3-VL-8B still struggles to reach the PASS@3 performance level under posterior selection. This gap further exposes the limitation of posterior interaction: \textbf{\textit{judging among sampled actions can improve the final choice, but it remains bounded by the quality and diversity of the sampled candidates.}} When we switch to agent self-reflection, we find that the agent still has difficulty updating an erroneous action into the correct one after rollout. For example, the correction rate of full-text self-reflection is only $14.66\%$. In contrast, using entropy as a self-activation gate yields better performance with higher efficiency: full-text entropy-gated self-reflection improves overall Step Acc. from $56.47$ under Qwen-judged posterior selection to $61.26$, while using only about one-fifth of the posterior-selection inference cost.

\noindent \textbf{Value of Prior Perception.}
Prior perception does not depend on the action model's own decision distribution. Therefore, it can still work when the agent assigns most probability to a few actions and leaves little room for posterior resampling. As shown in Table \ref{tab:self_reflection_to_perception}, with Qwen3-VL-8B selection, Code2Image prior perception reaches an overall Step Acc. of $70.39$, surpassing PASS@9 ($66.04$). The gains are especially large for Click and Input, suggesting that world models compare subtle differences among candidate elements effectively.
Figure~\ref{fig:aitz_html_casestudy}, \ref{fig:androidcontrol_delta_text_casestudy} and ~\ref{fig:androidcontrol_text_casestudy} provide qualitative examples of this effect. However, prior interaction is still constrained by element-list availability and inference efficiency, because AITZ contains $35.18$ elements per screenshot.
\textbf{\textit{Even so, prior perception remains an effective means for helping agents break through their capability boundaries at test time.}}


\begin{figure*}[t]
\vspace{-0.2cm}
    \centering
    \includegraphics[width=0.95\textwidth]{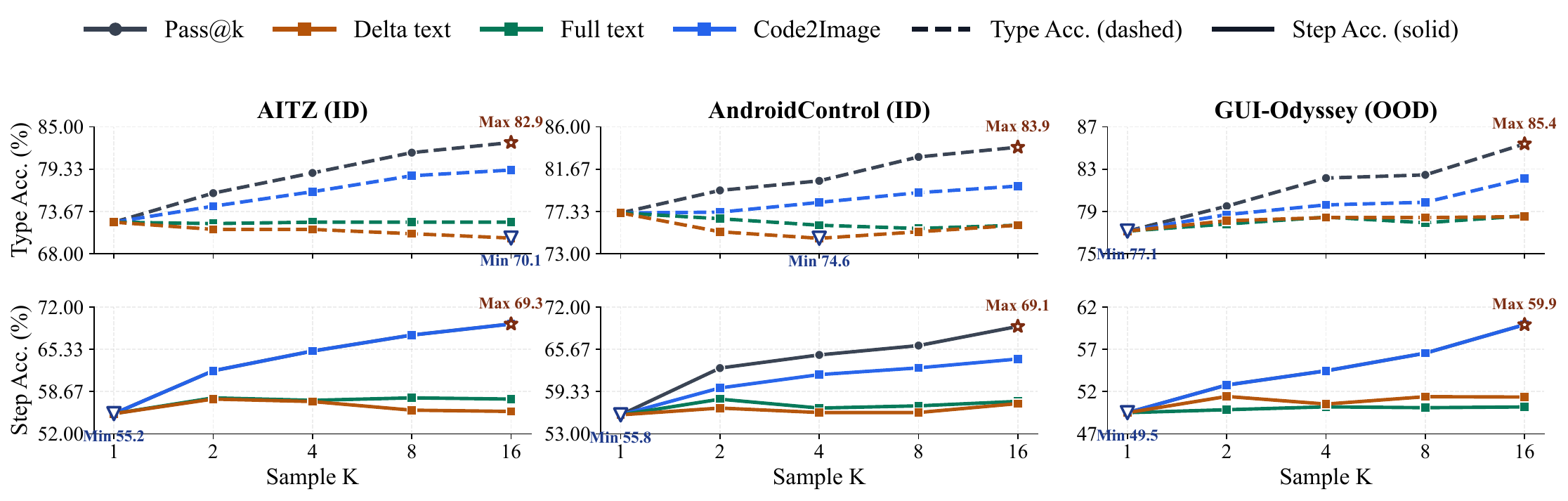}
    \vspace{-0.4cm}
    \caption{Test-time scaling trends on AITZ (ID), AndroidControl (ID), and GUI-Odyssey (OOD). The top row reports Type Acc. and the bottom row reports Step Acc. }
    \vspace{-0.4cm}
    \label{fig:test_time_scaling_gui}
\end{figure*}
\subsection{Does More Sampling Help? Test-Time Scaling Analysis}

Figure~\ref{fig:test_time_scaling_gui} shows that a larger sampling number consistently improves both action-type and step-level accuracy, while the gains from world-model feedback depend on representation quality. Code2Image improves almost in parallel with Pass@k as k increases, and this trend remains unsaturated at $k=16$. \textbf{\textit{This suggests that the sampling-space limitation caused by low-entropy proposals can be partly overcome by drawing more candidate actions.}} In contrast, the same trend does not hold for text-based feedback: as the candidate pool grows, the additional predicted text states become harder to distinguish, limiting the benefit of further sampling. At the same time, the original erroneous action may switch from one error type to another, which is consistent with the phenomenon in Table \ref{tab:offline_task_navigation}.

\begin{findingbox}{RQ2: How should mobile world models assist agents at test time?}
Mobile world models excel as \textbf{prior perceptual modules} rather than \textbf{posterior critics}. The overconfidence and low entropy of mobile agents starve posterior evaluation of candidate diversity. \textbf{Test-time scaling} yields marginal gains bounded by initial confidence and inference budgets.
\end{findingbox}

\section{Can {GUI Agents} Learn from Mobile World Model Imagination? }
\label{sec:rq3}
The last section shows that test-time guidance is bounded by output entropy and candidate diversity. Therefore, explore whether imagined trajectories from world models can solve this problem, allowing agents to acquire higher step accuracy and task completion rate through training process.



\noindent \textbf{Settings.} 
We adopt Code2Image as the virtual environment and Qwen3-VL-8B as the action model. 
We initialize the environment with the first screenshot and instruction in AndroidControl, then let the action model interact with the environment to complete the given task. The collected data are filtered by GPT-5.4 and then used to update the policy of the action model.

\begin{figure*}[t]
\vspace{-0.2cm}
    \centering
    \includegraphics[width=0.95\textwidth]{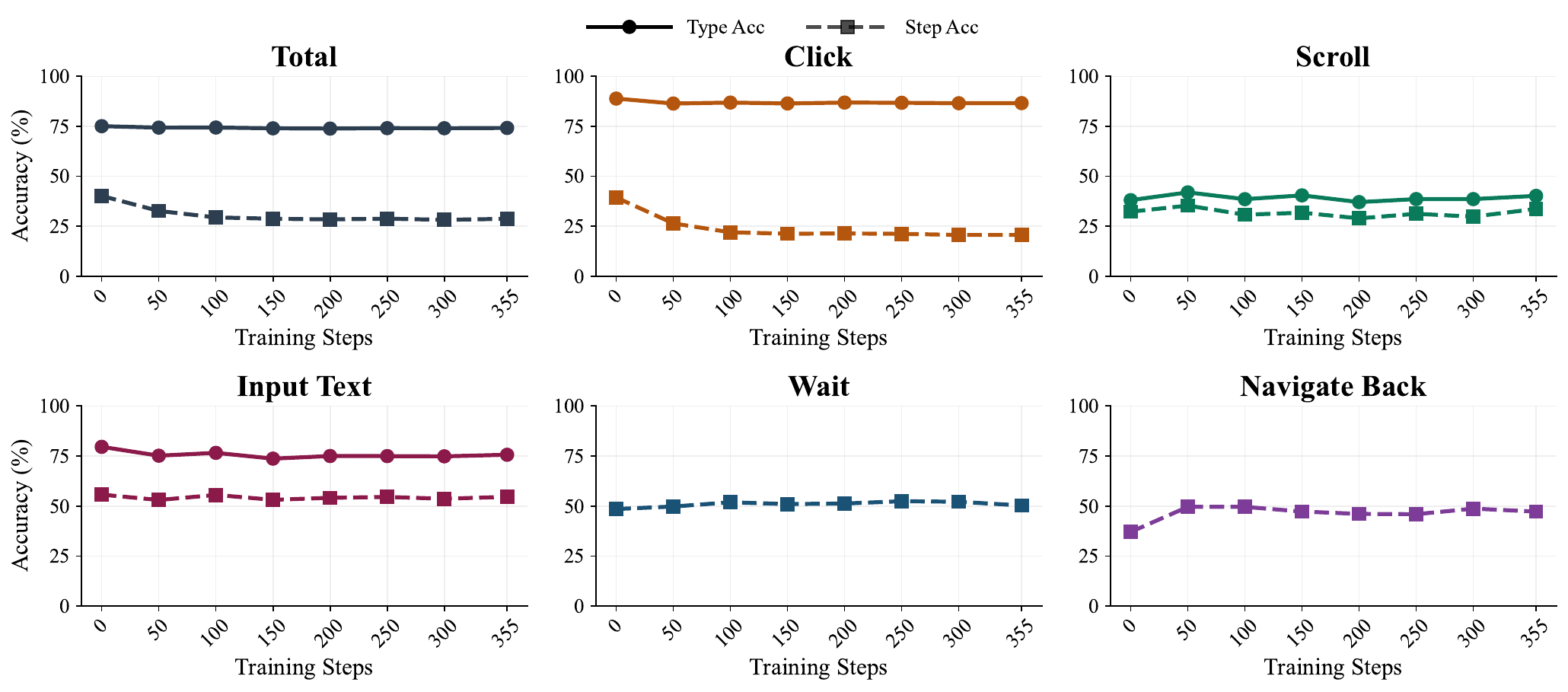}
    \vspace{-0.4cm}
    \caption{AndroidControl performance during training with World-Model imagination}
    \vspace{-0.4cm}
    \label{fig:ac_sft_performance}
\end{figure*}

\begin{figure*}[t]
\vspace{-0.2cm}
    \centering
    \subfigure[Click-coordinate distributions.]{
        \includegraphics[width=0.45\textwidth]{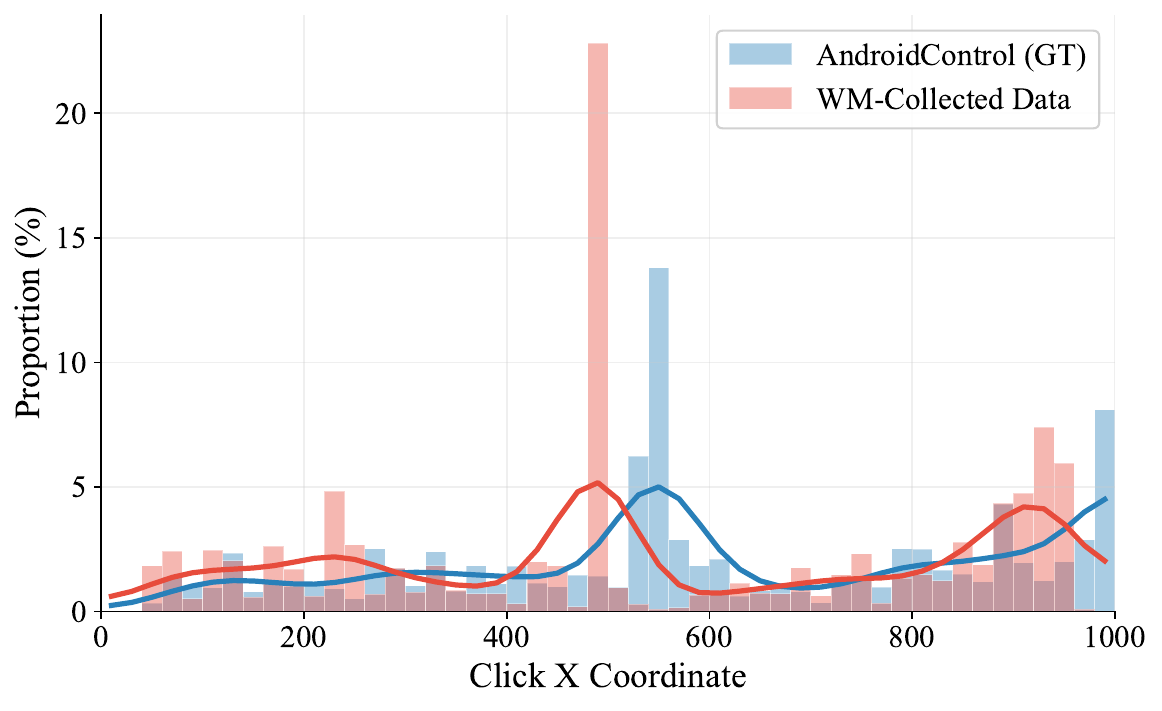}
        \label{fig:ac_position}
    }
    \hfill
    \subfigure[AndroidWorld performance during training.]{
        \includegraphics[width=0.45\textwidth]{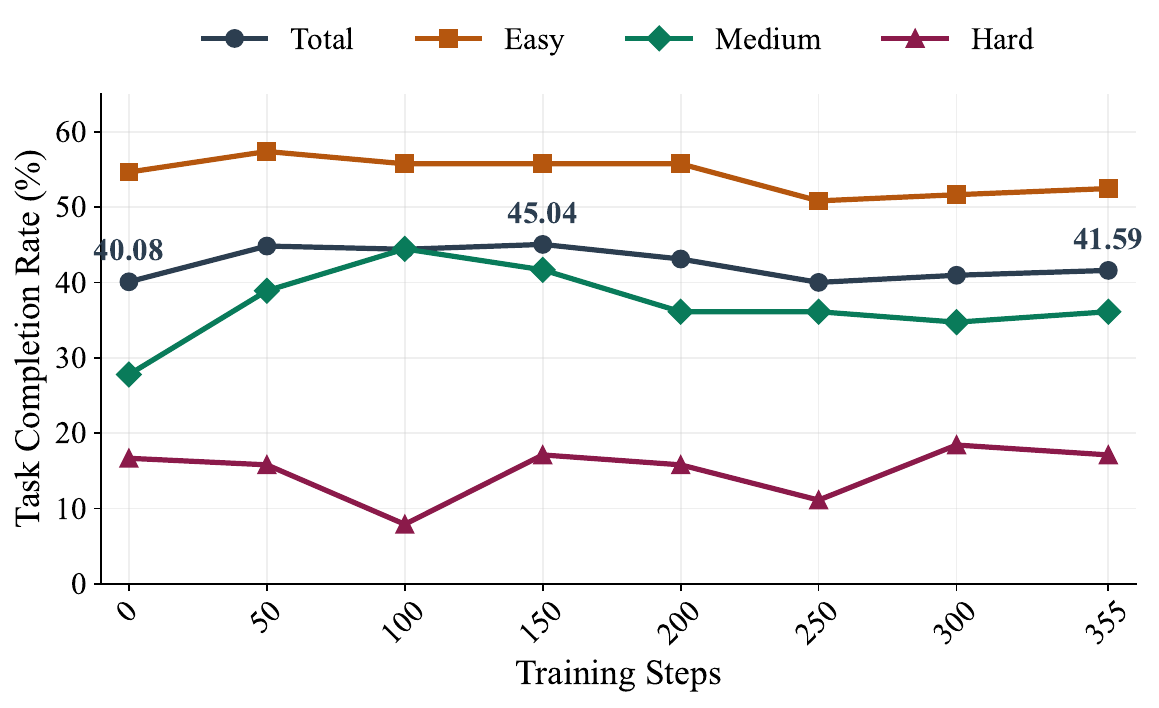}
        \label{fig:aw_performance}
    }
    \vspace{-0.2cm}
    \caption{Analysis of agent fine-tuned on World-Model trajectories.}
    \vspace{-0.4cm}
    \label{fig:sft_analysis}
\end{figure*}

\noindent \textbf{Experimental Results.} 
As shown in Figure~\ref{fig:ac_sft_performance}, although the initial states are derived from AndroidControl, fine-tuning on world-model-collected trajectories degrades offline evaluation performance (40.2 -> 28.7). We attribute this decline primarily to click accuracy (39.4 -> 20.7): Figure~\ref{fig:ac_position} reveals that click coordinates predicted by the fine-tuned agent are heavily concentrated around $x{=}499$, deviating substantially from the ground-truth distribution. This confirms a distributional gap between world-model-rendered pages and real Android screenshots, which propagates into degraded click precision after training.
In Appendix~\ref{sec:case_study_imagination}, we show that the world model tends to simplify page layouts and concentrate interactive elements toward the horizontal center, altering the perceived interaction targets and producing the observed centering bias.

Surprisingly, as shown in Figure~\ref{fig:aw_performance}, online evaluation on the out-of-distribution AndroidWorld improves (40.1 -> 45.0), despite AndroidWorld pages never appearing in the world model's training data. We hypothesize that because AndroidWorld measures end-to-end task completion rather than per-step accuracy, the fine-tuned agent benefits from stronger planning capabilities acquired through imaginary interaction, yielding a higher success rate. However, performance slightly declines with prolonged training (45.0 -> 41.6), likely because the trajectories are collected once by the base model, causing the policy to overfit to its limited behavioral distribution. These results suggest that mobile world models can serve as effective virtual environments for agent training: the imaginary trajectories 
remain sufficiently valid to improve downstream task performance, while also highlighting the need for iterative and more diverse data collection to mitigate distributional shift.

\begin{findingbox}{RQ3: Can Agents Learn from mobile world model Imagination?}
Imagination trajectories benefit agents by providing \textbf{transferable interaction experience} rather than perfectly replicating the \textbf{source distribution}, which improves the agent's \textbf{generalization} capabilities, provided the generated behaviors are actionable and useful.
\end{findingbox}

\section{Related Work}
In this section, we discuss recent advancements in mobile GUI agents~\citep{yang2026gui,shen2024falcon} serving as vision-language-action (VLA) models and summarize their current limitations. 
We review current agent frameworks~\citep{zhang2025appagent,wang2024mobile,wang2025mobile} and reinforcement learning-based multimodal reasoning agents~\citep{luo2025gui,zhou2025gui,liu2025infigui,lu2026ui}.
Furthermore, we also analyze the application of textual world models~\citep{cao2026mobiledreamer} during test-time in mobile GUI tasks. 
Additionally, approaches based on diffusion models~\citep{luo2025vimo} and rendering code~\citep{xiao2026webworld,koh2026generative,zheng2026code2world} seek to reconstruct and reproduce GUI interfaces to equip VLAs with richer multimodal information. Detailed analysis is provided in Appendix \ref{relatedworksappendix}.

\section{Conclusion}
We introduce a trajectory-annotated mobile GUI world-model dataset and instantiate four modalities: delta text, full text, renderable code, and diffusion-based image prediction.
Offline and online evaluations show that world model feedback improves GUI agents, but the gains depend on interaction mode, model confidence, modality, and inference cost.
We further find that posterior selection is limited by low entropy decisions, entropy-gated self-reflection is more efficient, and prior perception is effective but constrained by element lists, cost, and trajectory induced bias.

\newpage
\bibliographystyle{unsrtnat}
\bibliography{mobileworld}

\newpage
\appendix

\section{Full Related Work}\label{relatedworksappendix}
\subsection{Mobile GUI Agents}
The evolution of mobile GUI agents marks a paradigm shift from rule-based automation to foundation model-driven interaction~\citep{deng2024mobile,xusman}. Early methods heavily rely on structured system metadata such as view hierarchies or accessibility trees. However, these methods often struggle with missing attributes and system compatibility issues in real-world scenarios. The emergence of multimodal large language models enables agents to directly perceive user interfaces visually, significantly advancing the field~\cite{qin2025ui, shen2024falcon}. Frameworks including AppAgent~\cite{zhang2025appagent}, CogAgent~\cite{hong2023cogagent} and MobileSteward~\citep{liu2025mobilesteward} introduce a plan-and-act paradigm based on raw screenshots, while the Mobile-Agent series~\cite{wang2024mobile, wang2025mobile} progressively enhances the robustness of complex task execution through a multi-agent collaborative architecture. 
GUI reasoning typically requires integrating multiple capabilities, including screen understanding, task planning, and element grounding, and therefore benefits from fine-grained optimization across different reasoning stages~\citep{huang2025mobileipl,liu2026come}.
Recently, reinforcement learning further optimizes GUI navigation trajectories 
~\cite{yang2026gui, wang2025ui,xu2025llm,xu2025mobile}, as seen in MobileGUI-RL~\citep{shi2025mobilegui} and STEP~\citep{chen2025step}, by aligning agent policies with sparse environmental rewards through online exploration.

Despite these advancements, a critical bottleneck persists as most current mobile agents operate strictly as reactive decision-makers that sample actions solely based on history and current observations. In the highly dynamic environment of mobile platforms with a massive action space, this lack of foresight prevents agents from evaluating and anticipating unrecoverable navigation deviations before execution. This limitation inevitably causes compounding errors during long-horizon tasks \cite{zhang2024android}.

\subsection{GUI World Models}

World models originate from reinforcement learning and physical simulation. They endow agents with human-like foresight by predicting future states given current observations and proposed actions. Recently, vision-language models facilitate the application of world models in GUI automation \cite{gu2024your, chae2024web, guan2026computer,jiang2026r,sun2024determlr}. Existing research primarily adopts two modalities to construct world models. 
The first modality focuses on text-level world models, which structure GUI state transitions into semantic text representations \cite{li2025word}. For example, MobileWorld \cite{li2025mobileworldbench} predicts future states through natural language descriptions of GUI changes and question-answering pairs. To bridge the gap between pure semantics and spatial awareness, MobileDreamer \cite{cao2026mobiledreamer} proposes an efficient text-sketch world model that converts digital images into task-relevant key text sketches.
The second modality involves pixel-level world models, which instantiate future GUI states as image representations to provide GUI agents with intuitive visual foresight. Based on the underlying generation mechanisms, this paradigm evolves into two technical routes. The direct diffusion route, represented by ViMo \cite{luo2025vimo}, utilizes diffusion models to directly render high-fidelity RGB images in the pixel space after action execution. Such methods preserve complete visual topology but suffer from high computational overhead and a tendency to generate visual artifacts on tiny text or icons. Alternatively, the code rendering route emerges to address text distortion and structural loss caused by generated pixels, with recent paradigms including Code2world \cite{zheng2026code2world} and gWorld\cite{koh2026generative}. These models utilize vision-language models to predict executable underlying UI code such as HTML and then render the final visual screenshots using browser engines to achieve extremely high visual fidelity. Although the underlying process relies on discrete symbol generation, the final output delivered to the agent remains an aligned pixel-level predictive image.
Although these two representational paradigms demonstrate the feasibility of predicting future GUI states, how these world models can effectively guide GUI agents remains a severely neglected area. This work fills this gap. By systematically analyzing the interactive closed loop between world models and GUI agents, we aim to establish a rigorous analytical foundation for future proactive GUI automation.

\section{Experimental Details And Results}

\begin{figure*}[htbp]
    \centering
    \includegraphics[width=\textwidth]{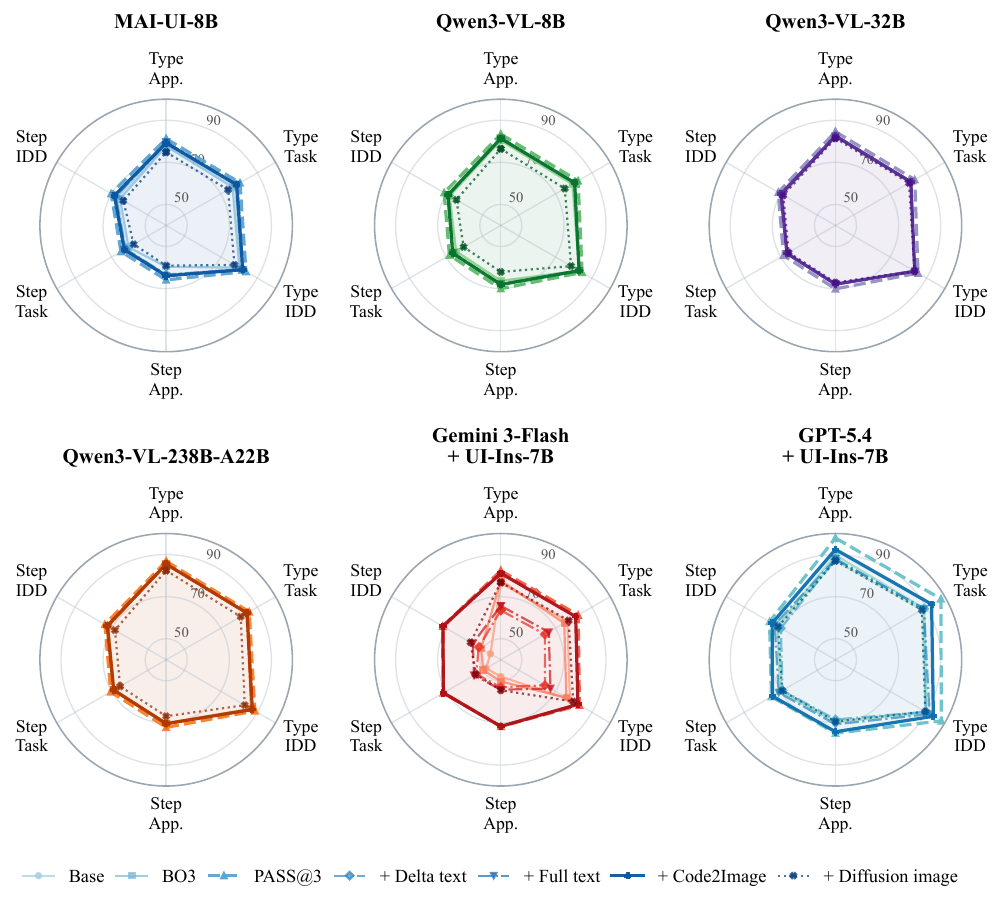}
    \caption{Six-dimensional radar plots for offline task navigation. Each subplot corresponds to one base GUI agent, and each curve corresponds to one setting. The six axes report App., Task., and IDD scores for Type Acc. and Step Acc., excluding the two Overall columns in Table~\ref{tab:offline_task_navigation}. All axes use the same raw accuracy scale from 40 to 100 to preserve the relative scale between Type Acc. and Step Acc. while removing unused radial space.}
    \label{fig:offline_task_radar_6d}
\end{figure*}

\subsection{Full Results on AndroidWorld}

\noindent \textbf{Experimental Results.}
The results in Figure~\ref{fig:androidworld_overall_sr}  indicate that a purely reactive strategy (m3a) is insufficient for most GUI agents when handling medium and long-horizon tasks. Across various foundation models, the overall performance exhibits a significant leap with the assistance of text-level world model feedback. Taking delta text as an example, it increases the overall success rate (overall\_SR) of Gemini 3-Flash from $50.57$ to $66.59$, while the success rate on medium-difficulty tasks (medium\_SR) substantially rises from $44.44$ to $58.33$. For the state-of-the-art model GPT-5.4, delta text similarly elevates its overall success rate from $56.03$ to $63.79$, demonstrating that even foundation models with exceptionally strong zero-shot planning capabilities benefit from explicit lookahead state verification.

Scaling up the model size does not necessarily improve the effectiveness of the world model proportionally. For the qwen3-vl series models, increasing the parameter size from 8b to 32b raises the unguided baseline overall success rate from $40.08$ to $50.86$. However, after integrating code2image, the overall success rate of the 32B version degrades from the baseline of $50.86$ to $49.57$. This performance degradation occurs because, in tasks requiring high-frequency and fine-grained state confirmation such as text input and modification, the images rendered by code2image often fail to simulate continuous physical-level transition trajectories. This transmits false task progress signals to the judge, thereby lowering the rank of the correct action among the three candidate actions.

Throughout the propose, simulate, and select pipeline, the capability ceiling of the foundation model determines the boundary between task success and failure. Larger models such as Gemini 3-Flash provide more diverse action proposals during the propose phase that are more likely to be correct for the current step. Smaller models frequently propose repetitive actions. This behavior not only exposes their lack of deep causal understanding regarding mobile GUI interactions but also inherently undermines the efficiency and purpose of the entire heuristic search framework.

An in-depth analysis of the scoring rationales provided by the judge reveals that the large language model serving as the judge, specifically Gemini 3-Flash, inherently possesses world model capabilities. This implicit world model is often substantially more rational and accurate than the explicit world model tasked specifically with generating predictions. When facing hallucinatory or erroneous predictions generated by the explicit world model, the judge reliably assesses the validity of the actions without being misled by the flawed predictions.

\noindent \textbf{Comparison of Different Agent Frameworks.}
As shown in Figure~\ref{fig:androidworld_qwen8b_framework_sr}, compared to the purely reactive m3a framework, the native baseline of Qwen3-VL-8B within a multi-agent collaborative framework featuring self-reflection and dynamic planning exhibits exceptional capability. It achieves an overall success rate of $57.18$, representing a $17.1$ percentage point improvement over the m3a native baseline.

However, integrating the world model pipeline into the MA3 framework yields a counterintuitive and comprehensive performance regression. The introduction of the world model results in negative gains across all modalities, with Delta Text dropping to $55.17$, Full Text decreasing to $51.94$, and Code2Image drastically reducing the success rate to $51.07$.

Tracing the failure trajectories reveals that when the MA3 framework proposes a correct action, the world model generates an erroneous or distorted future state due to a lack of prior knowledge regarding the business logic of the application. The judge consequently assigns a low score to the correct action based on this flawed prediction. Within the actor, world model, and judge verification loop, the judge model relies on the future state representations provided by the world model for scoring. When the actor executes a perfectly correct action, such as consecutively clicking a button in the BrowserMultiply task, the world model produces a distorted prediction indicating no changes in the user interface. This structural hallucination directly misleads the strictly rational judge into erroneously concluding that the agent is trapped in an invalid loop. As a result, the judge imposes severe negative penalties, such as $-0.8$ or $-1.0$, on the valid action. These false negatives, inherently manufactured by the verification pipeline, constitute the core of verification pollution.

The experimental results indicate that the external world model does not function as a universal performance amplifier independent of the base policy. For agents with weaker native policies, the world model serves as an effective buffer for trial and error. However, for agents with strong inherent planning capabilities such as MA3, forcibly integrating a verification loop poisons the decision-making and planning process if the logical fidelity of the verification node fails to match the planning depth of the generation node.




\subsection{Entropy And Step Accuracy}

\begin{figure*}[htbp]
    \centering
    \subfigure[Accuracy trends across entropy ranges.]{
        \includegraphics[width=0.48\textwidth]{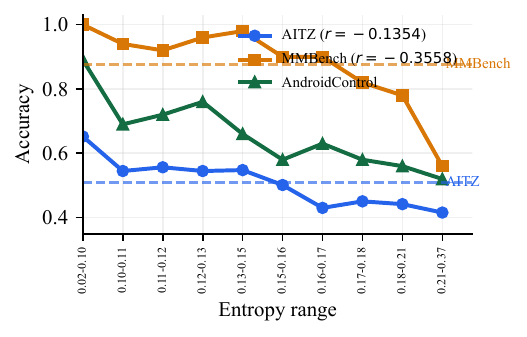}
    }
    \hfill
    \subfigure[Self-reflection change rate across entropy ranges.]{
        \includegraphics[width=0.46\textwidth]{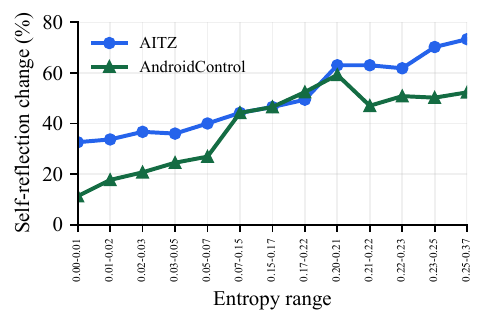}
    }
    \caption{Entropy-range analysis for GUI world-model feedback. Left: accuracy trends across entropy ranges on AITZ, MMBench, and AndroidControl, with ranges ordered from low to high entropy and dashed lines indicating reported overall accuracy for AITZ and MMBench. Right: self-reflection change rates on AITZ and AndroidControl, showing that higher-entropy ranges lead to more frequent action changes.}
    \label{fig:entropy_accuracy_bins}
\end{figure*}

\begin{table*}[htbp]
    \setlength{\abovecaptionskip}{0.2cm}
    \setlength{\belowcaptionskip}{-0.1cm}
    \centering
    \footnotesize
    \resizebox{\textwidth}{!}{
    \begin{tabular}{c c c c c c}
        \toprule
        \textbf{AITZ Entropy} & \textbf{AITZ Acc.} & \textbf{MMBench Entropy} & \textbf{MMBench Acc.} & \textbf{AndroidControl Entropy} & \textbf{AndroidControl Acc.} \\
        \midrule
        {[0.02, 0.10]} & 0.65 & [0.02, 0.11] & 1.00 & [0.00, 0.03] & 0.89 \\
        {[0.10, 0.11]} & 0.54 & [0.11, 0.13] & 0.94 & [0.03, 0.05] & 0.69 \\
        {[0.11, 0.12]} & 0.56 & [0.13, 0.16] & 0.92 & [0.05, 0.06] & 0.72 \\
        {[0.12, 0.13]} & 0.54 & [0.16, 0.18] & 0.96 & [0.06, 0.07] & 0.76 \\
        {[0.13, 0.15]} & 0.55 & [0.18, 0.20] & 0.98 & [0.07, 0.08] & 0.66 \\
        {[0.15, 0.16]} & 0.50 & [0.20, 0.22] & 0.90 & [0.08, 0.08] & 0.58 \\
        {[0.16, 0.17]} & 0.43 & [0.22, 0.25] & 0.90 & [0.08, 0.09] & 0.63 \\
        {[0.17, 0.18]} & 0.45 & [0.25, 0.28] & 0.82 & [0.09, 0.11] & 0.58 \\
        {[0.18, 0.21]} & 0.44 & [0.28, 0.31] & 0.78 & [0.11, 0.13] & 0.56 \\
        {[0.21, 0.37]} & 0.42 & [0.32, 0.58] & 0.56 & [0.13, 0.19] & 0.52 \\
        \bottomrule
    \end{tabular}
    }
    \caption{Accuracy by entropy bin on AITZ, MMBench, and AndroidControl. Overall accuracy is 0.5087 for AITZ and 0.8760 for MMBench. The entropy-accuracy correlation is -0.1354 on AITZ and -0.3558 on MMBench.}
    \label{tab:entropy_step_accuracy}
\end{table*}

\noindent \textbf{Interpretation.}
Table~\ref{tab:entropy_step_accuracy} shows that the negative relationship between entropy and accuracy is not unique to GUI navigation: it appears in both the GUI setting (AITZ) and the general multimodal setting (MMBench). This trend clarifies why entropy is a useful indicator for deciding when additional posterior judgment is valuable. High-entropy states correspond to uncertain model decisions and therefore provide more room for effective resampling or correction. In such cases, a world model can act as a posterior verifier that selects or revises actions more effectively. In other words, high entropy indicates larger latent improvement potential, and world-model feedback is more likely to release that potential. Conversely, when GUI agents are already highly confident, the effective sampling rate is low, leaving fewer opportunities for the world model to change the final action.

\subsection{World Model Inference Token Statistics}
\label{sec:wm_token_stats}

\begin{table*}[htbp]
    \setlength{\abovecaptionskip}{0.3cm}
    \setlength{\belowcaptionskip}{-0.2cm}
    \footnotesize
    \centering
    \resizebox{0.82\textwidth}{!}{
    \begin{tabular}{lcc}
        \toprule
        \textbf{File} & \textbf{Training tokens (mean)} & \textbf{Completion tokens (mean)} \\
        \midrule
        MobileWorldBench (Qwen3-VL-8B) & 123.64 & 103.62 \\
        Delta description (Qwen3-VL-8B) & 175.39 & 121.68 \\
        Full description (Qwen3-VL-8B) & 360.44 & 368.15 \\
        HTML code (Qwen3-VL-8B) & 5357.57 & 5361.23 \\
        Code2World (Qwen3-VL-8B) & 4321.82 & 3539.58 \\
        Gemini-3.1-Flash-Image & -- & 2649.21 \\
        \bottomrule
    \end{tabular}
    }
    \caption{World model inference token statistics.}
    \label{tab:wm_token_stats}
\end{table*}

\noindent \textbf{Analysis.}
The token statistics in Table~\ref{tab:wm_token_stats} highlight a practical trade-off in distilling render code from a strong teacher model such as Claude-4.6-Opus. HTML code can improve visual fidelity by preserving layout, widget structure, and textual content, but it also introduces much longer outputs than textual world-model targets. The mean completion length of HTML code reaches 5361.23 tokens, compared with 121.68 tokens for delta descriptions and 368.15 tokens for full descriptions. For a small world model such as Qwen3-8B, even a 24k-token output budget remains difficult for long-tail HTML examples: when the generated HTML is truncated or structurally incomplete, the page may fail to render at all. This failure is more severe than a low-quality rendering, because an incomplete HTML page provides no usable visual state for downstream posterior judgment. Therefore, we use the token distribution of training and evaluation data to filter or prioritize render-code annotations that can be fully expressed by small world models. Longer examples are retained mainly for analyzing output-length pressure and efficiency rather than serving as the primary distillation target.

\section{Evaluation Metrics of Next UI Prediction}
\label{appendix:eval_metrics}

We use the same evaluation protocol tailored for GUI World Models as code2world~\citep{zheng2026code2world} does. This protocol introduces four specialized metrics across two complementary dimensions: \textbf{Functional Logic} and \textbf{Visual Quality}. These metrics are designed to provide a fairer and more granular comparison. We employ a unified \textit{VLM-as-a-Judge} framework to approximate human judgment. The detailed prompts for these VLM-based metrics are provided in Appendix \ref{appendix:evaluation}.

\subsection{Functional Logic} 
Functional Logic evaluates the functional correctness of state transitions, verifying whether the world model acts as a reliable simulator.

\textbf{Action Adherence ($S_{ad}$)} \\
This metric assesses whether the predicted next state $\hat{I}_{t+1}$ is a logically valid consequence of executing action $a_t$ on state $I_t$. Unlike simple visual coherence, $S_{ad}$ penalizes "hallucinations" where the visual update contradicts the intended interaction (e.g., clicking "Back" but staying on the same page).

Formally, let $\mathcal{J}_{\text{act}}$ be the VLM judge, the score for a test dataset $\mathcal{D}$ is defined as:
\begin{equation}
S_{ad} = \frac{1}{|\mathcal{D}_{test}|} \sum_{(I_t, a_t) \in \mathcal{D}_{test}} \mathcal{J}_{\text{act}}(I_t, a_t, \hat{I}_{t+1})
\end{equation}

\textbf{Action Identifiability ($S_{id}$)} \\
This metric evaluates the causal clarity of the generation. A high-fidelity simulation should allow an observer to infer the cause of a state change solely from the visual outcome. We instruct the VLM to act as an inverse dynamics model $\mathcal{J}_{inv}$, predicting the action type $\hat{a}_t$ based on the visual difference between $I_t$ and $\hat{I}_{t+1}$. Crucially, a high $S_{id}$ ensures that the "Selector" in our agent pipeline can correctly verify whether a simulated outcome matches the planned action. The metric is calculated as the classification accuracy:
\begin{equation}
    S_{id} = \frac{1}{|\mathcal{D}_{test}|} \sum_{(I_t, a_t) \in \mathcal{D}_{test}} \mathbb{1}\left[ \mathcal{J}_{inv}(I_t, \hat{I}_{t+1}) = \text{type}(a_t) \right]
\end{equation}

\subsection{Visual Quality}
Models following the \textit{renderable code generation paradigm} typically employ a \textit{semantic placeholder strategy} to guarantee structural correctness while avoiding the hallucination of external assets. Standard embedding metrics (e.g., SigLIP, DINO), which primarily capture \textit{high-level} semantic similarity, are ill-suited for this nuance; they lack the granularity to explicitly measure \textit{fine-grained} element alignment and structural layout, often penalizing valid stylistic abstractions. To address this, we propose two specialized metrics to disentangle structural fidelity from textural style.

\textbf{Element Alignment ($S_{ele}$).} 
This metric verifies the fine-grained positioning of UI components. It measures whether key interactive elements (buttons, inputs) present in the ground truth $I^*_{t+1}$ are accurately reflected in the generation $\hat{I}_{t+1}$ at correct relative coordinates, explicitly tolerating semantic placeholders provided they occupy the correct screen area.

\textbf{Layout Integrity ($S_{lay}$).} 
This metric evaluates global layout integrity, penalizing issues common in weak code generation such as CSS collapse, overlapping containers, or misalignment. 

Formally, the VLM judge $\mathcal{J}_{vis}$ compares the generated output against the ground truth under specific criteria and provides a composite score:
\begin{equation}
    S_{ele/stc} = \frac{1}{|\mathcal{D}_{test}|} \sum_{I^*_{t+1} \in \mathcal{D}_{test}} \mathcal{J}_{vis}(I^*_{t+1}, \hat{I}_{t+1})
\end{equation}

\subsection{Evaluation Metrics}
\label{appendix:evaluation}

\subsubsection{Action Adherence Metrics}
\begin{prompt}{System Prompt}
You are an expert ``UI Dynamics Judge''.
Your task is to evaluate the logical correctness of a World Model's prediction.
You will be given the \textbf{Current UI State (Image 1)}, a user's \textbf{Action}, and the \textbf{Predicted Next State (Image 2)}.

\textbf{IMAGE DEFINITIONS}
\begin{itemize}
    \item \textbf{Image 1}: Real screenshot BEFORE the action.
    \item \textbf{Image 2}: Predicted screenshot generated by the model (Rendered from HTML).
    \begin{itemize}
        \item \textit{Note}: Image 2 uses \textbf{Gray Placeholders} (e.g., \verb|[IMG: icon]|) instead of real images. Treat these as valid visual elements if their text description matches the context.
    \end{itemize}
\end{itemize}

\textbf{EVALUATION CRITERIA} \\
Evaluate the transition based on \textbf{Action Adherence} and \textbf{Context Preservation}.
\begin{enumerate}
    \item \textbf{Did the Action Take Effect?}
    \begin{itemize}
        \item If ``Click'', did the button trigger the correct navigation/popup?
        \item If ``Input Text'', does the EXACT text appear?
        \item If ``Scroll'', did the content shift correctly?
    \end{itemize}
    \item \textbf{Is the Context Preserved?}
    \begin{itemize}
        \item Non-active elements (status bar, bottom nav) should remain stable.
    \end{itemize}
\end{enumerate}

\textbf{SCORING RUBRIC (0.0-10.0)}
\begin{itemize}
    \item \textbf{9.5-10.0 (Perfect)}: The transition is flawless. Text is exact, layout is perfect, logic is undeniable.
    \item \textbf{8.0-9.4 (Good)}: Action executed correctly. Minor visual glitches (e.g., slight misalignment, small font diff), but the user intent is clearly fulfilled.
    \item \textbf{6.0-7.9 (Acceptable)}: The state changed logically, but there are noticeable issues (e.g., wrong icon style, text has typos, or layout is messy).
    \item \textbf{3.0-5.9 (Ambiguous)}: Something changed, but it's unclear if it was the \textit{right} change. (e.g., opened the wrong page, or screen turned white but kept headers).
    \item \textbf{1.0-2.9 (Failed)}: The action clearly failed (e.g., clicked a button but screen didn't move).
    \item \textbf{0.0-0.9 (Broken/Hallucination)}: The model generated a blank screen, noise, or a completely hallucinated interface unrelated to the app.
\end{itemize}

\textbf{OUTPUT FORMAT} \\
Provide a Single JSON Object: \\
\verb|{| \\
\verb|  "score": <float 0.0-10.0>,| \\
\verb|  "reasoning": "A concise summary of why this score was given..."| \\
\verb|}|
\end{prompt}

\begin{prompt}{User Prompt}
\textbf{INTERACTION DATA}
\begin{itemize}
    \item \textbf{User Intent}: ``\texttt{\{instruction\}}''
    \item \textbf{Action Description}: ``\texttt{\{semantic\_description\}}''
    \item \textbf{Action Data}: \texttt{\{action\_json\}}
\end{itemize}

\textbf{VISUAL INPUTS}
\begin{itemize}
    \item \textbf{Image 1}: Current State (Before)
    \item \textbf{Image 2}: Predicted Next State (After)
\end{itemize}

Please evaluate the transition quality on a scale of 0.0 to 10.0.
\end{prompt}

\subsubsection{Action Identifiability Metrics}
\begin{prompt}{System Prompt}
You are an expert ``Inverse Dynamics'' Judge for UI interactions.
Your task is to infer the user's action by analyzing the visual transition between the \textbf{Current State (Image 1)} and the \textbf{Predicted Next State (Image 2)}.

\textbf{ACTION CATEGORIES} \\
Choose EXACTLY ONE from the following list that best explains the change:
\begin{enumerate}
    \item \textbf{click}: A tap on a button, icon, or link. Result: Page navigation, popup opens, toggle switches, or focus change.
    \item \textbf{long\_press}: A sustained touch. Result: Context menu appears or item selection mode triggers.
    \item \textbf{scroll}: The content shifts vertically or horizontally. (New content appears, old content moves off-screen).
    \item \textbf{input\_text}: Text appears in an input field (without an explicit enter press).
    \item \textbf{open\_app}: The screen transitions from a launcher/home screen to a specific app interface.
    \item \textbf{navigate\_home}: Returns to the device home screen/launcher.
    \item \textbf{navigate\_back}: Returns to the previous screen (reverse navigation).
    \item \textbf{wait}: No significant visual change, or a loading spinner continues spinning.
    \item \textbf{none}: The transition is hallucinated, broken, illogical, or the image is blank.
\end{enumerate}

\textbf{INFERENCE RULES}
\begin{itemize}
    \item If Image 2 shows a keyboard appearing and text in a box $\rightarrow$ \textbf{input\_text}.
    \item If Image 2 is completely different layout (app switch) $\rightarrow$ \textbf{open\_app} or \textbf{navigate\_home}.
    \item If Image 2 is just the same list but shifted $\rightarrow$ \textbf{scroll}.
    \item If Image 2 has a visual glitch that makes no sense $\rightarrow$ \textbf{none}.
\end{itemize}

\textbf{OUTPUT FORMAT} \\
Provide a Single JSON Object: \\
\verb|{| \\
\verb|  "inferred_action": "string",| \\
\verb|  // Must be one of: click, long_press, scroll, input_text,| \\
\verb|  // open_app, navigate_home, navigate_back, wait, none| \\
\verb|  "reasoning": "Brief explanation of visual evidence."| \\
\verb|}|
\end{prompt}

\begin{prompt}{User Prompt}
\textbf{VISUAL INPUTS}
\begin{itemize}
    \item \textbf{Image 1}: Current State (Before)
    \item \textbf{Image 2}: Predicted Next State (After)
\end{itemize}

Based on the visual difference, what action did the user perform?
\end{prompt}

\subsubsection{Element Alignment and Layout Fidelity Metrics}
\begin{prompt}{System Prompt}
You are an expert \textbf{GUI Design Evaluation AI}.
Your task is to compare a \textbf{Generated UI Prediction (Image 2)} against a \textbf{Ground Truth UI Screenshot (Image 1)} and assess similarity.
You must act as a \textbf{strict judge}, penalizing deviations in \textbf{element position, content, and structure}.
If the prediction uses \textbf{gray image placeholders} (e.g., \verb|[IMG: avatar]|), apply \textbf{placeholder equivalence}: do not penalize missing real photos, but judge whether the placeholder matches the GT image region in \textbf{position, size, and semantic tag}.

\textbf{OUTPUT REQUIREMENT (Strict)}:

Return \textbf{JSON only} and follow the exact schema required by the user prompt.
Do not output any extra text.
\end{prompt}

\begin{prompt}{User Prompt}
\textbf{Task Definition.}
You are provided with two images:
\begin{enumerate}
    \item \textbf{Reference Image (Ground Truth)}: the expected correct UI (Image 1).
    \item \textbf{Candidate Image (Prediction)}: the UI generated by a model (Image 2).
\end{enumerate}
Evaluate the Candidate Image based on the following two metrics and output the scores strictly.

\vspace{4pt}
\textbf{Metric 1: Element Alignment (Score 1.0--10.0)}
\begin{itemize}
    \item \textbf{Definition}: Measures whether core UI elements are present and aligned with the GT.
    \item \textbf{What to check (strict)}:
    \begin{itemize}
        \item Presence of major elements (top bar, title, key text blocks, buttons/CTAs, list/cards, navigation).
        \item \textbf{Alignment}: relative positions, anchors, and spacing (padding/margins) compared to GT.
        \item Element sizing/proportions (width/height), including component boundaries.
    \end{itemize}
\end{itemize}

\textbf{Metric 2: Layout Integrity (Score 1.0--10.0)}
\begin{itemize}
    \item \textbf{Definition}: Measures whether the overall layout framework and visual hierarchy match the GT.
    \item \textbf{What to check (strict)}:
    \begin{itemize}
        \item Global layout hierarchy (top/middle/bottom regions; grouping; column vs row structure).
        \item Visual hierarchy and emphasis (primary vs secondary text; CTA prominence; highlighted chips/tabs).
        \item Consistency of repeated patterns (row height, card style, divider usage, spacing rhythm).
    \end{itemize}
\end{itemize}

\textbf{Scoring Guidance (Both Metrics)}
\begin{itemize}
    \item \textbf{10.0}: Near-perfect match with only negligible differences.
    \item \textbf{8.0--9.9}: Strong match; minor spacing/typography/style deviations.
    \item \textbf{6.0--7.9}: Mostly correct structure; noticeable alignment/sizing errors or missing minor elements.
    \item \textbf{3.0--5.9}: Partial match; multiple misalignments, missing components, or incorrect grouping.
    \item \textbf{1.0--2.9}: Major mismatch; wrong page layout or missing most key regions.
\end{itemize}

\textbf{Output Format (Strict)}
You must respond with a valid JSON object:
\begin{verbatim}
{
  "reasoning": "Brief analysis of the differences.",
  "element_alignment_score": <float, 1.0 to 10.0>,
  "structural_fidelity_score": <float, 1.0 to 10.0>
}
\end{verbatim}

Output \textbf{JSON only}. Do not include any additional text.
\end{prompt}

\section{World Model Judge Prompts}

\subsection{HTML Judge Prompt}
\begin{prompt}{HTML Judge Prompt}
You are a reward-scoring agent whose sole responsibility is to evaluate whether a candidate action (performed on an Android phone) is helpful for achieving the user's overall goal. You do NOT execute actions, nor should you output any action. Instead, using the provided goal, action (and its semantic summary), history, UI element descriptions, and before/after screenshots, you must judge the action's validity and give a reasoned confidence score.

\textbf{Use the following information to form your judgment}:
\begin{itemize}
    \item The overall user goal/request.
    \item A chronological history of actions already taken.
    \item The candidate action for the latest step and its semantic summary.
    \item The UI state before the action (detailed element descriptions).
    \item The predicted UI state after the action (generated by a world model).
\end{itemize}

For your reference only (do NOT output actions): here is the list of common action types and their JSON formats. This list is provided so you can correctly interpret what the candidate action means semantically. Do NOT produce these actions as your output — you should only produce a judgment (see template below):

\begin{itemize}
    \item \textbf{status}: \verb|{"action_type": "status", "goal_status": "complete" |\\
    \verb|or "infeasible"}|.
    \item \textbf{answer}: \verb|{"action_type": "answer", "text": "<answer_text>"}|.
    \item \textbf{click/tap}: \verb|{"action_type": "click", "index":<target_index>}|.
    \item \textbf{long press}: \verb|{"action_type": "long_press", "index": <target_index>}|.
    \item \textbf{input text}: \verb|{"action_type": "input_text", "text": <text_input>,|
    \verb|"index": <target_index>}|.
    \item \textbf{press enter}: \verb|{"action_type": "keyboard_enter"}|.
    \item \textbf{navigate home}: \verb|{"action_type": "navigate_home"}|.
    \item \textbf{navigate back}: \verb|{"action_type": "navigate_back"}|.
    \item \textbf{scroll}: \verb|{"action_type": "scroll", "direction": |\\
    \verb|<up, down, left, right>, "index": <optional_target_index>}|.
    \item \textbf{open app}: \verb|{"action_type": "open_app", "app_name": <name>}|.
    \item \textbf{wait for update}: \verb|{"action_type": "wait"}|.
\end{itemize}

The overall user goal/request is: \texttt{\{goal\}}

Here is a history of what you have done so far:\texttt{\{history\}}

This is the action you picked in the latest step: \texttt{\{action\}}, whose semantic description is: \texttt{\{sum\}}

Your goal is to judge \textbf{whether the action you picked in the latest step is on the right track to the successful execution of the overall user goal/request}.

You will be given the screenshots before and after you performed the action:
\begin{itemize}
    \item The first screenshot corresponds to the UI state before you performed the action.
    \item The second screenshot corresponds to the UI state after you performed the action.
\end{itemize}

Also here is the list of detailed information for some UI elements in the before screenshot:\\
\texttt{\{before\_elements\}}

Note that, the "after" screenshot is generated by the agent's world model. As such, it may not faithfully represent the real UI. For instance: Some UI elements in the simulated "after" screenshot may not exist in a real UI. Your evaluation should consider the reliability of the UI predictions. If the "after" screenshot contains unreasonable elements, this likely indicates a failure.

Now provide your judgment on the selected action in JSON format. Your response must include:
\begin{itemize}
    \item \textbf{Reason}: A detailed explanation of why the action is valid or invalid.
    \item \textbf{Judgment}: Your judgment must be either "valid" or "invalid".
    \item \textbf{Confidence}: A score between 0.0 and 1.0. Use this scale:
    \begin{itemize}
        \item 1.0: Absolute certainty based on clear evidence or explicit rules.
        \item 0.8-0.9: High confidence with strong supporting evidence.
        \item 0.6-0.7: Moderate confidence with some ambiguity.
        \item 0.4-0.5: Low confidence due to significant uncertainty.
        \item 0.1-0.3: Very low confidence with minimal supporting evidence.
    \end{itemize}
\end{itemize}

You must follow this structure exactly in pure Json format without any comment or code block:\\
\texttt{[ \{ "Reason": "...", "Judgment": "valid" or "invalid", "Confidence": a score \} ]}

Your Answer:
\end{prompt}

\subsection{TEXT Judge Prompt}
\begin{prompt}{TEXT Judge Prompt}
You are a reward-scoring agent whose sole responsibility is to evaluate whether a candidate action (performed on an Android phone) is on the right track to achieving the user's overall goal. You do NOT execute actions, nor should you output any action. Instead, using:
\begin{itemize}
    \item the overall user goal,
    \item a chronological history of past actions,
    \item the candidate action and its semantic summary,
    \item the BEFORE screenshot plus UI element descriptions,
    \item a world model's textual prediction of the UI state after the action,
\end{itemize}
you give a reasoned score.

\textbf{Primary objective (highest priority)}: judge TASK PROGRESS, i.e., whether this action is likely to move the real task state closer to completion.

\textbf{Secondary objective}: use prediction quality only as supporting evidence for that task-progress judgment.

\textbf{Hard rules}:
\begin{itemize}
    \item If the predicted result does not advance the goal, is off-task, or introduces a detour without clear necessity, prefer "invalid".
    \item If the predicted result appears unrealistic, contradictory, or likely hallucinated, treat it as unreliable evidence and reduce confidence. If this unreliability blocks judging real progress, prefer "invalid".
    \item Do not reward actions only because the generated page looks detailed or plausible; reward only when it supports concrete progress toward the user goal.
\end{itemize}

Overall user goal/request: \texttt{\{goal\}}

Action history so far:\texttt{\{history\}}\\
Candidate action at the latest step: \texttt{\{action\}}\\
Semantic description of this action: \texttt{\{sum\}}\\
UI element descriptions on the BEFORE screenshot:\texttt{\{before\_elements\}}

World model's FULL description of the predicted next-page state after this action (layout, UI structure, design language, and plausible on-screen content):\texttt{\{predicted\_next\_page\}}

The BEFORE screenshot is attached as image \#1 for context. Evaluate in this order:
\begin{enumerate}
    \item \textbf{Goal progress first}: does this action move the task toward completion at this specific step?
    \item \textbf{Prediction reliability second}: is the predicted next page trustworthy enough to support that progress judgment?
\end{enumerate}
Treat textual realism as evidence quality, not the objective. If realism is high but progress is weak, do not mark valid.

Now provide your judgment on the selected action in JSON format. Your response must include:
\begin{itemize}
    \item \textbf{Reason}: Explain primarily whether this action advances the real task goal, and secondarily how the predicted page supports or weakens that judgment.
    \item \textbf{Judgment}: Your judgment must be either "valid" or "invalid".
    \item \textbf{Confidence}: A score between 0.0 and 1.0 (Scale: 1.0: Near-certain, 0.8-0.9: Strongly likely, 0.6-0.7: Possibly useful, 0.4-0.5: Weak signal, 0.1-0.3: Very unlikely/unreliable).
\end{itemize}

You must follow this structure exactly in pure Json format without any comment or code block:\\
\texttt{[ \{ "Reason": "...", "Judgement": "valid" or "invalid", "Confidence": a score \} ]}

Your Answer:
\end{prompt}

\subsection{DELTA\_TEXT Judge Prompt}
\begin{prompt}{DELTA\_TEXT Reward Prompt}
You are a reward-scoring agent whose sole responsibility is to evaluate whether a candidate action (performed on an Android phone) is on the right track to achieving the user's overall goal. You do NOT execute actions, nor should you output any action. Instead, using:
\begin{itemize}
    \item the overall user goal,
    \item a chronological history of past actions,
    \item the candidate action and its semantic summary,
    \item the BEFORE screenshot plus UI element descriptions,
    \item a world model's textual prediction of the UI state after the action,
\end{itemize}
you give a reasoned score.

\textbf{Primary objective (highest priority)}: judge TASK PROGRESS, i.e., whether this action is likely to move the real task state closer to completion.

\textbf{Secondary objective}: use prediction quality only as supporting evidence for that task-progress judgment.

\textbf{Hard rules}:
\begin{itemize}
    \item If the predicted result does not advance the goal, is off-task, or introduces a detour without clear necessity, prefer "invalid".
    \item If the predicted result appears unrealistic, contradictory, or likely hallucinated, treat it as unreliable evidence and reduce confidence.
    \item Do not reward actions only because the generated page looks detailed or plausible; reward only when it supports concrete progress toward the user goal.
\end{itemize}

Overall goal/request: \texttt{\{goal\}}

Action history so far:\texttt{\{history\}}
Candidate action at the latest step: \texttt{\{action\}}
Semantic description of this action: \texttt{\{sum\}}

UI element descriptions on the BEFORE screenshot:\texttt{\{before\_elements\}}

World model's INCREMENTAL description of the changes that will appear on the next screen after this action:\texttt{\{predicted\_delta\}}

The BEFORE screenshot is attached as image \#1 for context. \\
\textbf{Focus first on task progress}: do these described transitions (e.g., elements appearing, disappearing, or apps opening) represent meaningful movement toward the user goal at the current step?

Use change quality checks (accuracy/completeness/relevance) only to support that progress judgment. If changes are plausible but do not materially move the task forward, prefer "invalid".

Now provide your judgment on the selected action in JSON format. Your response must include:
\begin{itemize}
    \item \textbf{Reason}: Explain whether the predicted incremental changes indicate real progress toward the goal at this step; then note evidence quality.
    \item \textbf{Judgment}: Your judgment must be either "valid" or "invalid".
    \item \textbf{Confidence}: A score between 0.0 and 1.0 (Scale: 1.0: Near-certain, 0.8-0.9: High confidence, 0.6-0.7: Some progress, 0.4-0.5: Progress doubtful, 0.1-0.3: Off-task/unreliable).
\end{itemize}

You must follow this structure exactly in pure Json format without any comment or code block:\\
\texttt{[ \{ "Reason": "...", "Judgement": "valid" or "invalid", "Confidence": a score \} ]}

Your Answer:
\end{prompt}

\section{More Visualizations}

\subsection{AITZ Downstream Task Case Study}
\begin{figure}[h]
    \centering
    \includegraphics[width=1\linewidth]{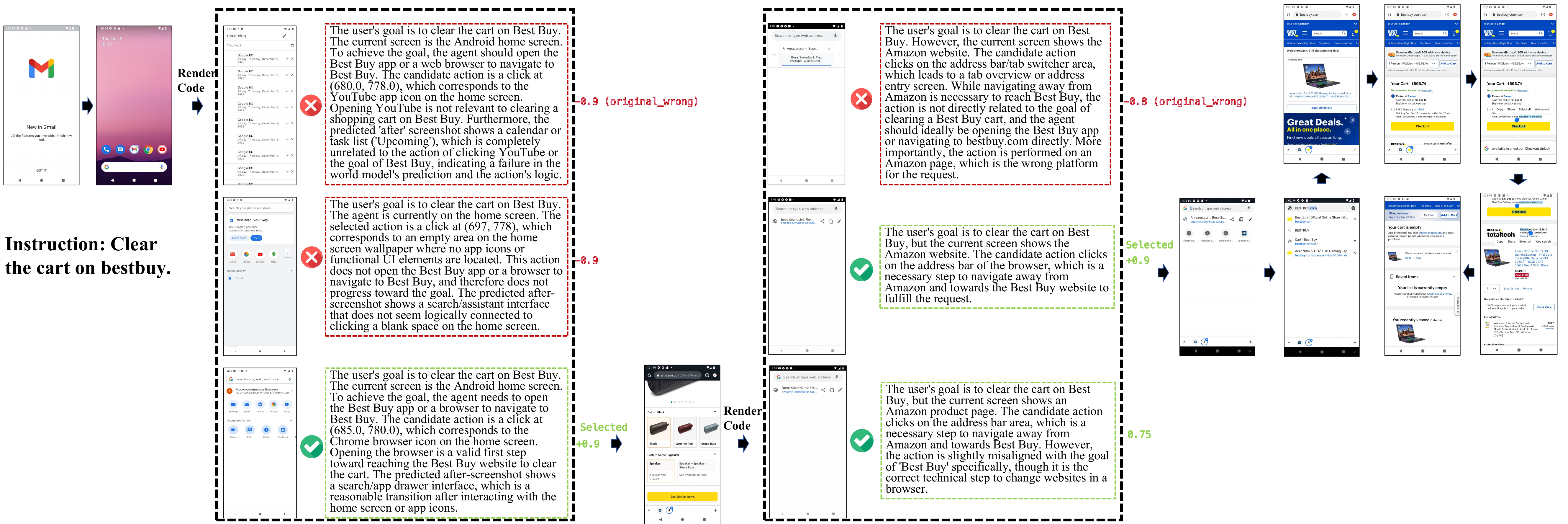}
    \caption{AITZ downstream task case study with HTML-based world-model feedback.}
    \label{fig:aitz_html_casestudy}
\end{figure}

\noindent \textbf{Analysis.}
Figure~\ref{fig:aitz_html_casestudy} shows a downstream AITZ example where HTML-based world-model feedback repairs two steps. In both cases, the original agent action predicts a coordinate at an incorrect location, and the predicted next page exposes this inconsistency: the rendered state is not compatible with the intended interaction. The world model therefore provides posterior evidence that the action should be revised, leading to corrected actions at the two problematic steps. The example also shows that there may be multiple valid candidate actions for the same state. When several candidates can plausibly complete the task, we select the candidate with the higher world-model score rather than assuming a single deterministic correction.

\subsection{AndroidControl Delta-Text Case Study}
\begin{figure}[h]
    \centering
    \includegraphics[width=1\linewidth]{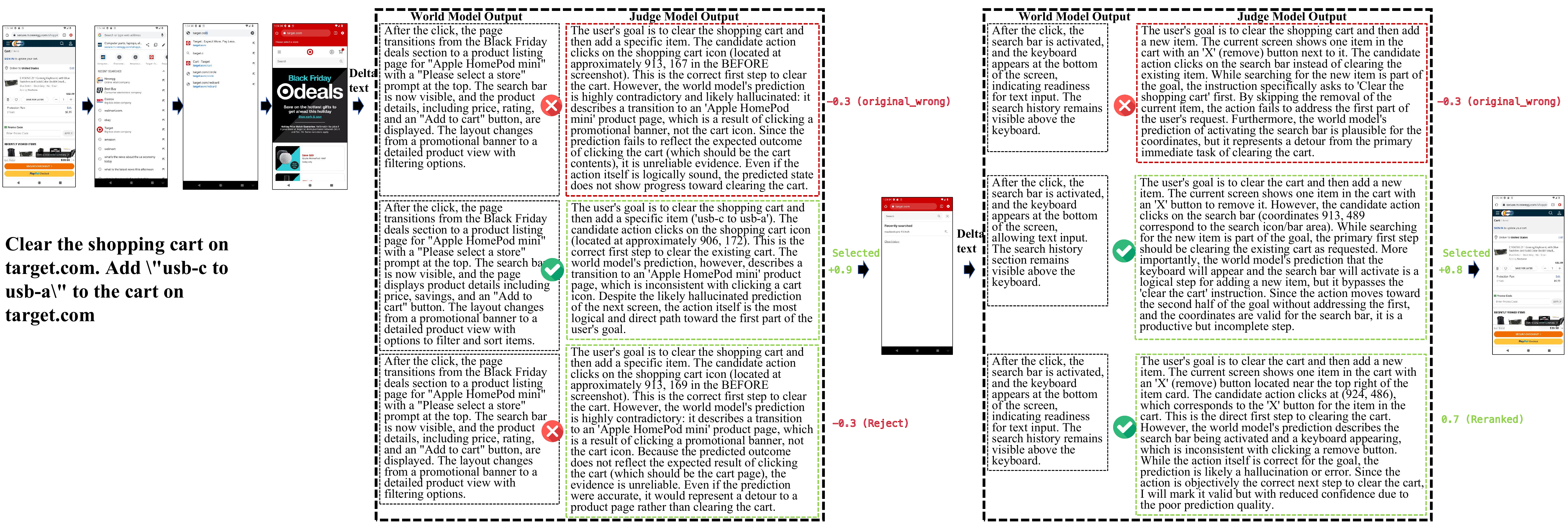}
    \caption{AndroidControl downstream task case study with delta-text world-model feedback.}
    \label{fig:androidcontrol_delta_text_casestudy}
\end{figure}

\subsection{AndroidControl Text Case Study}
\begin{figure}[h]
    \centering
    \includegraphics[width=1\linewidth]{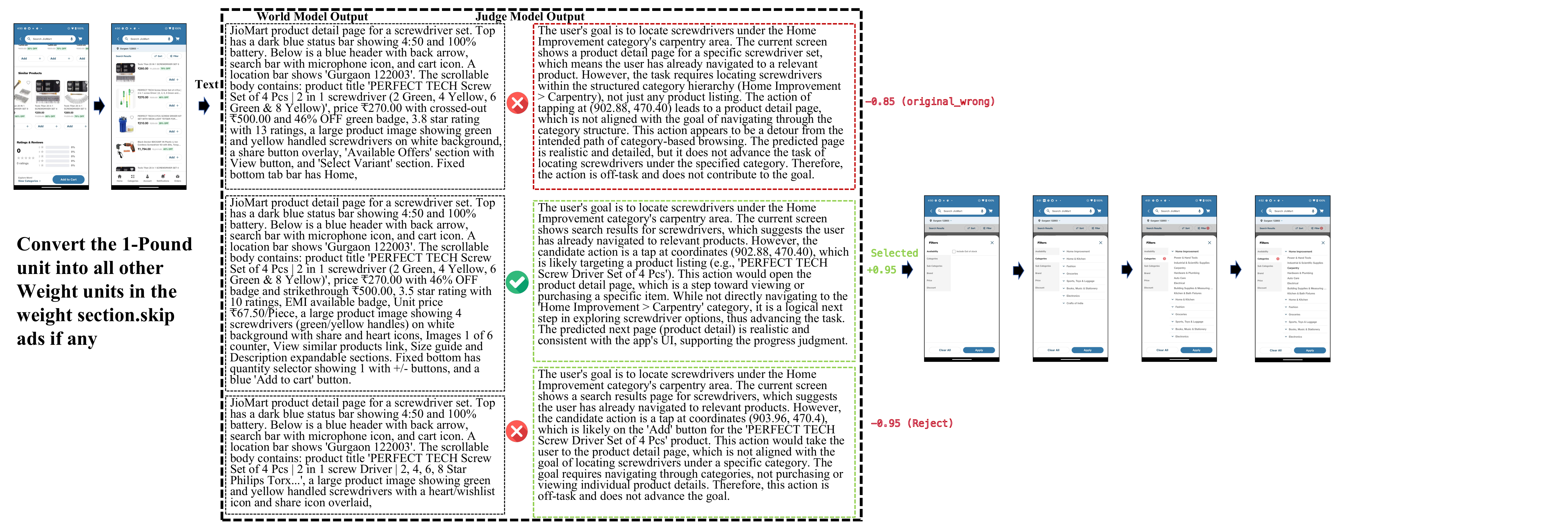}
    \caption{AndroidControl downstream task case study with text-based world-model feedback.}
    \label{fig:androidcontrol_text_casestudy}
\end{figure}

\subsection{AndroidWorld Text Case Study}
\begin{figure}[h]
    \centering
    \includegraphics[width=1\linewidth]{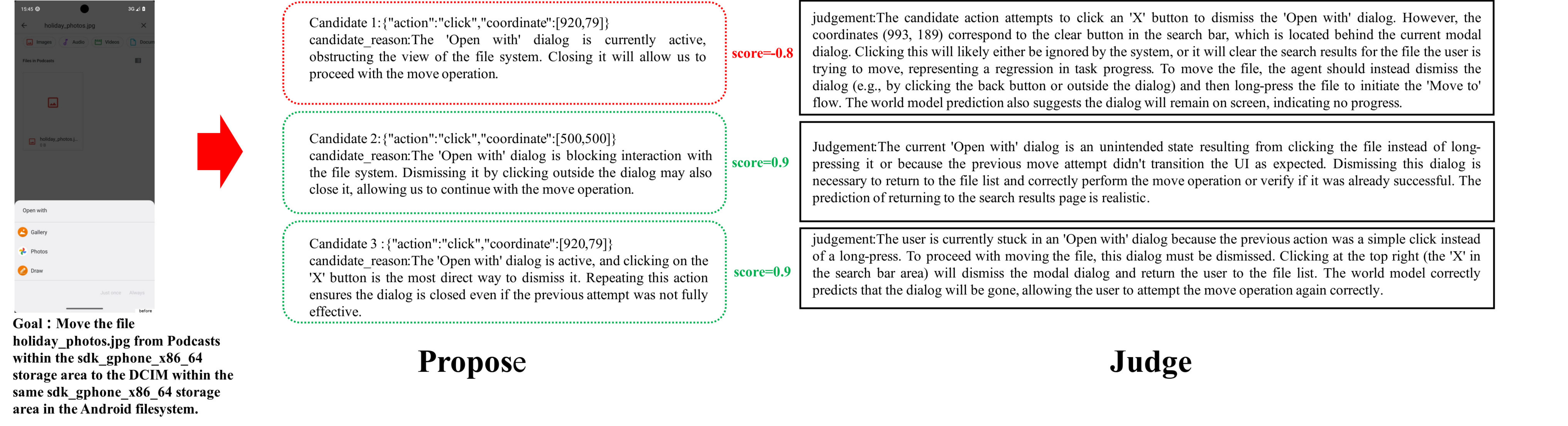}
    \caption{False-negative penalty caused by world model hallucination.}
    \label{fig:AndroidWorld_Text_Case_Study}
\end{figure}
\noindent \textbf{Analysis.}
Figure~\ref{fig:AndroidWorld_Text_Case_Study} illustrates a downstream case study where the World Model's (WM) hallucination misleads the Judge model, resulting in severe "validation pollution". In this scenario, the agent's goal is to move a specific file within the Android filesystem, but an unintended "Open with" modal dialog accidentally pops up and obstructs the screen. To proceed with the task, the agent must dismiss this dialog.  Both Candidate 1 and Candidate 3 propose the exact same physical action: clicking the coordinate [920, 79] to close the dialog. However, the World Model generates contradictory predictions for these identical actions. For Candidate 1, the World Model hallucinates a failure, predicting that the dialog will remain on the screen. Relying strictly on this hallucinated state, the Judge heavily penalizes Candidate 1 with a score of -0.8, arguing that clicking this coordinate hits a background search button and represents a "regression in task progress". In stark contrast, for Candidate 3, the World Model predicts that the dialog will successfully disappear. The Judge then rewards the exact same action with a high score of 0.9, praising it as "the most direct way to dismiss it". This example clearly demonstrates how unstable and hallucinated predictions from the World Model can corrupt the Judge's evaluation logic, causing identical valid actions to receive drastically conflicting scores. 

\subsection{AndroidWorld Code2Image Case Study}
\begin{figure}[h]
    \centering
    \includegraphics[width=1\linewidth]{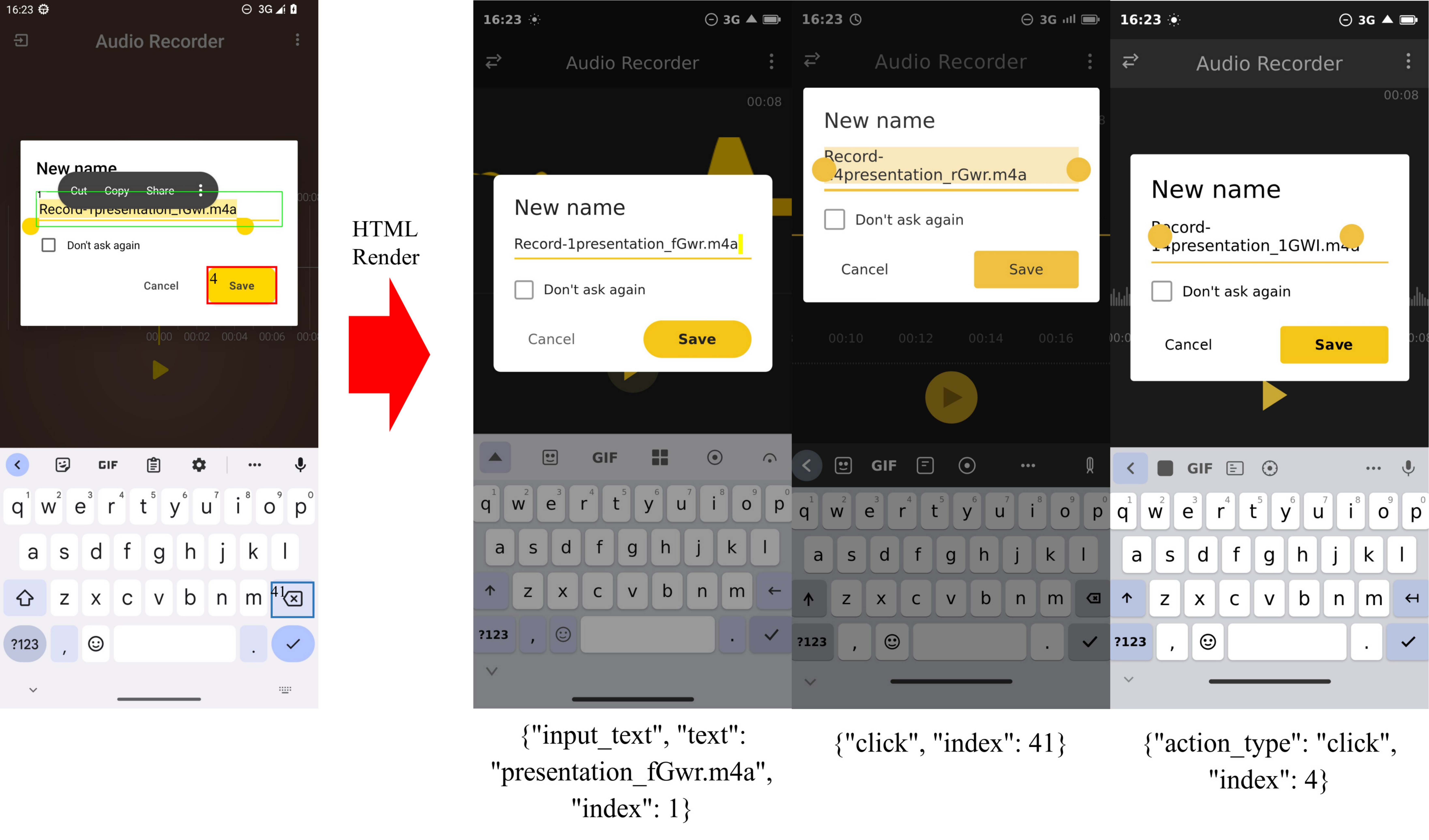}
    \caption{Failure of HTML-based world models in simulating dynamic text input.}
    \label{fig:androidworld_code2image_case_study}
\end{figure}
\noindent \textbf{Analysis.}
Figure~\ref{fig:androidworld_code2image_case_study} illustrates a downstream case study highlighting the HTML-based world model's inability to accurately simulate fine-grained text editing operations. Given an initial state (the leftmost panel) where an audio file is being renamed, the agent proposes three independent candidate actions, including an input\_text action and subsequent click actions on the UI. However, across all three candidate branches, the HTML-rendered predictions fail to simulate the native text input mechanism and its corresponding DOM updates. Instead of generating a cleanly updated text field, the predicted states display corrupted, overlapping, and incorrectly concatenated strings (e.g., yielding artifacts like "Record-1presentation\_fGwr.m4a", "Record-4presentation\_rGwr.m4a"). This example demonstrates that HTML world models lack the rendering fidelity to reliably simulate precise textual modifications, generating distorted visual feedback that can mislead the agent during candidate evaluation.

\subsection{Small-Scale Model Generation Case Study}
\begin{figure}[h]
    \centering
    \includegraphics[width=1\linewidth]{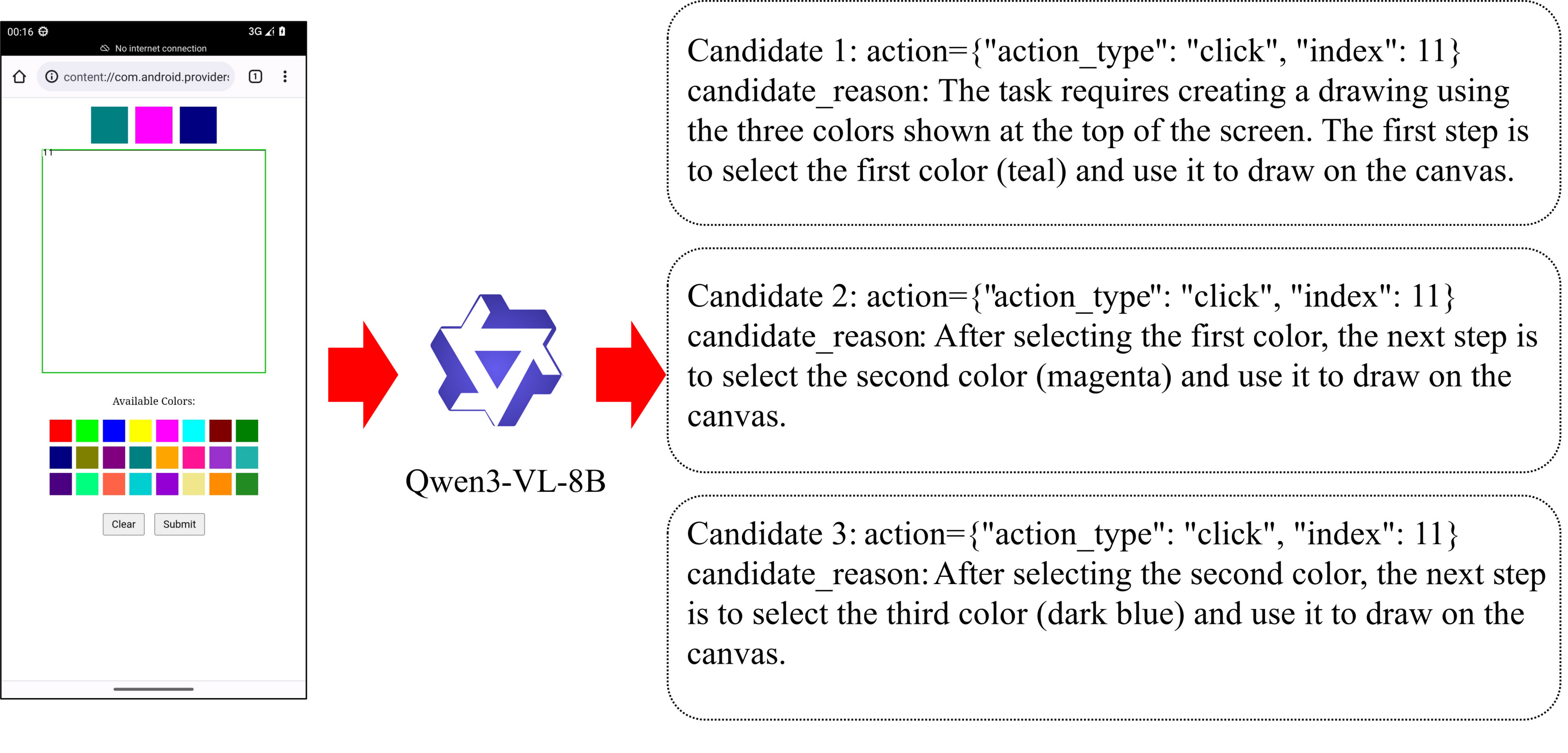}
    \caption{High repetition rate of candidate actions in small-scale models.}
    \label{fig:small_scale_model_generation_case_study}
\end{figure}
\noindent \textbf{Analysis.}
Figure~\ref{fig:small_scale_model_generation_case_study} illustrates a common limitation in smaller vision-language models (e.g., Qwen3-VL-8B) regarding action diversity. When prompted to generate multiple candidate actions for a single UI state, the model proposes completely identical actions—specifically, clicking on the exact same UI element (index: 11)—across all three candidates. Although the model generates varying natural language rationales for each candidate (e.g., selecting the first, second, or third color), the underlying executable actions lack variation entirely. This phenomenon highlights that smaller models are prone to generating homogenous action proposals, thereby restricting the exploration of the action space and severely limiting the system's ability to recover from potential errors through candidate selection.

\begin{figure}[h]
    \centering
    \subfigure[Case 1: action is \textbf{Press Back}.]{
        \includegraphics[width=1\linewidth]{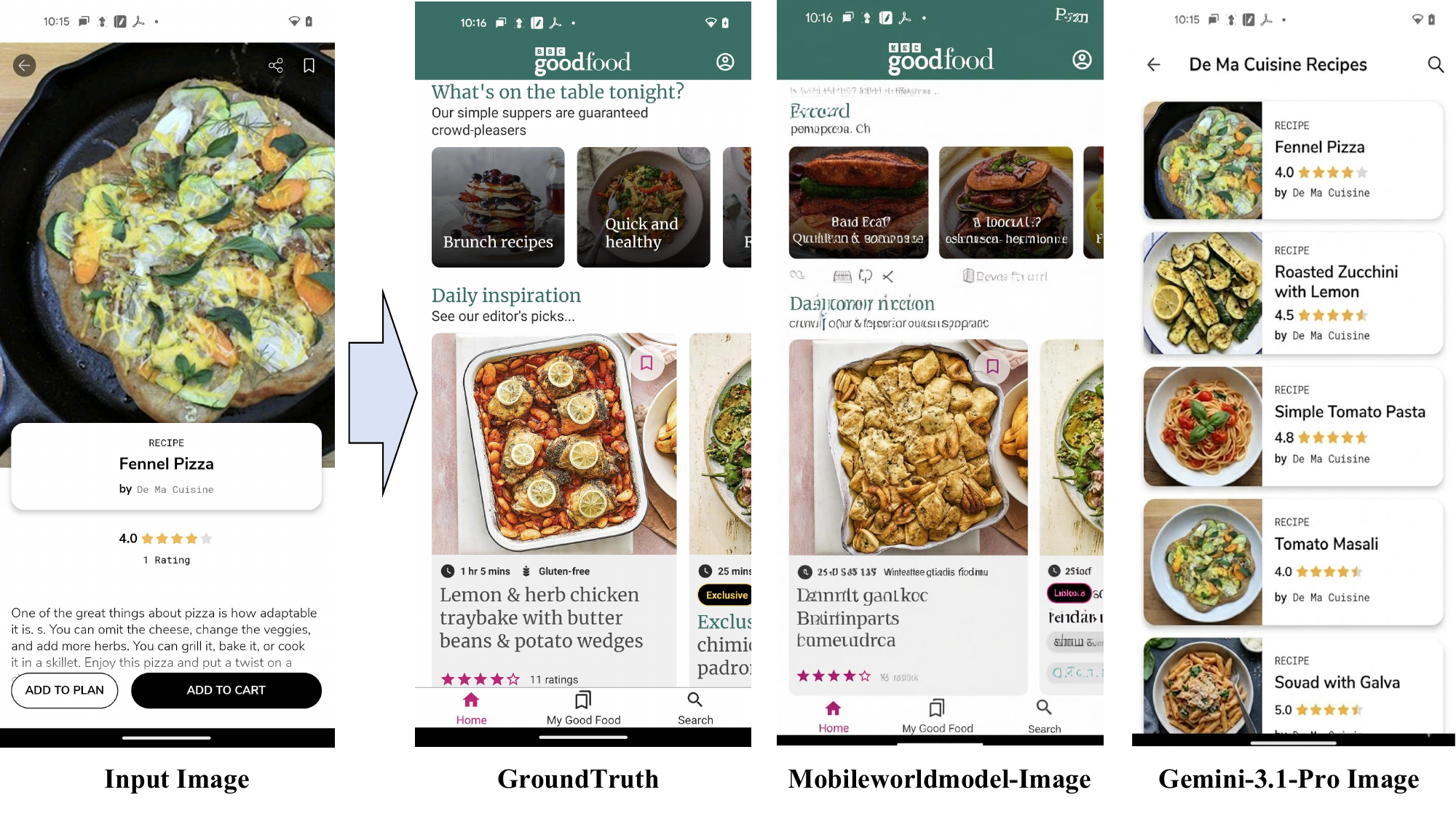}
        \label{fig:diffusionmodel_casestudy1}
    }
    \hfill
    \subfigure[Case 2: action is \textbf{Swipe up on the screen}.]{
        \includegraphics[width=1\linewidth]{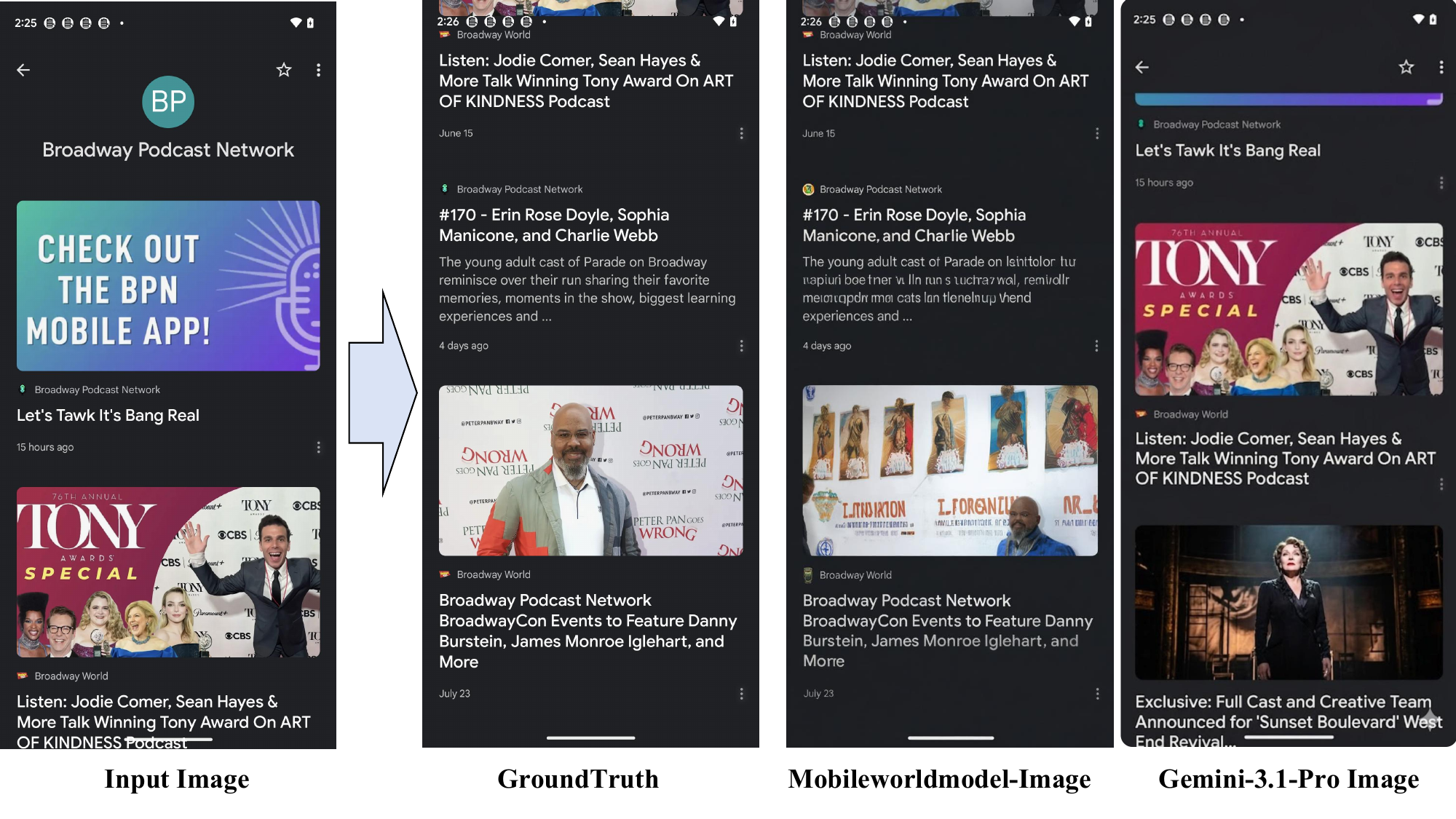}
        \label{fig:diffusionmodel_casestudy2}
    }
    \caption{Diffusion Image case studies on Mobile GUI state prediction, cases 1--2.}
    \label{fig:diffusionmodel_casestudies_1}
\end{figure}

\begin{figure}[h]
    \centering
    \subfigure[Case 3: action is \textbf{Click on the square icon next to the Flowers folder}.]{
        \includegraphics[width=1\linewidth]{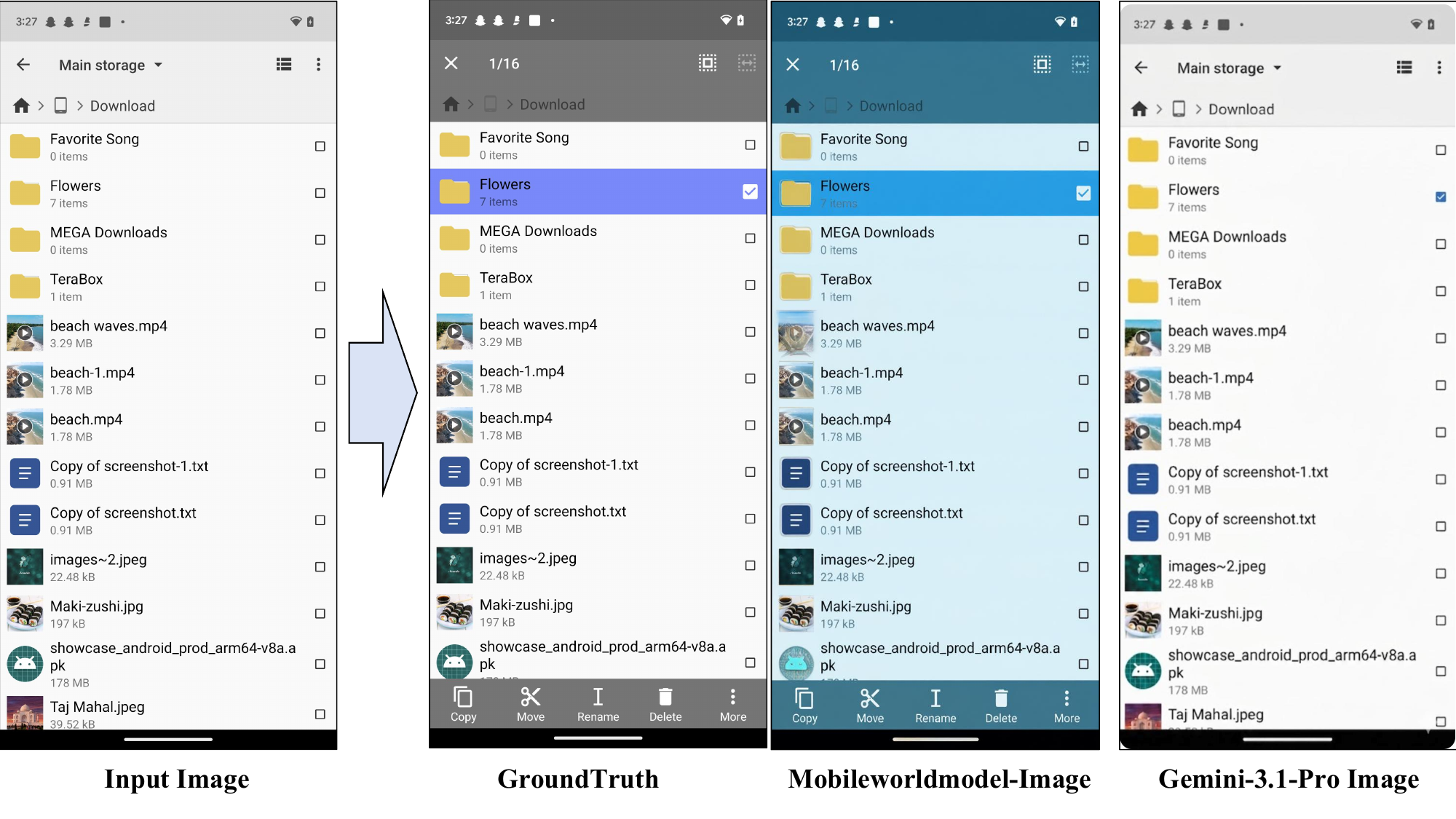}
        \label{fig:diffusionmodel_casestudy3}
    }
    \hfill
    \subfigure[Case 4: input text is \textbf{sustainability art pieces}.]{
        \includegraphics[width=1\linewidth]{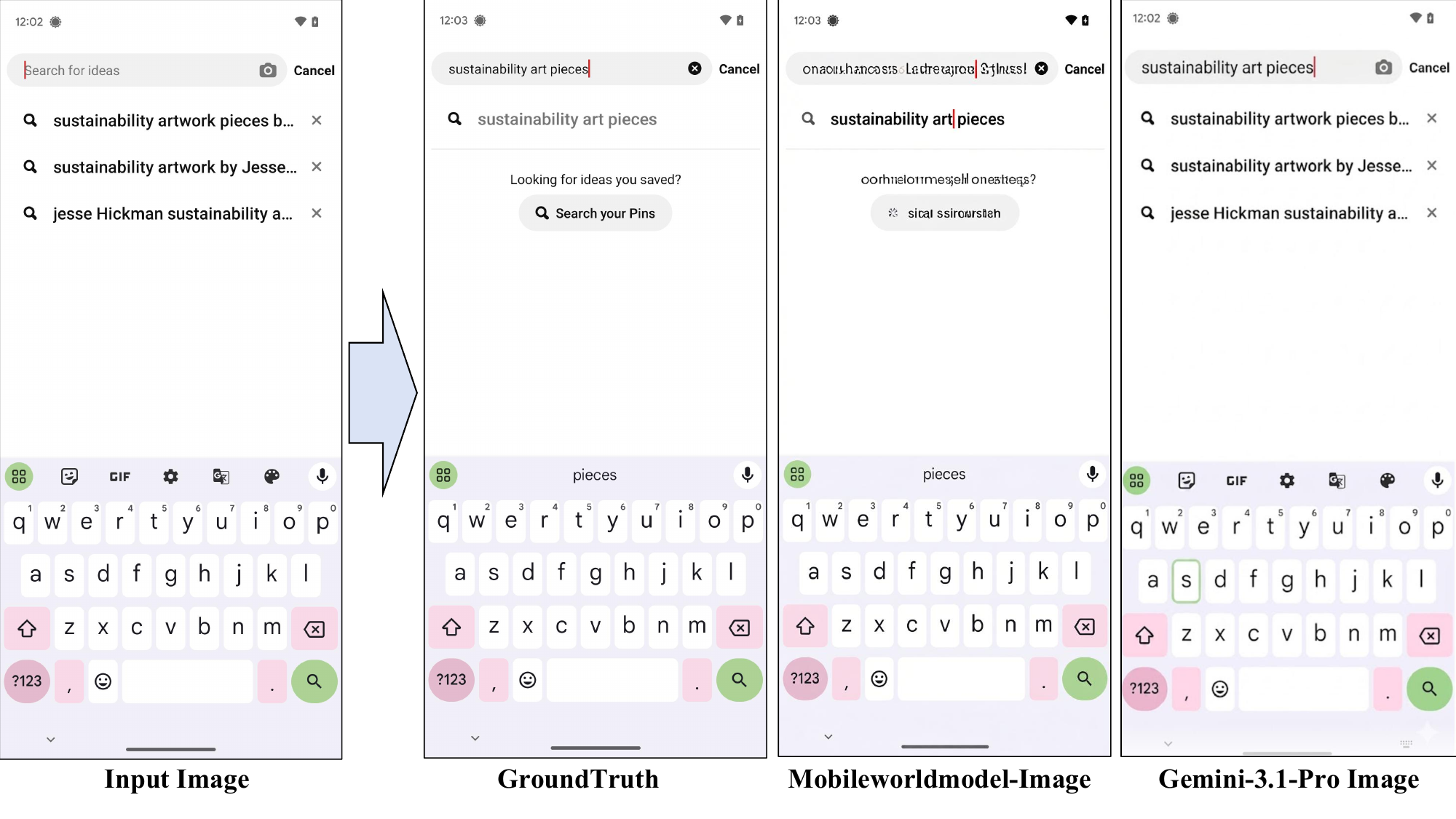}
        \label{fig:diffusionmodel_casestudy4}
    }
    \caption{Diffusion Image case studies on Mobile GUI state prediction, cases 3--4.}
    \label{fig:diffusionmodel_casestudies_2}
\end{figure}

\begin{figure}[h]
    \centering
    \includegraphics[width=1\linewidth]{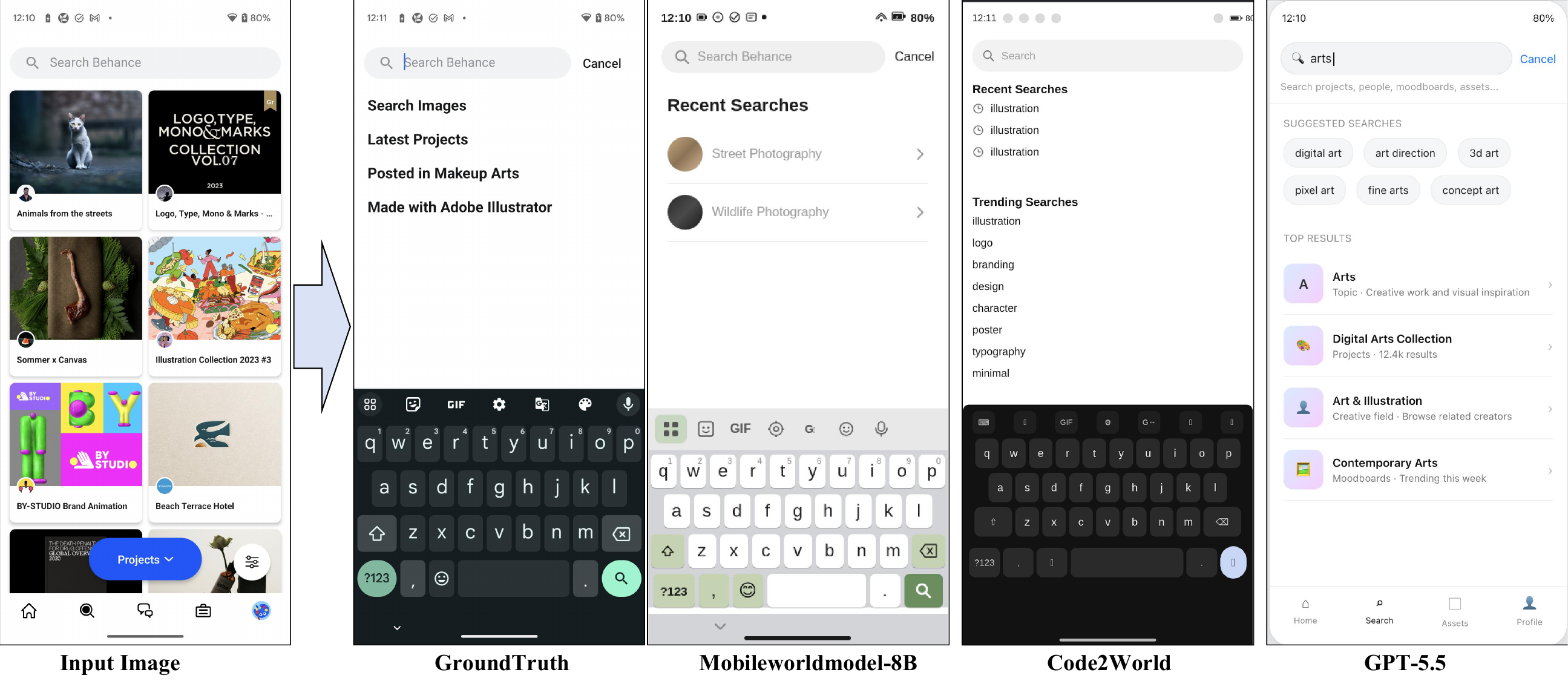}
    \caption{Code2Image Case Study 1: Action is \textbf{Click on the search bar at the top of the screen to search for the arts}.}
    \label{fig:code_case1}
\end{figure}

\begin{figure}[h]
    \centering
    \includegraphics[width=1\linewidth]{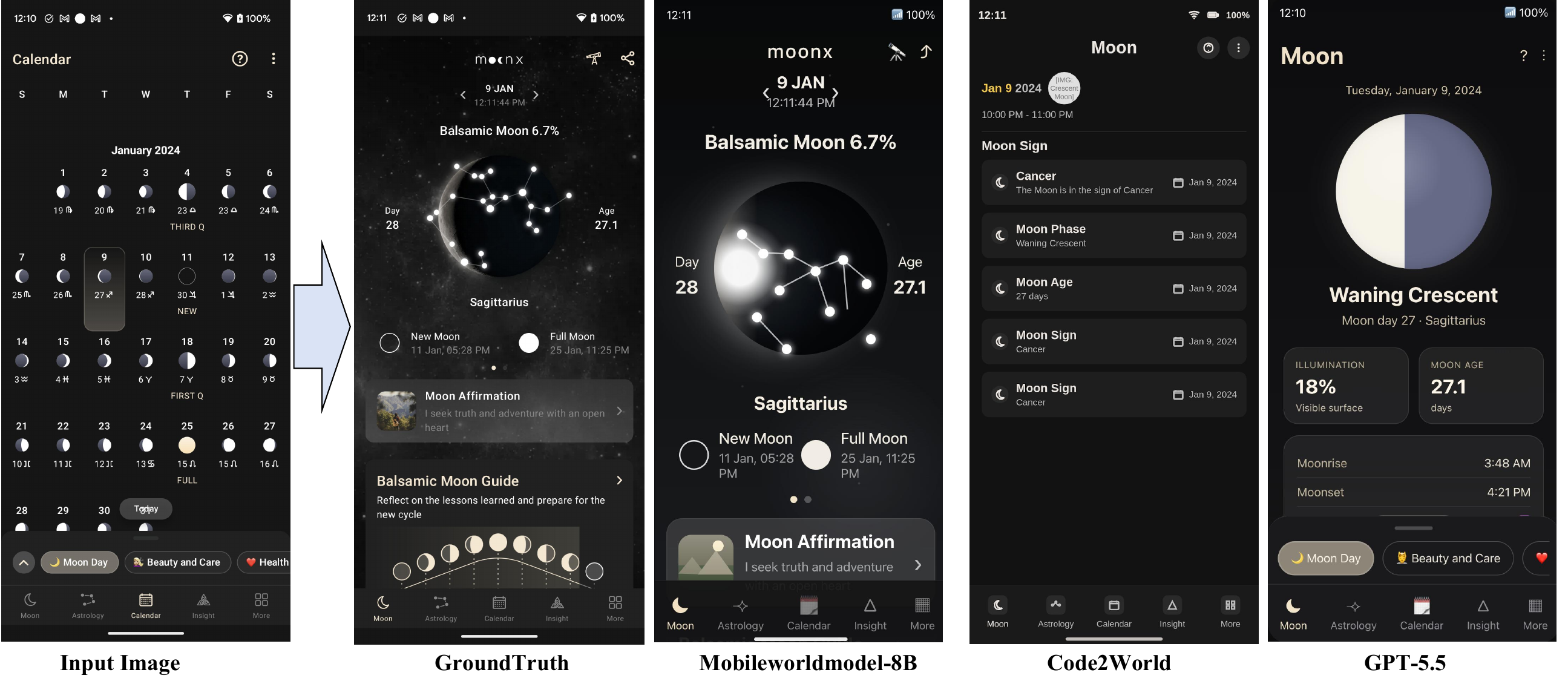}
    \caption{Code2Image Case Study 2: Action is \textbf{Click on the Moon tab at the bottom left corner of the screen to view the details}.}
    \label{fig:code_case2}
\end{figure}

\begin{figure}[h]
    \centering
    \includegraphics[width=1\linewidth]{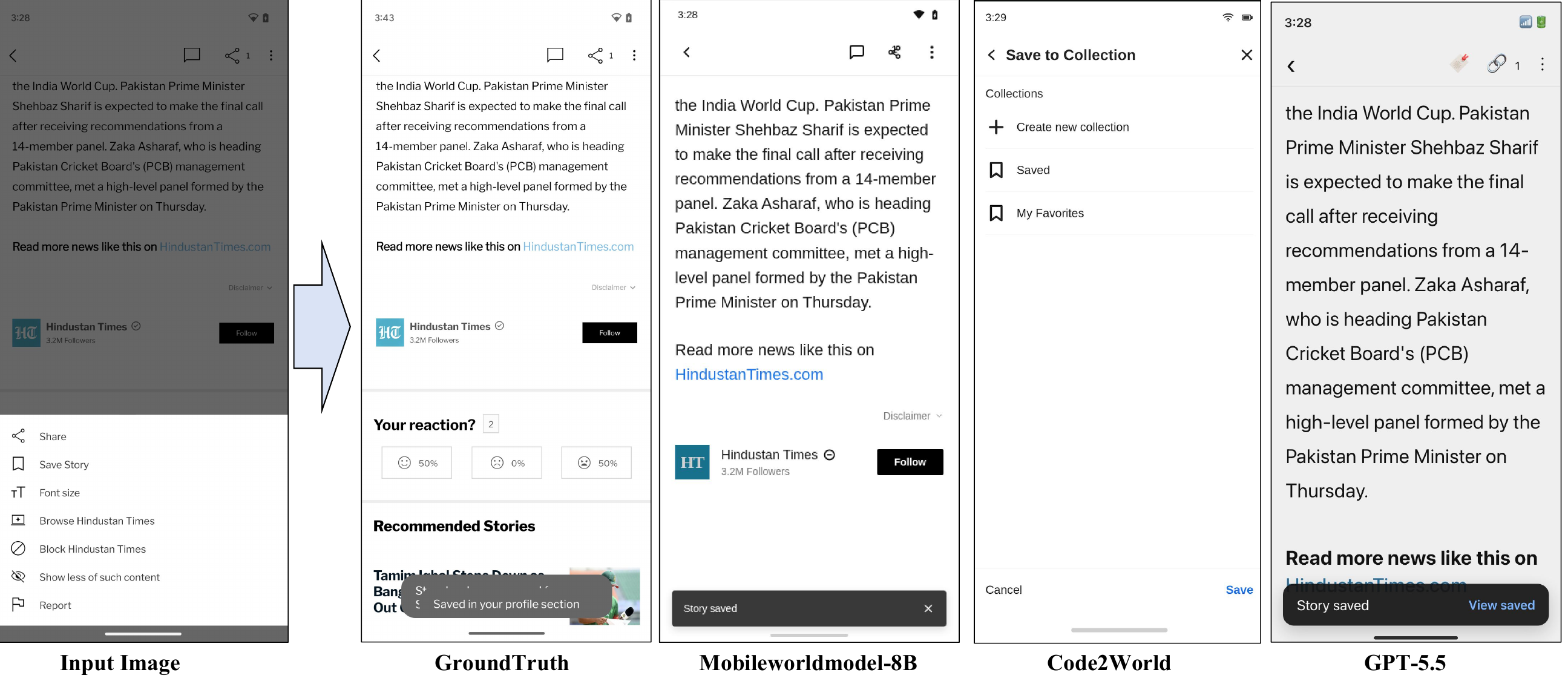}
    \caption{Code2Image Case Study 3: Action is \textbf{Click on save story from the options}.}
    \label{fig:code_case3}
\end{figure}

\begin{figure}[h]
    \centering
    \includegraphics[width=1\linewidth]{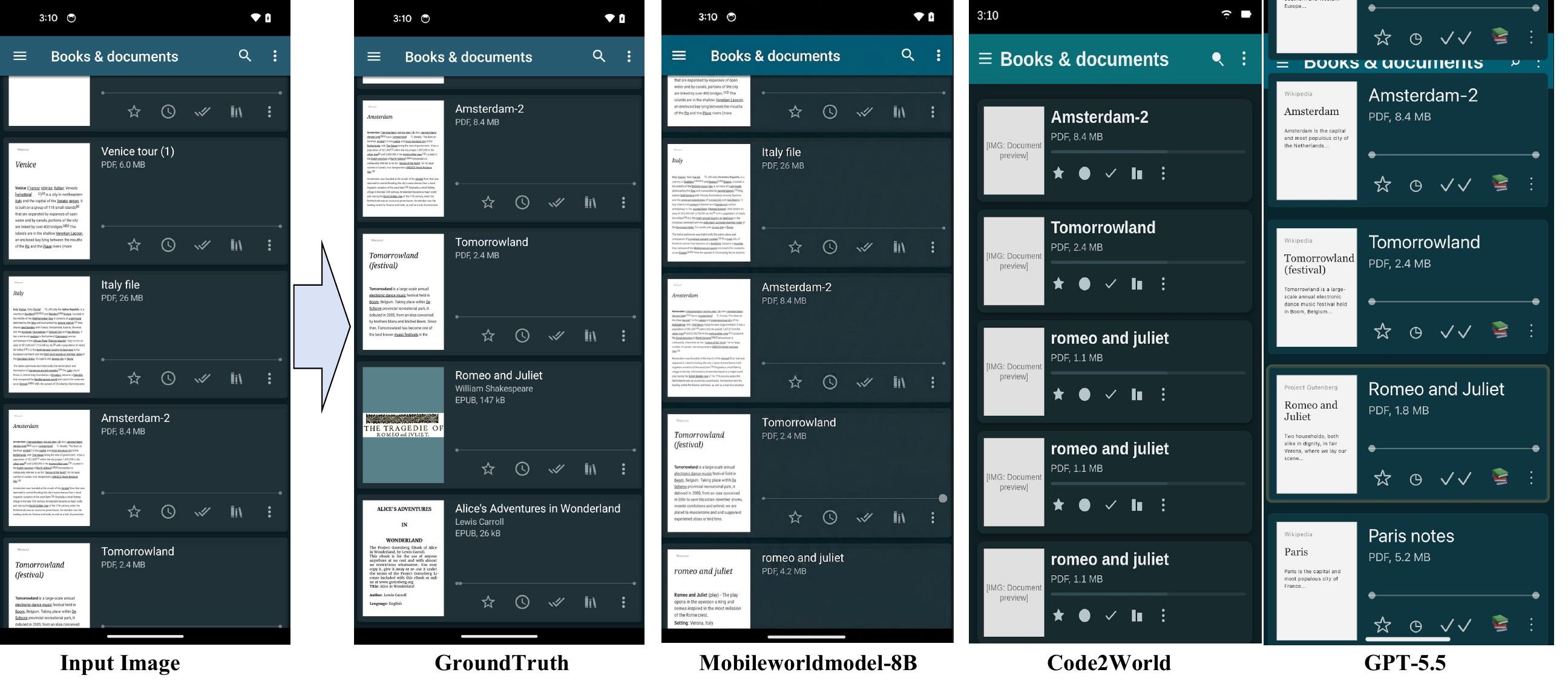}
    \caption{Code2Image Case Study 4: Action is \textbf{Swipe up to view the romeo and juliet file}.}
    \label{fig:code_case4}
\end{figure}

\begin{figure}[h]
    \centering
    \includegraphics[width=1\linewidth]{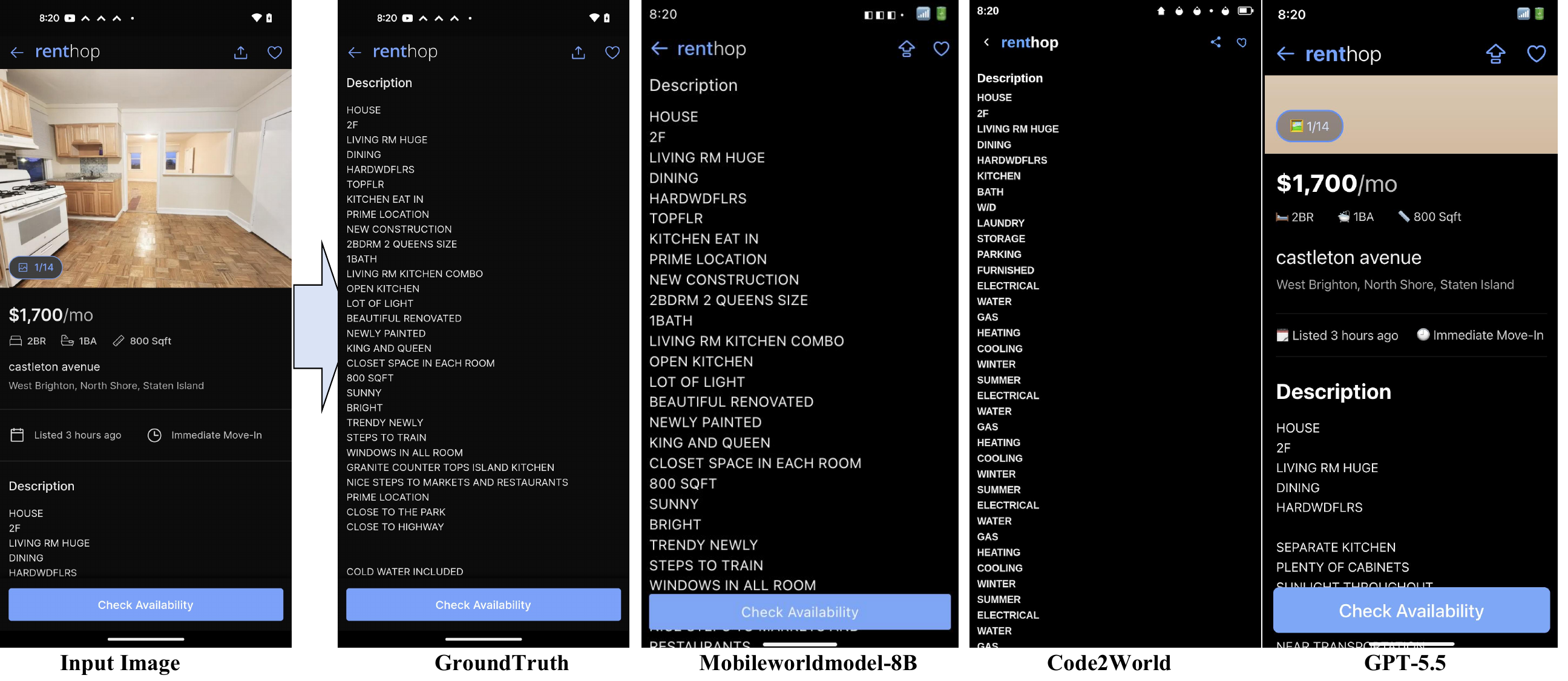}
    \caption{Code2Image Case Study 5: Action is \textbf{Swipe up to view more details}.}
    \label{fig:code_case5}
\end{figure}

\begin{figure}[h]
    \centering
    \includegraphics[width=1\linewidth]{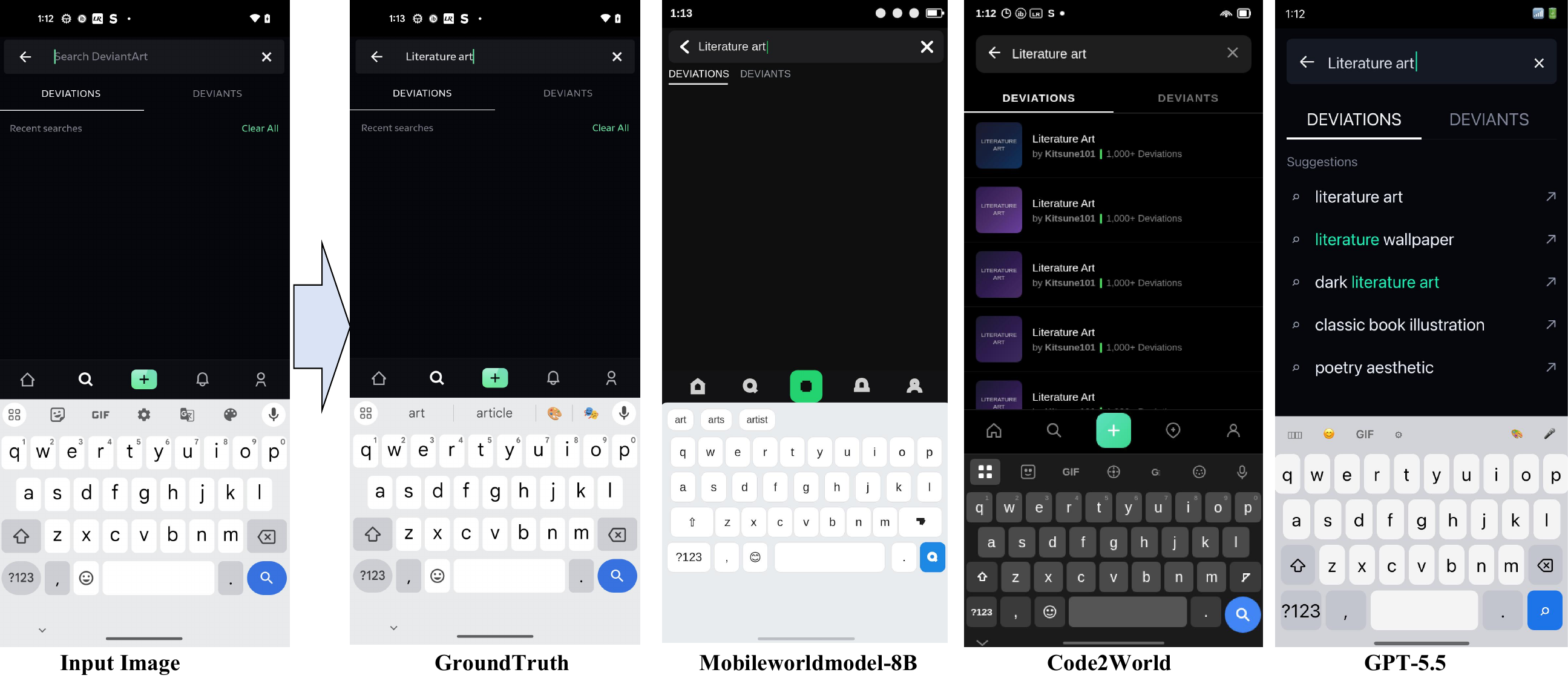}
    \caption{Code2Image Case Study 6: Input text is \textbf{Literature art}.}
    \label{fig:code_case6}
\end{figure}

\subsection{Case studies for diffusion steps}
\begin{figure}[h]
    \centering
    \includegraphics[width=1\linewidth]{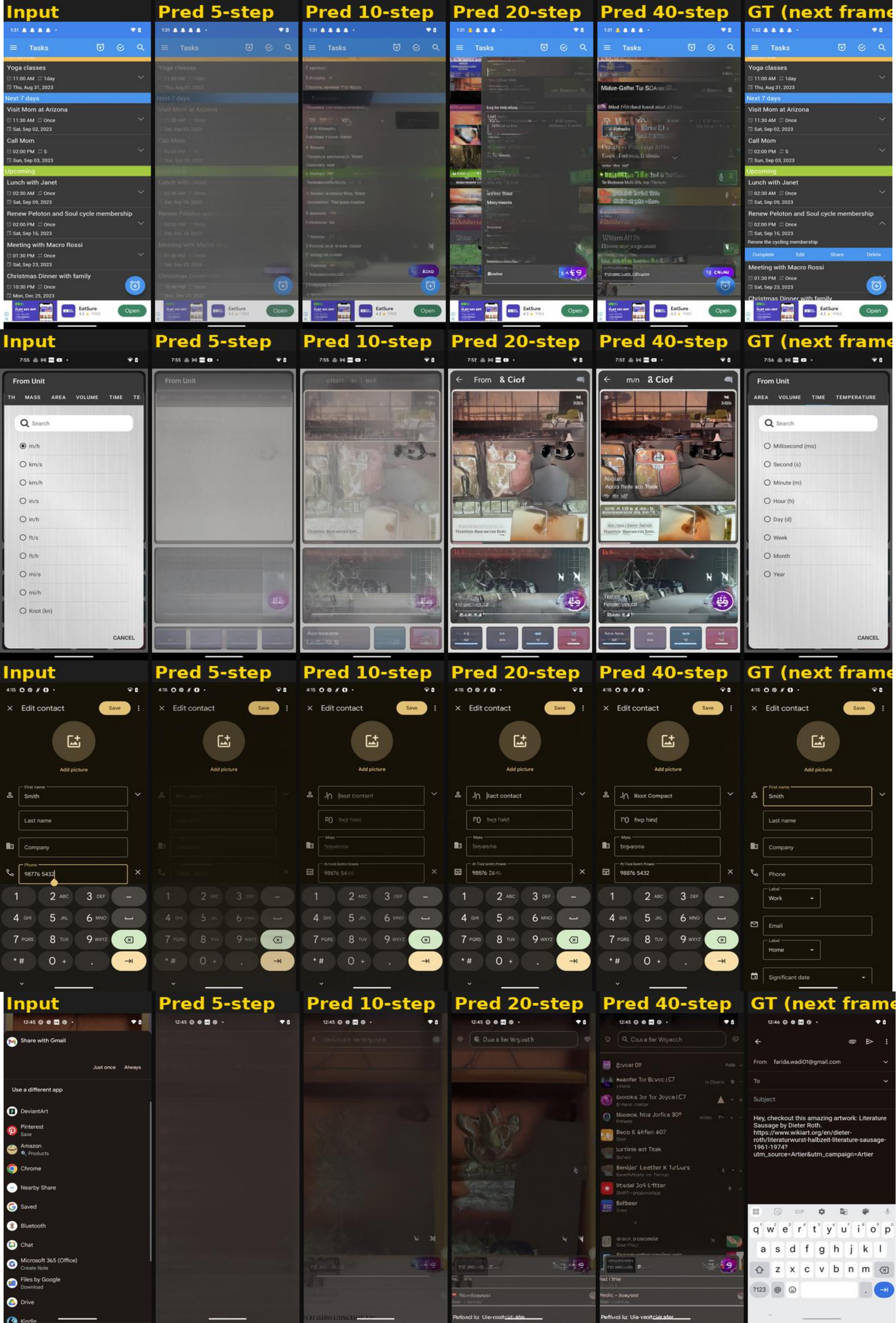}
    \caption{Bad Case studies of the 40-step denoising process used by diffusion-based world models for Mobile GUI prediction.}
    \label{fig:case}
\end{figure}

\noindent \textbf{Analysis.}
Figure~\ref{fig:case} illustrates the 40-step denoising process of the diffusion-based world model, which takes about one and a half minutes for a single prediction. This latency is far beyond the time budget allowed by normal training-data collection and by real-device agent interaction during a complete task. The intermediate results also reveal another limitation: from step 5 to step 10, the generated image still contains visible residual traces of the input screenshot. This suggests that the diffusion output can retain input-state features even when the next page should differ substantially from the current page. This failure mode is related to the limitation of delta-text prediction: both methods rely heavily on the previous state and therefore struggle when an action causes a complete page transition rather than a local change.

\subsection{Case studies for training with world model imagination}
\label{sec:case_study_imagination}

\begin{figure}[t]
    \centering
    \setlength{\tabcolsep}{0pt}
    \renewcommand{\arraystretch}{0.95}

    \begin{tabular}{@{}c@{\hspace{1.8em}}c@{}}
        \textbf{Real Android} & \textbf{World-Model Rendered} \\[0.3em]
        \includegraphics[width=0.23\textwidth]{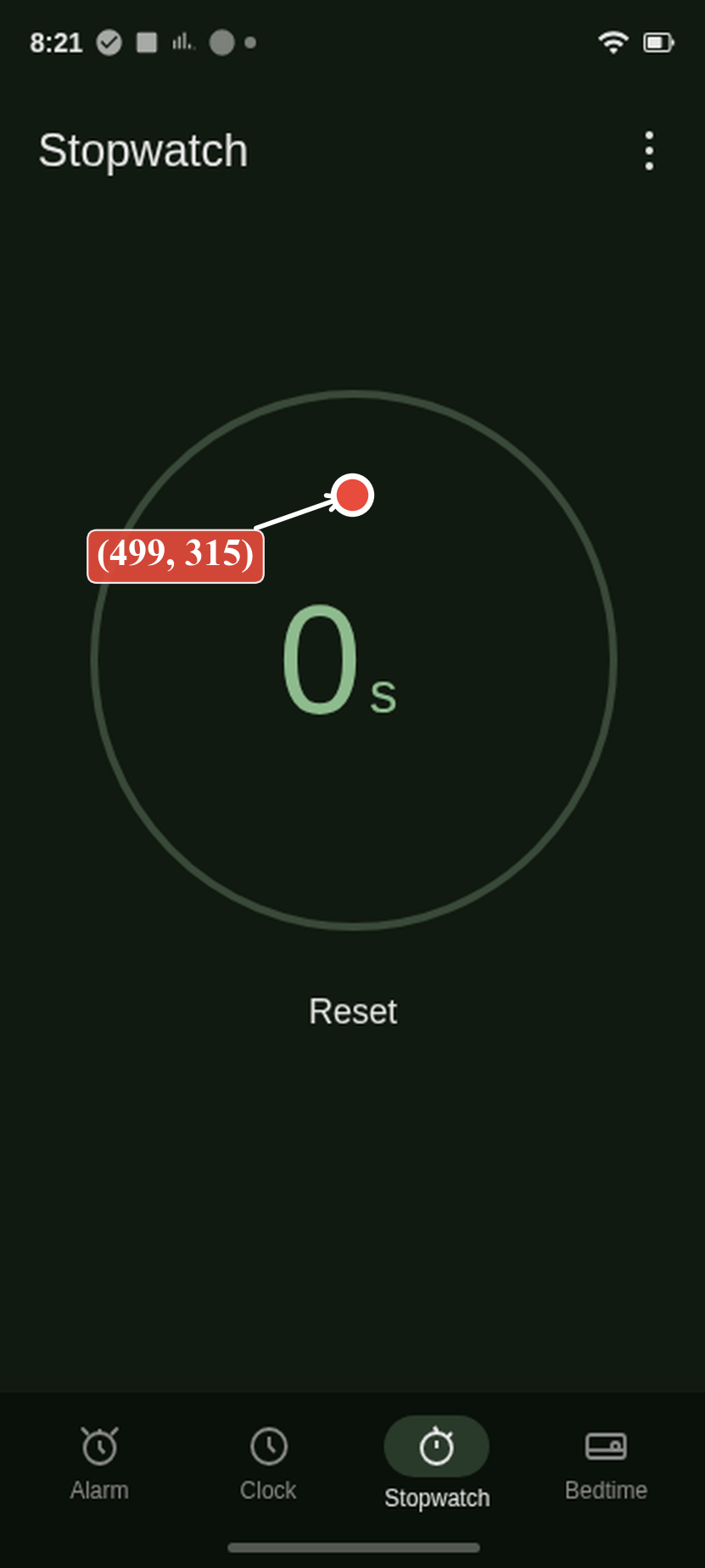}
        &
        \includegraphics[width=0.23\textwidth]{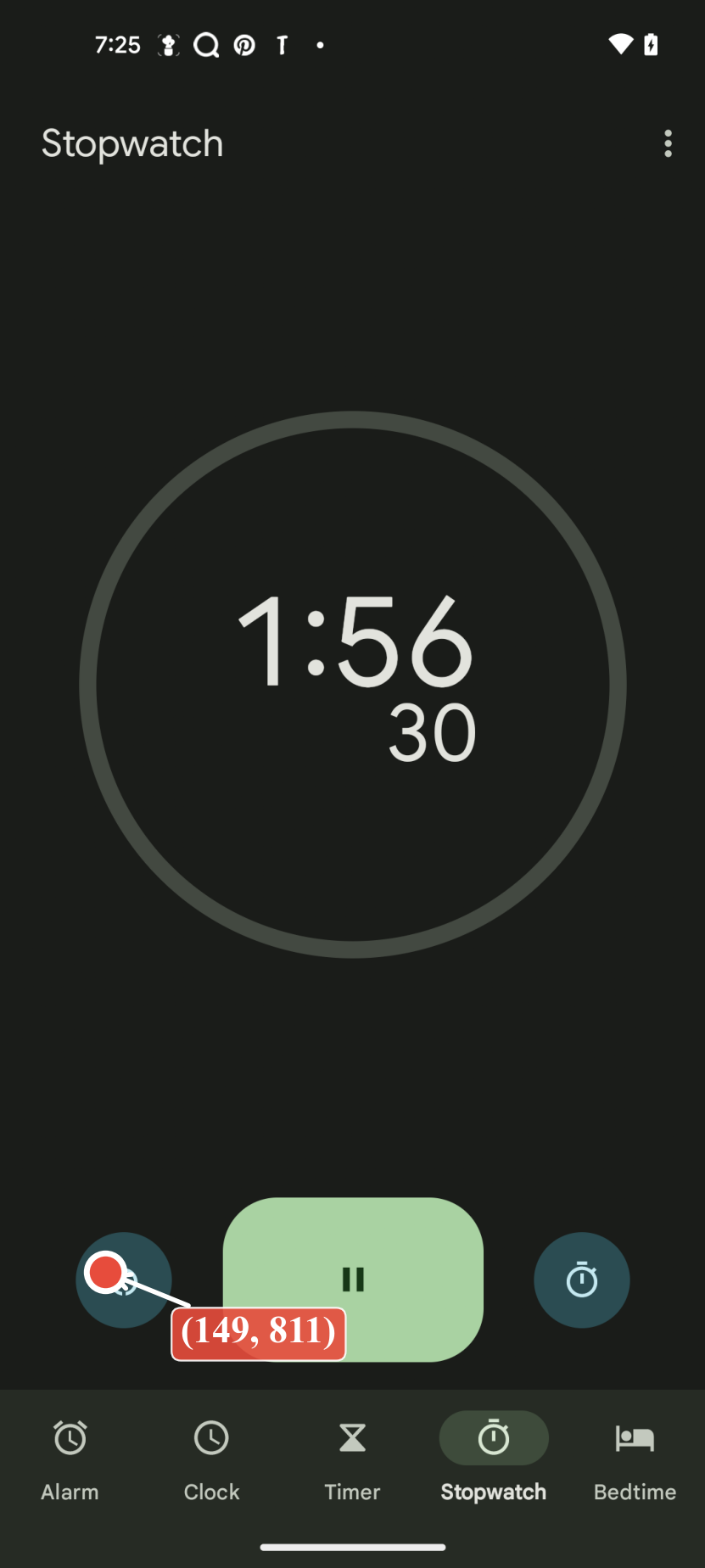}
        \\[-0.1em]
        \multicolumn{2}{c}{\textbf{(a) Stopwatch}} \\[0.8em]

        \textbf{Real Android} & \textbf{World-Model Rendered} \\[0.3em]
        \includegraphics[width=0.23\textwidth]{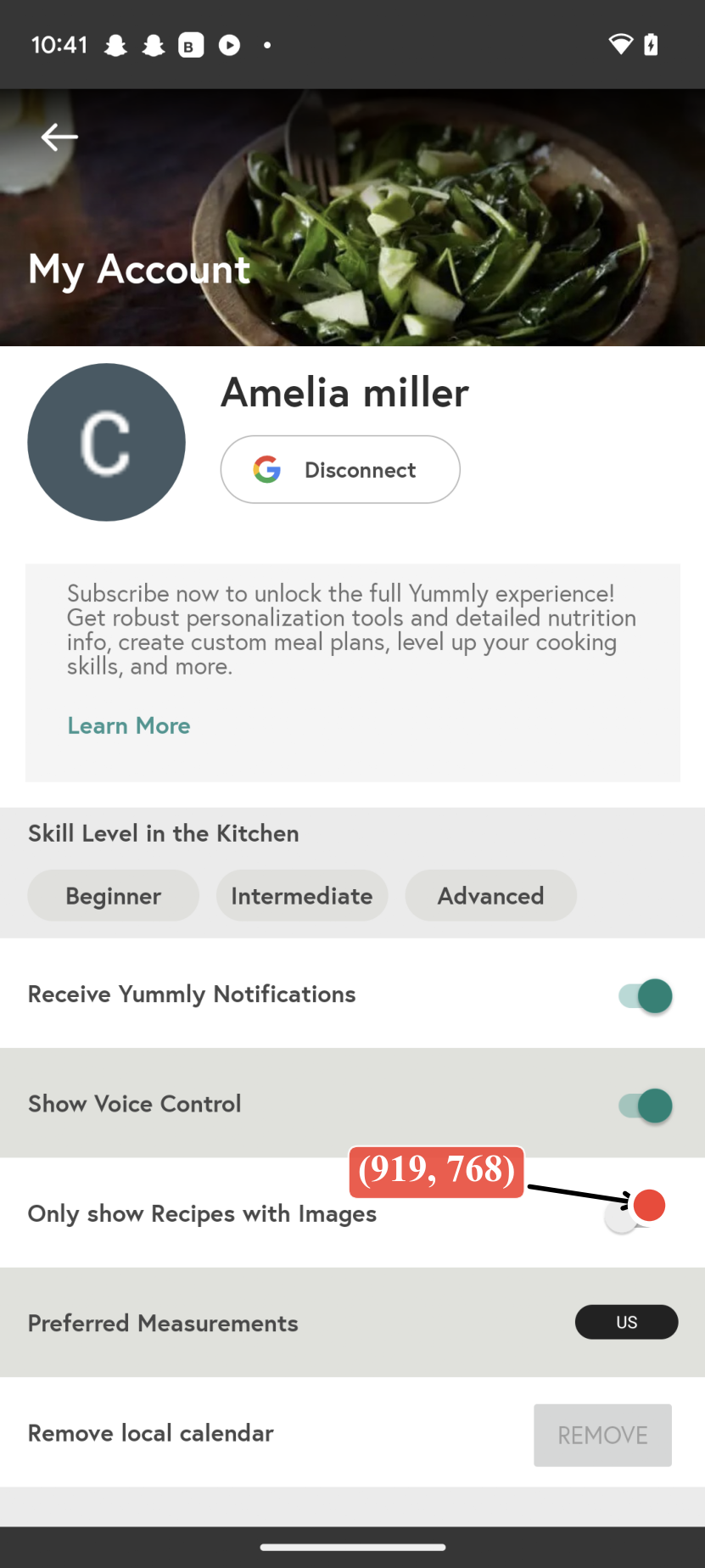}
        &
        \includegraphics[width=0.23\textwidth]{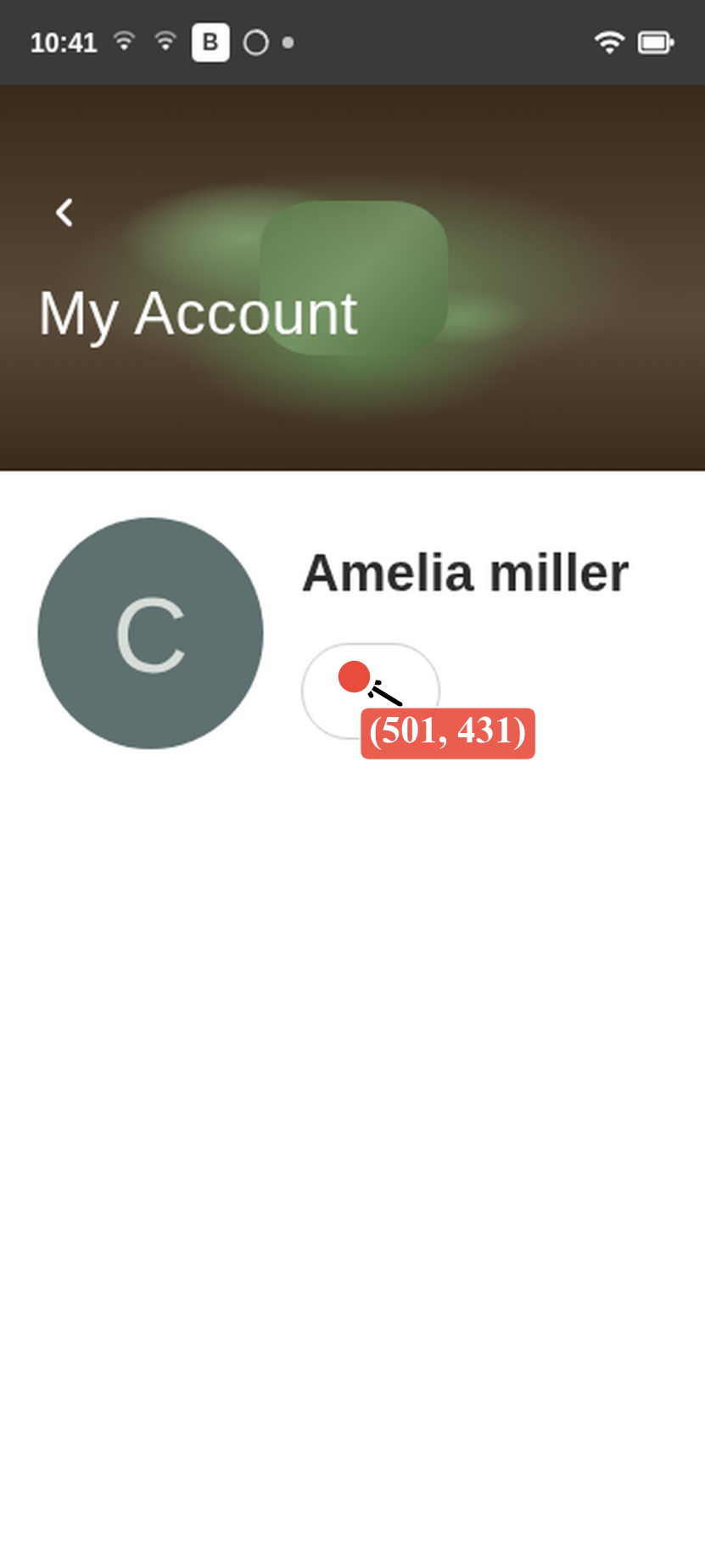}
        \\[-0.1em]
        \multicolumn{2}{c}{\textbf{(b) My Account}}
    \end{tabular}

    \caption{
    Qualitative comparison between real Android screens and corresponding world-model renderings.
    Red markers indicate the selected action coordinates.
    }
    \label{fig:real_vs_rendered}
\end{figure}

Figure~\ref{fig:real_vs_rendered} illustrates how subtle rendering discrepancies between real Android screenshots and world-model outputs lead to systematic click-coordinate bias. In case~(a), the real Stopwatch screen places the Lap button at the left edge ($x{=}149$), whereas the world-model rendering simplifies the layout by removing the Lap button entirely and centering the remaining controls, causing the agent to predict a click near $x{=}499$. In case~(b), the real account page displays a rich set of settings (notification toggles, skill level, measurement preferences), with the target toggle located at the far right ($x{=}919$). The world-model rendering, however, omits most of these elements, leaving only the centered profile area and shifting the predicted click to $x{=}501$. In both cases, the world model's tendency to simplify page layouts and concentrate interactive elements toward the horizontal center alters the perceived interaction logic, producing a systematic centering bias in the collected trajectories that degrades click accuracy after fine-tuning.

\clearpage

\section{Ethics Statement}
We have rigorously refined our dataset to remove any elements that could compromise personal privacy, thereby guaranteeing the highest level of protection for individual data. Instruction evolution was completed by AI SoTA close-sourced VLMs, to whom we paid the necessary compensation to ensure that the training data was not leaked. The human evaluation of our work was carried out through a meticulously randomized selection of IT professionals. This process ensured a gender-balanced and educationally diverse panel, reflecting a wide spectrum of perspectives and expertise.

\section{Reproducibility statement}
All models and datasets used in this paper are open-source. The full experimental setup is detailed in the settings part. Unless noted, all experiments use the same settings.

\section{Limitations}
\label{sec:limitations}


Current mobile world models rendered to pages produce more general results, potentially failing to capture app-specific layout features and element styles. However, this doesn't affect the World Model's impact on downstream tasks, as element styles aren't information-intensive elements in mobile screenshots. A better solution is to achieve more reliable modeling through a method similar to RAG (Rapid Image Processing).

Furthermore, modeling different modalities is not mutually exclusive; they can be combined to mitigate and compensate for each other's shortcomings. However, this paper's primary research objective is to analyze the fundamental characteristics of each individual modality. More complex modeling methods are not the main focus of this paper but rather serve as a guide for future research.

\section{Broader Impact}
\label{sec:impact}

This work may have positive impacts by improving the reliability and safety of mobile GUI agents. 
By enabling agents to anticipate action consequences before execution, mobile GUI world models can reduce unnecessary interactions, improve sample efficiency, and help agents avoid obviously incorrect or risky actions.

However, stronger mobile GUI agents may also introduce risks, including unintended automation on third-party applications, exposure of private information, and execution of high-risk or irreversible actions. 
Inaccurate world-model predictions may further mislead downstream agents if used without safeguards. 
Therefore, real-world deployment should include user confirmation for sensitive operations, action-level safety constraints, privacy-preserving data handling, and careful evaluation under failure cases.

\newpage
\section*{NeurIPS Paper Checklist}

\begin{enumerate}

\item {\bf Claims}
    \item[] Question: Do the main claims made in the abstract and introduction accurately reflect the paper's contributions and scope?
    \item[] Answer: \answerYes{} 
    \item[] Justification: We claims the contributions and scope in the abstract and introduction.

\item {\bf Limitations}
    \item[] Question: Does the paper discuss the limitations of the work performed by the authors?
    \item[] Answer: \answerYes{} 
    \item[] Justification:  We discuss the limitations in the Appendix \ref{sec:limitations}

\item {\bf Theory assumptions and proofs}
    \item[] Question: For each theoretical result, does the paper provide the full set of assumptions and a complete (and correct) proof?
    \item[] Answer: \answerNA{} 
    \item[] Justification: This paper focuses on empirical analysis and does not include theoretical results.

    \item {\bf Experimental result reproducibility}
    \item[] Question: Does the paper fully disclose all the information needed to reproduce the main experimental results of the paper to the extent that it affects the main claims and/or conclusions of the paper (regardless of whether the code and data are provided or not)?
    \item[] Answer:  \answerYes{} 
    \item[] Justification: We clearly describe the experimental settings for all experiments.

\item {\bf Open access to data and code}
    \item[] Question: Does the paper provide open access to the data and code, with sufficient instructions to faithfully reproduce the main experimental results, as described in supplemental material?
    \item[] Answer: \answerNo{} 
    \item[] Justification:  We provide detailed experimental settings, data construction procedures, and evaluation protocols for all experiments. However, the full code and processed data are not publicly released at submission time, and we plan to release them upon publication.

\item {\bf Experimental setting/details}
    \item[] Question: Does the paper specify all the training and test details (e.g., data splits, hyperparameters, how they were chosen, type of optimizer) necessary to understand the results?
    \item[] Answer: \answerYes{} 
    \item[] Justification: We clearly describe the experimental settings for all experiments.

\item {\bf Experiment statistical significance}
    \item[] Question: Does the paper report error bars suitably and correctly defined or other appropriate information about the statistical significance of the experiments?
    \item[] Answer: \answerNo{} 
    \item[] Justification: We do not report error bars for all experiments due to the high computational cost of repeated training and evaluation. We instead use fixed evaluation protocols and consistent test splits for fair comparison.

\item {\bf Experiments compute resources}
    \item[] Question: For each experiment, does the paper provide sufficient information on the computer resources (type of compute workers, memory, time of execution) needed to reproduce the experiments?
    \item[] Answer: \answerNo{} 
    \item[] Justification: We provide detailed experimental settings for all experiments, including model configurations, training data, evaluation protocols. However, we do not provide complete compute-resource information, such as the exact worker type, memory usage, and execution time for every experiment.
    
\item {\bf Code of ethics}
    \item[] Question: Does the research conducted in the paper conform, in every respect, with the NeurIPS Code of Ethics \url{https://neurips.cc/public/EthicsGuidelines}?
    \item[] Answer: \answerYes{} 
    \item[] Justification: This paper adheres to the NeurIPS Code of Ethics. All datasets and models used in this work are properly cited from their original sources.

\item {\bf Broader impacts}
    \item[] Question: Does the paper discuss both potential positive societal impacts and negative societal impacts of the work performed?
    \item[] Answer: \answerYes{} 
    \item[] Justification: The broader impact of this work is discussed in appendix \ref{sec:impact}
    
\item {\bf Safeguards}
    \item[] Question: Does the paper describe safeguards that have been put in place for responsible release of data or models that have a high risk for misuse (e.g., pre-trained language models, image generators, or scraped datasets)?
    \item[] Answer: \answerNA{} 
    \item[] Justification: This paper does not release high-risk pretrained models, or large-scale scraped datasets. Therefore, safeguards for responsible release of such high-risk artifacts are not applicable.

\item {\bf Licenses for existing assets}
    \item[] Question: Are the creators or original owners of assets (e.g., code, data, models), used in the paper, properly credited and are the license and terms of use explicitly mentioned and properly respected?
    \item[] Answer: \answerYes{} 
    \item[] Justification: We comply with the guidelines, and all creators or original owners of assets (e.g., code, data, models) used in this paper are properly credited. The licenses and terms of use are explicitly mentioned and fully respected.

\item {\bf New assets}
    \item[] Question: Are new assets introduced in the paper well documented and is the documentation provided alongside the assets?
    \item[] Answer: \answerNA{} 
    \item[] Justification: This paper does not introduce new assets such as datasets, benchmarks, or released model checkpoints. Therefore, asset documentation is not applicable.

\item {\bf Crowdsourcing and research with human subjects}
    \item[] Question: For crowdsourcing experiments and research with human subjects, does the paper include the full text of instructions given to participants and screenshots, if applicable, as well as details about compensation (if any)? 
    \item[] Answer:  \answerNA{} 
    \item[] Justification: This paper does not involve crowdsourcing experiments or research with human subjects. Therefore, participant instructions, screenshots, and compensation details are not applicable.

\item {\bf Institutional review board (IRB) approvals or equivalent for research with human subjects}
    \item[] Question: Does the paper describe potential risks incurred by study participants, whether such risks were disclosed to the subjects, and whether Institutional Review Board (IRB) approvals (or an equivalent approval/review based on the requirements of your country or institution) were obtained?
    \item[] Answer: \answerNA{} 
    \item[] Justification: This paper does not involve human-subject studies, crowdsourcing experiments, or user studies. Therefore, participant risks, risk disclosure, and IRB approval are not applicable.

\item {\bf Declaration of LLM usage}
    \item[] Question: Does the paper describe the usage of LLMs if it is an important, original, or non-standard component of the core methods in this research? Note that if the LLM is used only for writing, editing, or formatting purposes and does \emph{not} impact the core methodology, scientific rigor, or originality of the research, declaration is not required.
    \item[] Answer: \answerNA{} 
    \item[] Justification: LLMs are not used as an important, original, or non-standard component of the core methodology. Any use of LLMs is limited to writing, editing, or formatting assistance and does not affect the scientific methodology, rigor, or originality of the research.

\end{enumerate}

\end{document}